\documentclass[final,5p,times,twocolumn]{elsarticle}
\biboptions{sort&compress} 
\usepackage{amsmath,amssymb,amsfonts}
\usepackage{algorithmic}
\usepackage[caption=false]{subfig}

\usepackage{newtxtext,newtxmath}  

\usepackage{graphicx}
\usepackage{textcomp}
\usepackage{xcolor}

\usepackage{array}       
\usepackage{booktabs}    
\usepackage{tabularx}
\usepackage{multirow}
\usepackage[table]{xcolor}
\usepackage[linesnumbered,ruled]{algorithm2e}
\usepackage[colorlinks=true, 
            linkcolor=blue,
            citecolor=blue,
            urlcolor=blue]{hyperref}

\usepackage{nameref}
\usepackage{makecell}
\usepackage[utf8]{inputenc}
\usepackage{enumitem}

\journal{Information Fusion}
\begin{document}

\begin{frontmatter}

\title{A self-evolving multi-role collaborative framework with fine-grained difficulty guidance for innovative mathematical problem generation}



\author[label1]{Yifei Sun\corref{cor1}} 
\ead{yifeis@snnu.edu.cn} 
\cortext[cor1]{Corresponding author at: School of Physics and
Information Technology, Shaanxi Normal University, Xi’an 710119, China}

\author[label1]{Yongan Li}
\author[label2]{A.K. Qin}
\author[label1]{Sicheng Hou}
\author[label3]{Tamas Pflanzner}

\affiliation[label1]{organization={School of Physics and
Information Technology, Shaanxi Normal University},
            city={Xi’an},
            postcode={710119},
            country={China}}

\affiliation[label2]{organization={Department of Computing Technologies, Swinburne University of Technology},
            city={Melbourne},
            postcode={3122},
            country={Australia}}

\affiliation[label3]{organization={Department of Software Engineering, University of Szeged},
            city={Szeged},
            postcode={6720},
            country={Hungary}}

\begin{abstract}
Mathematical problem generation (MPG) is a significant research direction in the field of intelligent education. In recent years, the rapid development of large language models (LLMs) has enabled new technological approaches to problem-generation tasks. Although existing LLMs can achieve high correctness rates, they generally lack innovation and exhibit poor discrimination. In this paper, we propose the task of innovative math problem generation (IMPG). To solve the IMPG task, this paper proposes a self-evolving, multi-role collaborative framework with fine-grained difficulty guidance. First, a multi-role collaborative mechanism comprising a sampler, generator, evaluator, state machine, and memory is constructed, ensuring the correctness of generated problems through iterative optimization informed by self-assessment and external feedback. Second, we introduce an improved difficulty model to quantify difficulty and provide fine-grained guidance. We adopt the data-driven association-guided path sampling (DAPS) algorithm to enhance the semantic rationality of sampled encodings. Third, we construct the HSM3K-CN dataset, which comprises high-quality high school math problems. A multi-stage training pipeline is adopted, incorporating continual pre-training (CPT), supervised fine-tuning (SFT), and group relative policy optimization (GRPO), to enhance the generation and evaluation capabilities of the base model. Finally, system self-evolution is achieved by transferring evaluation capabilities from the expert model to the apprentice model via distillation. Experiments show that, compared to baseline models, our proposed method significantly improves the innovation of the generated problems while maintaining a high correctness rate.
\end{abstract}




\end{frontmatter}

\section{Introduction}
Mathematical exercises serve as an essential means for objectively assessing students' cognitive states. Mathematics problem banks require a substantial volume of high-quality problems for continuous iteration and expansion. However, designing high-quality mathematical problems typically requires domain experts to invest substantial time and effort, making this process labor-intensive and unscalable. This consequently limits the supply of personalized teaching and assessment resources. As a result, automatic problem generation (APG) technology for mathematics has emerged in response to this demand. It has rapidly become a research hotspot in the field of artificial intelligence in education.

In the early stages of research on problem generation, researchers focused on using traditional natural language processing (NLP) techniques to address the automatic generation of math word problems (MWPs). Early studies primarily adopted template-based generation approaches\cite{xu_procedural_2021}, utilizing predefined problem templates and mathematical domain knowledge to generate problems. With the advancement of deep learning techniques, research began shifting toward neural network-driven generation methods. The core innovation of this phase was the introduction of attention mechanisms to model the complex correspondences between mathematical expressions and problem text\cite{wu_automatic_2022,zhou_towards_2019}. However, deficiencies remained in terms of controllability during the generation process. Consequently, researchers further focused on achieving more fine-grained control during generation. On one hand, answer set programming-based methods were introduced\cite{polozov2015personalized}, implementing personalized problem generation through declarative constraints and discourse-tropes systems. On the other hand, research began to develop targeted constraint mechanisms\cite{wang_math_2021} to model mathematical consistency explicitly.

In recent years, the rise of LLMs has brought revolutionary breakthroughs to MPG through their powerful capabilities in natural language understanding, logical reasoning, and content generation. The application of LLMs in MPG tasks has begun to attract research attention\cite{maity_future_2024}. Early researchers first focused on evaluating the fundamental capabilities of general-purpose LLMs such as ChatGPT in MPG tasks\cite{pham_chatgpt_2024}. As research progressed, simple instruction-based generation could no longer meet the requirements. Fine-grained controllable generation has become a mainstream direction. Researchers considered multi-objective modeling for MPG, combining external knowledge bases and adaptive retrieval-augmented generation (RAG) frameworks to improve the alignment accuracy between generated content and multiple objectives\cite{sun_multi-objective_2025}. Compared to external knowledge injection relying on retrieval augmentation, other studies focus on building specialized domain models through deeper model fine-tuning\cite{feng_exploring_2024,chen_controlmath:_2024,liu_comet_2024,sun_multi-objective_2025,christ2024mathwell}. With the development of LLM technology, multi-agent collaboration has emerged as a new direction for addressing APG problems. Several studies have proposed various agent collaboration paradigms that jointly solve complex APG problems through division of labor, evaluation and discussion, and iterative refinement among multiple agents\cite{tonga2025automatic,chen_controlmath:_2024,sun_multi-objective_2025,karbasi2025multi,lee2025vista}.

\begin{figure}[!htbp]
\centering
\includegraphics[width=\columnwidth]{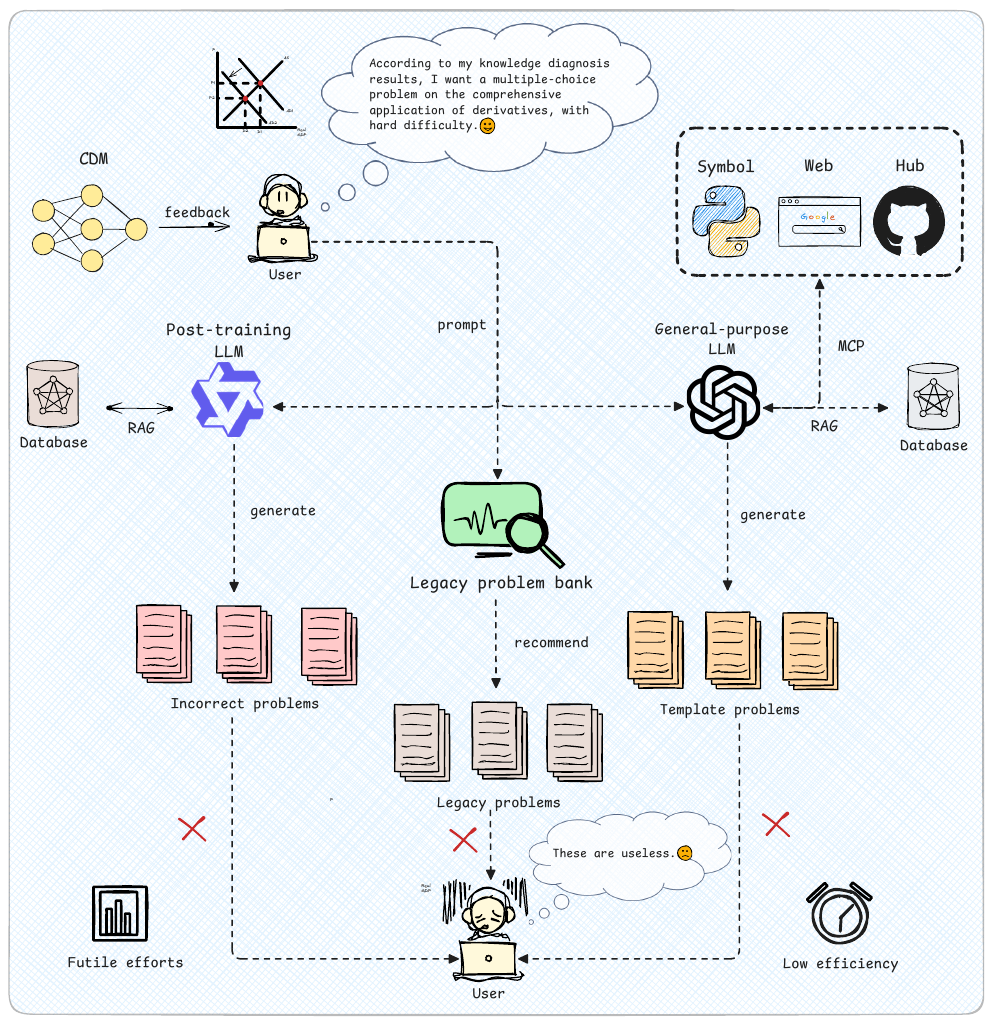}
    \caption{The limitations of the post-training LLMs and general-purpose LLMs in IMPG task. CDM represents the cognitive diagnosis model. MCP denotes the model context protocol for standardized communication between LLMs and external tools. These limitations will lead to futile efforts and low efficiency for users.}
\label{fig_1}
\end{figure}

Despite significant advances in controllability and diversity achieved by existing APG methods, which demonstrate high accuracy, these methods still face numerous challenges in practical applications. First, whether using open-source models or closed-source commercial models, the generated problems commonly suffer from insufficient innovation and poor discrimination. One of the core reasons is the inability to control problem difficulty at a finer granularity. Second, regarding the IMPG task, we observed that after the base model underwent SFT, the innovation of the generated problems improved, while their correctness declined. This phenomenon reveals an inherent conflict between correctness and innovation in MPG tasks, which we term the \textbf{Innovation Curse}. These limitations are illustrated in \hyperref[fig_1]{Fig. 1}.

To address the two challenges above, this paper proposes a self-evolving, multi-role collaborative framework with fine-grained difficulty guidance. First, we design a multi-role collaboration mechanism and achieve iterative optimization of problem quality via two modes: \textbf{apprentice mode} and \textbf{expert mode}. In the apprentice mode, a single LLM serves as both the generator and evaluator, whereas the expert mode introduces an expert LLM as the evaluator to provide more accurate evaluation. Second, we introduce an improved difficulty model during fine-tuning and inference that employs a 16-bit difficulty encoding scheme to quantify problem difficulty combinatorially. We utilize a data-driven association-guided path sampling algorithm that mines encoding association patterns from historical problems. This approach maps problem difficulty levels to 16-bit difficulty encodings, which enhance the LLM's ability to generate innovative mathematics problems after decoding. Finally, we implement system self-evolution via expert-to-apprentice distillation, which improves the apprentice's evaluation capability. For SFT training and data distillation, we construct the HSM3K-CN dataset containing 3,160 high-quality high school mathematics problems and adopt a multi-stage training pipeline that incorporates continual pre-training, SFT, and GRPO to enhance the generation and evaluation capabilities of the base model. Experimental results demonstrate that, compared to baseline models, our method significantly improves problem innovation and discrimination while maintaining high correctness rates.

The contributions of this paper can be summarized as follows:

\begin{itemize}[leftmargin=*,noitemsep,topsep=0pt]
\item We propose a multi-role collaborative framework where a single LLM plays multiple roles and incorporates external expert guidance. Through self-assessment and expert feedback, the framework improves the correctness of generated problems.

\item An improved difficulty model is introduced to the MPG process for the first time, enhancing the LLM's ability to generate high-innovation mathematics problems. A data-driven association-guided path sampling algorithm is proposed to ensure the semantic validity of sampling, which mines encoding association patterns from historical problems.

\item This work pioneers the integration of a self-evolution method into the training process for the MPG task, which enhances the evaluation performance of the apprentice model while reducing the reliance on manually annotated data.

\item A Chinese high school mathematics dataset HSM3K-CN containing 3,160 raw samples is established, from which we derive two training subsets HSM3K-CN-gen and HSM3K-CN-eval containing 6,320 samples in total. The generation and evaluation capabilities of the base model are significantly enhanced through fine-tuning with these datasets.
\end{itemize}

\section{Related work}
LLMs have demonstrated powerful content generation capabilities in the educational field. Existing research can be primarily categorized into two types: single-LLM-based methods and multi-agent collaborative frameworks. This section will elaborate on the specific related work from these two perspectives, laying the foundation for our proposed method.
\subsection{Large language model}
In early research on problem generation, researchers primarily focused on utilizing structured methods to generate problems. Nandhini et al.\cite{nandhini2011math} compared two fundamental approaches for generating MWPs: the template-based method and the context-free grammar, providing foundational insights for subsequent, more complex generation models. Xu et al.\cite{xu_procedural_2021} proposed a general mathematical problem PCG pipeline that offered a modular, highly controllable production pipeline that significantly reduced manual editing time. Polozov et al.\cite{polozov2015personalized} pioneered the integration of personalized learning concepts into MWP generation by proposing a system based on answer-set programming. Compared with structured methods that rely on predefined rules and logic programming, subsequent research increasingly shifted toward neural networks for end-to-end MWP generation to achieve greater flexibility and higher generation quality. Zhou et al.\cite{zhou_towards_2019} were the first to propose an end-to-end neural network model called MAGNET for generating MWPs from given equations and topic words. Wu et al.\cite{wu_automatic_2022} introduced a novel model named MWPGen that extracts associative information between topic words and expressions to ensure the relevance of generated problems. Liu et al.\cite{liu2021mathematical} proposed a model called MaKE that integrates external commonsense knowledge graphs with mathematical equations for problem generation. Wang et al.\cite{wang_math_2021} combined pre-trained language models and introduced two major constraints to enhance generation quality. Jiao et al.\cite{jiao2023automatic} decomposed difficulty into mathematical difficulty and linguistic difficulty, employing energy-based language models and samplers to generate high-quality and diverse MWPs across different difficulty levels. Additionally, Niyarepola et al.\cite{niyarepola2022math} utilized multilingual pre-trained models for MWP generation, providing a viable pathway for building educational resources in low-resource languages. Yuan et al.\cite{yuan2020automatic} investigated how to automatically generate concise headlines for lengthy and complex online math problems. The research by Shridhar et al.\cite{shridhar2022automatic} did not focus on generating problems themselves, but rather on generating socratic subquestions to guide students in problem-solving. Although Jiao et al.\cite{jiao2023automatic} pioneered difficulty decomposition in MWP generation, this method can only perform difficulty decomposition for simple MWP problems with linear logic, and is entirely infeasible for complex MPG tasks.

With the emergence of LLMs, researchers have shifted toward leveraging their powerful generation capabilities for MPG tasks.
Feng et al.\cite{feng_exploring_2024} explored the effectiveness of utilizing LLMs to automatically generate distractors for math multiple-choice problems. Liu et al.\cite{liu_comet_2024} fine-tuned the COMET multimodal model using the tower construction method for controllable generation, simulation generation, and problem-solving. Xie et al.\cite{xie2024adversarial} pioneered a new paradigm for generating adversarial MWPs by converting original math problems into code and constructing abstract syntax trees, then generating new problems by modifying numerical values at the tree's leaf nodes while imposing educational constraints. Shah et al.\cite{shah2024ai} proposed an "AI-assisted, human-in-the-loop" framework aimed at generating high-difficulty math problems that require the integration of multiple skills. Chen et al.\cite{chen_controlmath:_2024} introduced a data generation method called ControlMath, designed to address the problem of model overfitting on specific datasets. To address the issue of monotonous problem scenarios caused by limited input information in traditional models, Qin et al.\cite{qin2024math} proposed an MWP generation method based on decoupled memory retrieval. In the process of utilizing AI to assist teaching, Tonga et al.\cite{tonga2025automatic} explored the use of LLM to automatically generate guiding hints for students who answered incorrectly. In educational practice, problems not only need to be correct but also require different cognitive depths to meet diverse instructional needs at various levels. Addressing this need, Yu et al.\cite{yu2025recall} developed a framework called QG-DOK aimed at automatically generating math problems with varying cognitive depths. Regarding the evaluation of existing LLMs' capabilities in MPG, Pham et al.\cite{pham_chatgpt_2024} conducted the first comprehensive and in-depth assessment of ChatGPT's ability to generate pre-college mathematics problems. Maity et al.\cite{maity_future_2024} provided a systematic review of the potential of using LLMs for MPG and answer evaluation. To make generated math problems more relevant to students' real lives, Hwang et al.\cite{hwang2024using} proposed an innovative system that generates geometry word problems with different difficulty levels by identifying real-world contexts. To better meet fine-grained control requirements, Sun et al.\cite{sun_multi-objective_2025} proposed a multi-objective math problem generation task and designed an adaptive multi-level retrieval-augmented framework. Building on the above research, scholars further focused on ensuring the educational appropriateness and authenticity of generated problems, as well as their ability to satisfy multiple pedagogical objectives. Christ et al.\cite{christ2024mathwell} emphasized that an educational MWP must not only be solvable with accurate answers but also be appropriate in theme, difficulty, and language for K-8 grade students. Mahran et al.\cite{mahran2025investigating} designed and implemented an automated multilingual pipeline to systematically investigate language bias in LLMs' mathematical problem-solving capabilities. 

\subsection{Multi-agent collaboration}
Researchers have found that single models often encounter bottlenecks when handling highly complex reasoning or generation tasks. To overcome this limitation, multi-agent systems, which leverage multiple agents for collaborative work, have emerged as a promising research direction. Jin et al.\cite{jin2025comprehensive} provided a comprehensive review and classification of multi-agent collaborative decision-making. In a survey, Sharma et al.\cite{sharma2025small} proposed a disruptive viewpoint that when building agent systems, "bigger" is not always "better". Huang et al.\cite{huang2025routereval} introduced RouterEval, a comprehensive benchmark specifically designed for LLM routing, exploring how to intelligently select the most appropriate model for specific tasks within a system containing multiple models. Hu et al.\cite{hu2024self} proposed EvoMAC, a self-evolving multi-agent collaborative network paradigm for software development. Chen et al.\cite{chen2025tumix} introduced the TUMIX framework, a multi-agent approach for test-time scaling via "tool-use mixture."

Multi-agent collaboration has proven effective in solving complex scientific problems. Chen et al.\cite{chen2024comm} proposed the CoMM framework aimed at enhancing LLMs' ability to solve complex scientific problems. Lei et al.\cite{lei2024macm} introduced the MACM prompting method, specifically designed to solve complex mathematical problems, which abstracts problems into a series of conditions and an ultimate goal. Also focusing on zero-shot mathematical problem-solving, Keating et al.\cite{keating2024zero} proposed a two-agent collaborative framework called Aurek through the lens of conversation programming. Mushtaq et al.\cite{mushtaq2025harnessing} designed a multi-agent LLM framework to assist with senior design projects in engineering programs.

Multi-agent frameworks not only effectively solve problems but also demonstrate tremendous potential in generating high-quality, customized problems. Karbasi et al.\cite{karbasi2025multi} proposed a multi-agent collaborative framework for MPG, aimed at dynamically generating math problems that meet specific difficulty and cognitive requirements for intelligent tutoring systems. Lee et al.\cite{lee2025vista} developed the VISTA framework, a multi-agent system focused on utilizing LLM to automatically generate math problems with visual aids. Our work is inspired by this but adopts a hybrid multi-role collaborative framework that differs from traditional multi-agent collaboration. Its distinctive feature is that it avoids the high computational overhead associated with deploying and maintaining multiple independent agents. Moreover, our unique quintuple collaborative framework can enhance problem innovation while ensuring that the correctness rate does not decline.

Furthermore, we observe that these LLMs exert only coarse-grained control over problem generation in the MPG process, predominantly focusing on relatively simple MWPs at the K-9 grades. When it comes to complex MPG tasks, existing approaches demonstrate limited innovation. Our work incorporates an improved difficulty model into MPG tasks to enhance problem innovation and realize self-evolution of the collaborative framework via distillation.

\section{Methodology}
In this section, we first define the IMPG task that incorporates a difficulty model within a multi-role collaborative framework. Subsequently, we introduce the individual components of the framework, the difficulty model, sampling methods, and the training and self-evolution pipeline.

\subsection{Task definition}
\begin{figure*}[tbp]
\centering
\includegraphics[width=0.90\textwidth]{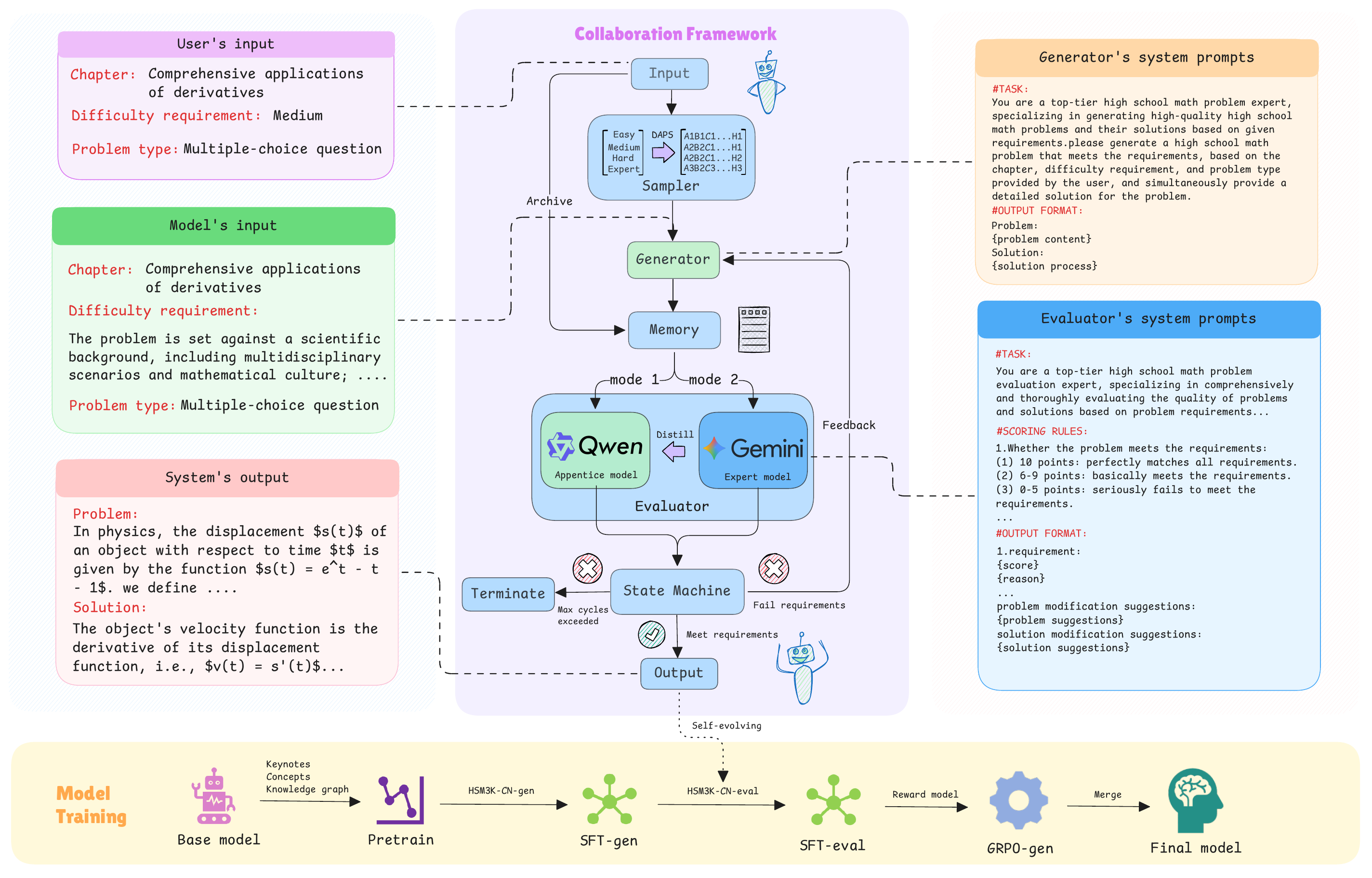}
\caption{Overall framework for multi-role collaboration.}
\label{fig_2}
\end{figure*}

Given a user's problem requirement $q = \{C^*, D_{level}, T\}$, where $C^*$ represents the target chapter, $D_{level} \in \{\text{Easy, Medium, Hard, Expert}\}$ denotes four difficulty levels, and $T$ signifies the problem type. This study introduces a fine-grained difficulty model that maps the difficulty requirement into a 16-bit encoding space $\mathcal{E} = \{L\omega \mid L \in \mathcal{L}, \omega \in [1, n_L]\}$, where $\mathcal{L} = \{A, B, C, D, E, F, G, H\}$ denotes eight dimensions of difficulty and $n_L \in \{2, 3, 4\}$ denotes the number of different levels per dimension. The objective of the IMPG task is to generate a mathematical problem $\mathcal{P} = \{M, S\}$, where $M$ is the problem statement and $S$ is the solution. The generation process must satisfy the following constraints:

\begin{equation}
\begin{aligned}
&\mathcal{P}^* = \arg\max_{\mathcal{P}} \mathbb{P}(\mathcal{P} | q, \mathcal{E}) \cdot \mathcal{I}(\mathcal{P}) \\
&\text{s.t.} \quad \mathcal{C}_{\text{correctness}}(\mathcal{P}) \wedge \mathcal{C}_{\text{difficulty}}(\mathcal{P}, D_{level})
\end{aligned}
\end{equation}
where $\mathcal{E}$ is a 16-bit encoding combination sampled from the difficulty encoding space, $\mathcal{I}(\mathcal{P})$ is the problem innovation evaluation model. $\mathcal{C}_{\text{correctness}}$ and $\mathcal{C}_{\text{difficulty}}$ represent the mathematical correctness and difficulty consistency constraints, respectively. 

To achieve the aforementioned objective, the sampler employs the DAPS algorithm to sample a compliant encoding combination $\mathcal{E} = \mathcal{S}(D_{level})$ from the difficulty encoding space. The generator then produces a preliminary problem $\mathcal{P}_{\text{pre}} = \mathcal{G}(q')$ based on the extended requirement $q' = \{C^*, \mathcal{E}, T\}$. The evaluator conducts a multi-dimensional assessment of the generated result, yielding $\{s, r\} = \mathcal{V}(\mathcal{P}_{\text{pre}}, q)$, where $s$ is a comprehensive quality score and $r$ represents revision suggestions. The iterative refinement process is formulated as:

\begin{equation}
\mathcal{P}^{(t+1)} = \begin{cases}
\mathcal{P}^{(t)}, & \text{if } s^{(t)} \geq \tau \\
\mathcal{G}(q', r^{(t)}, \mathcal{P}^{(t)}), & \text{if } s^{(t)} < \tau
\end{cases}
\end{equation}
where $\tau$ is a predefined quality threshold.

\subsection{Multi-role collaborative framework}
The overall framework for multi-role collaborative generation is illustrated in \hyperref[fig_2]{Fig. 2}, which clearly shows the interactions among the components. The complete system prompts for the generator and evaluator can be found in \hyperref[appendix_A]{Appendix A}. The specific implementation process for problem generation is detailed in \hyperref[alg_1]{Algorithm 1}.

\subsubsection{Quintet-collaboration role framework and closed-loop optimization}
We construct a novel quintet-collaboration system comprising the sampler, generator, evaluator, state machine, and a memory module. Each role has distinct responsibilities, yet cooperates with others to form a complete closed-loop MPG process.

The sampler is the starting point of the entire process. Its core task is to convert the problem difficulty level into a 16-bit difficulty encoding, which is subsequently decoded into specific, structured difficulty requirements. Upon first receiving the structured input from the sampler, the generator begins creating the specific mathematical problem and its solution. Starting from the second iteration, the generator receives feedback from the evaluator, progressively correcting errors, optimizing expressions to improve the core innovation of the problem. The evaluator rigorously evaluates the problems generated by the generator and provides corresponding suggestions for revision. Notably, the evaluator is designed as a completely stateless role. The state machine is the system's decision-making hub, responsible for determining the following action based on the evaluation results. It outputs the final result when the quality standard is met, returns to the generator for further improvement if the quality is substandard and the maximum number of iterations has not been reached, and makes a termination decision when the maximum iteration count is reached. The memory module manages framework state by storing and updating three core elements: the user's problem requirements, current problem content, and corresponding solution.

\subsubsection{Homogeneous multi-role collaboration mechanism based on a single large model}
Traditional multi-agent collaborative frameworks often rely on a combination of heterogeneous models, with different roles performed by distinct models or model variants. Although this design can leverage the specific strengths of each model, it also introduces issues, including high coordination costs, complex deployment, and knowledge inconsistency between models. We propose a novel homogeneous multi-role collaboration mechanism based on a single fine-tuned large model.

Within this framework, both the generator and evaluator roles are performed by the same domain-fine-tuned large model. However, through different system prompts and dialogue state management mechanisms, they exhibit distinct behavioral patterns. The generator role focuses on creating high-quality mathematics problems and their solutions based on chapters, difficulty levels, and problem types. In contrast, the evaluator role adopts the perspective of an expert in problem design, conducting a comprehensive review of the generated problems across ten fine-grained evaluation dimensions.

\begin{algorithm}[!htbp] 
\caption{Multi-role collaborative mathematical problems generation process}
\KwIn{$R$: User requirement (chapter, difficulty levels, problem type); $M$: Operation mode (apprentice or expert); $\tau_{max}$: Maximum iteration cycles; $\theta$: Quality thresholds for evaluation dimensions}
\KwOut{$P_{final}$: Final problem statement; $S_{final}$: Final solution}

\tcp{Initialization}
Initialize Generator $\mathcal{G}$, Evaluator $\mathcal{E}$, State Machine $\mathcal{D}$, Memory $\mathcal{M}$\;
$cycle \leftarrow 0$\;
$state \leftarrow$ "active"\;

\tcp{Difficulty encoding via DAPS}
$D_{decoded} \leftarrow \text{DAPS}(R.difficulty)$\;
$R_{formatted} \leftarrow \text{Format}(R.chapter, D_{decoded}, R.type)$\;

\tcp{Initial generation}
$(P_0, S_0) \leftarrow \mathcal{G}.\text{Generate}(R_{formatted})$\;
$\mathcal{M}.\text{Initialize}(R_{formatted}, P_0, S_0)$\;
$cycle \leftarrow cycle + 1$\;

\tcp{Iterative refinement loop}
\While{$state$ == "active" \textbf{and} $cycle \leq \tau_{max}$}{
    \tcp{Evaluation phase}
    $Input_{eval} \leftarrow \mathcal{M}.\text{GetFormattedMemory}()$\;
    
    \eIf{$M$ == "apprentice mode"}{
        $Eval_{output} \leftarrow \mathcal{E}_{local}.\text{Evaluate}(Input_{eval})$\;
    }{
        $Eval_{output} \leftarrow \mathcal{E}_{expert}.\text{Evaluate}(Input_{eval})$\;
    }
    
    \tcp{Score parsing}
    $Scores \leftarrow \text{ParseScores}(Eval_{output})$\;
    $Feedback \leftarrow \text{ExtractFeedback}(Eval_{output})$\;
    
    \tcp{State machine operation}
    $(action, suggestions) \leftarrow \mathcal{D}.\text{Judge}(Scores, Feedback, cycle, R_{formatted}, \theta)$\;

    \tcp{Action execution}
    \uIf{$action$ == "output\_results"}{
        $state \leftarrow$ "completed"\;
        $P_{final} \leftarrow \mathcal{M}.problem$\;
        $S_{final} \leftarrow \mathcal{M}.solution$\;
    }
    \uElseIf{$action$ == "return\_generator"}{
        $Input_{gen} \leftarrow \text{FormatFeedback}(R_{formatted}, suggestions)$\;
        $(P_{new}, S_{new}) \leftarrow \mathcal{G}.\text{Refine}(Input_{gen})$\;
        $\mathcal{M}.\text{Update}(P_{new}, S_{new})$\;
        $cycle \leftarrow cycle + 1$\;
    }
    \Else{
        \tcp{Max\_cycles\_exceeded}
        $state \leftarrow$ "terminated"\;
        $P_{final} \leftarrow \mathcal{M}.problem$\;
        $S_{final} \leftarrow \mathcal{M}.solution$\;
    }
}

\Return{$P_{final}$, $S_{final}$}
\label{alg_1}
\end{algorithm}

\subsubsection{Dual-mode mechanism through organic integration of internal optimization and external guidance}
The superior evaluation capability of the evaluator stems from its unique dual-mode design, namely the \textbf{apprentice mode} and the \textbf{expert mode}. These two modes correspond to two evaluation paths of internal optimization and external guidance, respectively.

On the one hand, the apprentice mode is essentially an internal iterative optimization mechanism based on the homogeneous model. In this mode, the fine-tuned local model undertakes evaluation responsibilities. The evaluator critically reviews the generator's output, identifies existing problems, and provides specific revision suggestions. This mode offers the advantage of cost-effectiveness but is also limited by the single model's capability boundary and blind-spot consistency.

On the other hand, the expert mode introduces an external, top-tier large model to act as the evaluator. Leveraging its superior reasoning abilities, broader knowledge base, and more nuanced judgment criteria, the evaluator can identify and correct more subtle issues, thereby raising generation quality. However, this mode also incurs additional API call costs and network latency.

\subsection{Data-driven association-guided path sampling}
\subsubsection{Requirements}
In IMPG tasks, precisely quantifying problem difficulty has consistently been a core challenge. Traditional metrics of problem difficulty are often coarse-grained descriptions for problem generation with LLMs. The core requirements of this study are to quantify problem difficulty using the difficulty model before problem generation, generate problem-encoding combinations via a sampling algorithm, and obtain specific, structured difficulty requirements upon decoding.

However, simple random sampling (RS) cannot guarantee coherence of the combinations. For example, a problem involving simple numerical computation might be paired with a requirement for a deeply innovative contextual design. Constrained random sampling (CRS) based on human expert rules can eliminate some unreasonable combinations. However, this process is not only time-consuming and labor-intensive, but also becomes increasingly difficult to formulate manual rules when the dimensions of the difficulty model increase or when complex nonlinear correlations exist among dimensions.

Therefore, the DAPS algorithm is developed. It mines encoding co-occurrence patterns from historical problem data and constructs a probability transition matrix to guide path search. Its high semantic validity and automated nature make it a reliable choice for constructing complex problem difficulty models.

\subsubsection{Difficulty model}
The improved comprehensive difficulty model established in this study decomposes problem difficulty into eight relatively independent dimensions, each further characterized by its distinct encoding levels. The specific difficulty dimensions, encoding levels, connotations, and their corresponding weight grades are detailed in \hyperref[appendix_B]{Appendix B}.

Let the encoding space be $\mathcal{L} = \{A, B, C, D, E, F, G, H\}$, and the set of levels corresponding to each dimension is denoted as ${L}\omega$. 
A complete difficulty encoding is an eight-tuple vector:
\begin{equation}
\mathbf{x} = (a, b, c, d, e, f, g, h) \in
\{L\omega \mid L \in \mathcal{L}, \omega \in [1, n_L]\}
\end{equation}
where each component represents the selected level value for the corresponding dimension. For example, $\mathbf{x} = (1, 2, 1, 1, 2, 1, 1, 2)$ corresponds to the encoding A1B2C1D1E2F1G1H2. 
Notably, $\mathbf{x}$ carries a dual meaning, representing both the encoding level and the weight grade.

To reflect the differential contribution of each dimension to the overall difficulty, a weight vector $\boldsymbol{\sigma} = (\sigma_A, \sigma_B, \ldots, \sigma_H)^T$ is introduced.
The difficulty coefficient is defined as the normalized summation of the weighted levels of each dimension:
\begin{equation}
D(\mathbf{x}) = \frac{1}{|\mathcal{L}|}
\sum_{L \in \mathcal{L}} \sigma_L \cdot x_L = \frac{1}{8}
\sum_{i=1}^{8} \sigma_i \cdot x_i
\label{eqa4}
\end{equation}
where $x_i$ represents the weight grade of the $i$-th component of the vector $\mathbf{x}$. This weight grade distinguishes the varying impact of different levels within each difficulty factor on the overall difficulty coefficient.

Based on cognitive theory and pedagogical practice, the difficulty coefficient is divided into four grade intervals: $\mathcal{I}_{Easy}$, $\mathcal{I}_{Medium}$, $\mathcal{I}_{Hard}$, and $\mathcal{I}_{Expert}$.

\subsubsection{Algorithm}
\paragraph{Problem description}
Let each specific level in the encoding space be represented as a node. Define the set of nodes as $\mathcal{V}$. Let $\mathcal{X}_{\text{valid}} \subset \mathcal{V}^8$ represent the subset of valid encodings. The sampling problem can be formalized as given a target difficulty level $\Omega \in \{\text{Easy}, \text{Medium}, \text{Hard}, \text{Expert}\}$, the objective is to generate an encoding $\mathbf{x}^* \in \mathcal{X}_{\text{valid}}$ such that $D(\mathbf{x}^*) \in \mathcal{I}_\Omega$. The complete data-driven association-guided path sampling algorithm is described in \hyperref[alg_2]{Algorithm 2}.

\paragraph{Construction of the association network and transition matrix}

The foundation of the DAPS algorithm is to learn the association structure between encoding nodes from historical data. Assume there is a labeled dataset $\mathcal{D} = \{\mathbf{x}^{(1)}, \mathbf{x}^{(2)}, \ldots, \mathbf{x}^{(N)}\}$ containing $N$ problems, where each $\mathbf{x}^{(k)}$ is a complete difficulty encoding vector.

For any two nodes $v_i, v_j \in \mathcal{V}$, their Jaccard association coefficient is defined as:
\begin{equation}
J(v_i, v_j) = \frac{C_{ij}}{N_i + N_j - C_{ij}}
\end{equation}
where $C_{ij}$ denotes the number of problems containing both encoding $i$ and $j$, and $N_i$ and $N_j$ denote the total number of problems containing encoding $i$ and $j$, respectively. This coefficient measures the relative frequency of two encodings co-occurring in actual problems, with a value range of $[0, 1]$.

Construct a $21 \times 21$ raw association matrix $\mathbf{J}$, whose $(i,j)$-th element is $J_{ij} = J(v_i, v_j)$. It should be noted that the Jaccard coefficient itself is symmetric, i.e., $J_{ij} = J_{ji}$, but directionality must be introduced when constructing the transition probabilities.

To convert the association strength into transition probabilities, each column of the matrix is normalized. Specifically, for a node $v_j$, its transition probability vector to all other nodes is defined as:
\begin{equation}
P_{ij} = \frac{J_{ij}}{\sum_{k=1}^{21} J_{kj}}
\end{equation}
The resulting matrix $\mathbf{P} = [P_{ij}]_{21 \times 21}$ is thus a column stochastic matrix, satisfying $\sum_{i=1}^{21} P_{ij} = 1, \; \forall j$. The matrix element $P_{ij}$ represents the probability of transitioning to node $v_i$ next, given that the current state is node $v_j$.

It is worth noting that although the original Jaccard matrix is symmetric, the transition matrix $\mathbf{P}$ is generally non-symmetric, i.e., $P_{ij} \neq P_{ji}$, because the normalization factors for each column differ. This non-symmetry reflects the true conditional probability relationship $P(v_i | v_j) \neq P(v_j | v_i)$. For example, almost all problems contain $G_1$, but only some problems contain $H_3$. Thus, $P(H_3|G_1)$ will be small, while $P(G_1|H_3)$ may be large. This asymmetry causes the encoding distributions generated from different starting points and paths to vary, thereby enhancing the diversity of the sampling.

\paragraph{Random walk and dimensional constraints}

A random walk process subject to dimensional constraints is executed on the probability space defined by the data-driven association transition matrix $\mathbf{P}$ to generate the difficulty encoding. The constraints are hard, i.e., prohibiting repeated visits to the same dimension and requiring complete traversal of all dimensions. Although the transition probabilities depend on the history of visited dimensions, the process constitutes a Markov chain on an augmented state space when the state $S_t = (V_t, \mathcal{D}_t)$ comprises the current node and the set of visited dimensions.

\textbf{Initialization:} Randomly select a starting node $V_0 \sim \text{Uniform}(\mathcal{V})$ from the node set $\mathcal{V}$ according to a uniform distribution. Define a node-to-dimension mapping function $\text{dim}: \mathcal{V} \to \mathcal{L}$, for example, $\text{dim}(A2) = A$. Initialize the set of visited dimensions $\mathcal{D}_0 = \{\text{dim}(V_0)\}$ and the path sequence $\mathcal{T}_0 = \{V_0\}$.

\textbf{Iterative transition:} At time step $t$ ($t = 0, 1, \ldots, 6$), perform the following steps:

First, candidate space construction. Define the set of currently transferable nodes as $\mathcal{C}_t = \{v \in \mathcal{V} : \text{dim}(v) \notin \mathcal{D}_t\}$
This ensures the monotonicity constraint of dimensions—each dimension can only be visited once. Second, conditional transition probability calculation. Let the current node be $V_t = v_j$. Extract the $j$-th column from the transition matrix $\mathbf{P}$ to obtain the original transition probability vector $\mathbf{p}_j = (P_{1j}, P_{2j}, \ldots, P_{21,j})^T$. Perform probability truncation and re-normalization on the candidate nodes:
\begin{equation}
\tilde{P}(v_i | V_t = v_j, \mathcal{D}_t) =
\begin{cases}
\frac{P_{ij}}{\sum_{v_k \in \mathcal{C}_t} P_{kj}} & \text{if } v_i \in \mathcal{C}_t \\
0 & \text{otherwise}
\end{cases}
\end{equation}
This defines a conditional transition probability that depends on the history of visited states. Third, probabilistic sampling. Sample the next node according to the re-normalized probability distribution $V_{t+1} \sim \text{Categorical}\{\tilde{P}(v | V_t, \mathcal{D}_t) : v \in \mathcal{C}_t\}$. Finally, state update. According to $\mathcal{D}_{t+1} = \mathcal{D}_t \cup \{\text{dim}(V_{t+1})\}$ and $\mathcal{T}_{t+1} = \mathcal{T}_t \cup \{V_{t+1}\}$, we update the set of visited dimensions and the path sequence.


\textbf{Termination condition:} When $|\mathcal{D}_t| = 8$, the walk terminates. At this point, the path $\mathcal{T} = \{V_0, V_1, \ldots, V_7\}$ contains nodes from 8 different dimensions.

\textbf{Path standardization:} Since the walk order is random, the generated path may not conform to the standard dimensional order. The path needs to be rearranged according to the dimensional lexicographical order $\mathbf{x} = \text{sort}(\mathcal{T}, \text{key}=\text{dim})$. The sorting is performed according to the lexicographical order of dimension labels A-H. This yields the encoding vector in standard form.

\paragraph{Rejection sampling and difficulty constraints}

While the random walk ensures combinatorial validity, a rejection sampling mechanism is employed to enforce the target difficulty constraints.

\textbf{Batch generation strategy:} To improve efficiency, $M$ independent random walks are executed simultaneously in each round to generate a set of candidate encodings $\mathcal{S}_m = \{\mathbf{x}^{(1)}, \mathbf{x}^{(2)}, \ldots, \mathbf{x}^{(M)}\}$ where each $\mathbf{x}^{(k)}$ is produced by an independent random walk process.

\textbf{Difficulty calculation:} For each candidate encoding $\mathbf{x} \in \mathcal{S}_m$, calculate its difficulty coefficient by \hyperref[eqa4]{Eq. (4)}.

\begin{algorithm}[!htbp]
\caption{Data-driven association-guided path sampling}
\KwIn{$\mathbf{P}$: Transition probability matrix ($21 \times 21$); $\Omega$: Target difficulty level; $M$: Batch size; $\boldsymbol{\sigma}$: Weight vector; $\mathcal{I}_\Omega$: Difficulty interval for level $\Omega$; $\mathcal{M}$: Difficulty model mapping dictionary}
\KwOut{$\mathbf{x}^*$: A valid difficulty encoding; $\mathbf{R}$: Decoded difficulty requirements}
$\mathcal{S}_{\text{accept}} \leftarrow \emptyset$\;
\While{$\mathcal{S}_{\text{accept}} = \emptyset$}{
    \tcp{Batch generation of candidate encodings}
    \For{$m = 1$ \KwTo $M$}{
        \tcp{Initialize random walk}
        $V_0 \sim \text{Uniform}(\mathcal{V})$\;
        $\mathcal{D}_0 \leftarrow \{\text{dim}(V_0)\}$\;
        $\mathcal{T}_0 \leftarrow \{V_0\}$\;
        $t \leftarrow 0$\;
        \tcp{Constrained random walk process}
        \While{$|\mathcal{D}_t| < 8$}{
            $\mathcal{C}_t \leftarrow \{v \in \mathcal{V} : \text{dim}(v) \notin \mathcal{D}_t\}$\;
            \tcp{Compute conditional transition probabilities}
            \ForEach{$v_i \in \mathcal{C}_t$}{
                $\tilde{P}(v_i | V_t, \mathcal{D}_t) \leftarrow \frac{P_{ij}}{\sum_{v_k \in \mathcal{C}_t} P_{kj}}$, where $V_t = v_j$\;
            }
            \tcp{Sample next node}
            $V_{t+1} \sim \text{Categorical}\{\tilde{P}(v | V_t, \mathcal{D}_t) : v \in \mathcal{C}_t\}$\;
            \tcp{Update state}
            $\mathcal{D}_{t+1} \leftarrow \mathcal{D}_t \cup \{\text{dim}(V_{t+1})\}$\;
            $\mathcal{T}_{t+1} \leftarrow \mathcal{T}_t \cup \{V_{t+1}\}$\;
            $t \leftarrow t + 1$\;
        }
        \tcp{Standardize path to encoding}
        $\mathbf{x}^{(m)} \leftarrow \text{sort}(\mathcal{T}_t, \text{key}=\text{dim})$\;
        \tcp{Calculate difficulty coefficient}
        $D(\mathbf{x}^{(m)}) \leftarrow \frac{1}{8} \boldsymbol{\sigma}^T \mathbf{x}^{(m)}$\;
        \tcp{Constraint verification}
        \If{$D(\mathbf{x}^{(m)}) \in \mathcal{I}_\Omega$}{
            $\mathcal{S}_{\text{accept}} \leftarrow \mathcal{S}_{\text{accept}} \cup \{\mathbf{x}^{(m)}\}$\;
        }
    }
}
\tcp{Randomly select from accepted encodings}
$\mathbf{x}^* \sim \text{Uniform}(\mathcal{S}_{\text{accept}})$\;
\tcp{Decode encoding to difficulty requirements}
$\mathbf{R} \leftarrow \text{Decode}(\mathbf{x}^*, \mathcal{M})$\;
\Return{$\mathbf{x}^*$, $\mathbf{R}$}
\label{alg_2}
\end{algorithm}

\textbf{Decision criterion:}
Given the target difficulty level $\Omega$, filter the encodings that satisfy the constraint $\mathcal{S}_{\text{accept}} = \{\mathbf{x} \in \mathcal{S}_m : D(\mathbf{x}) \in \mathcal{I}_\Omega\}$. If $\mathcal{S}_{\text{accept}} \neq \emptyset$, the sampling is successful. Randomly select one and return it by $\mathbf{x}^* \sim \text{Uniform}(\mathcal{S}_{\text{accept}})$. If $\mathcal{S}_{\text{accept}} = \emptyset$, reject all candidates in this round and re-execute the walk generation process.

Let $n$ be the $n$-th attempt, the probability of successful sampling is:
\begin{equation}
P_{\text{success}}(n) = 1 - [1 - P(\mathcal{S}_{\text{accept}} \neq \emptyset)]^n
\end{equation}
This probability approaches 1 exponentially fast as $n$ increases, ensuring the algorithm has a finite expected termination time.

Define the constraint satisfaction rate for a single walk as $\alpha = P(D(\mathbf{x}) \in \mathcal{I}_\Omega \, | \, \mathbf{x} \sim \text{DAPS})$, where $\mathbf{x} \sim \text{DAPS}$ denotes an encoding generated by the DAPS algorithm. This probability depends on the match between the encoding distribution guided by the transition matrix $\mathbf{P}$ and the target difficulty interval. Since $\mathbf{P}$ is learned from real problem bank data, the generated encoding distribution will naturally concentrate in the dense regions of the historical data. Therefore, $\alpha$ is generally much higher than the acceptance rate of fully random sampling.

\subsection{Post-training and self-evolving}

\begin{figure*}[htbp]
\centering
\includegraphics[width=0.90\textwidth]{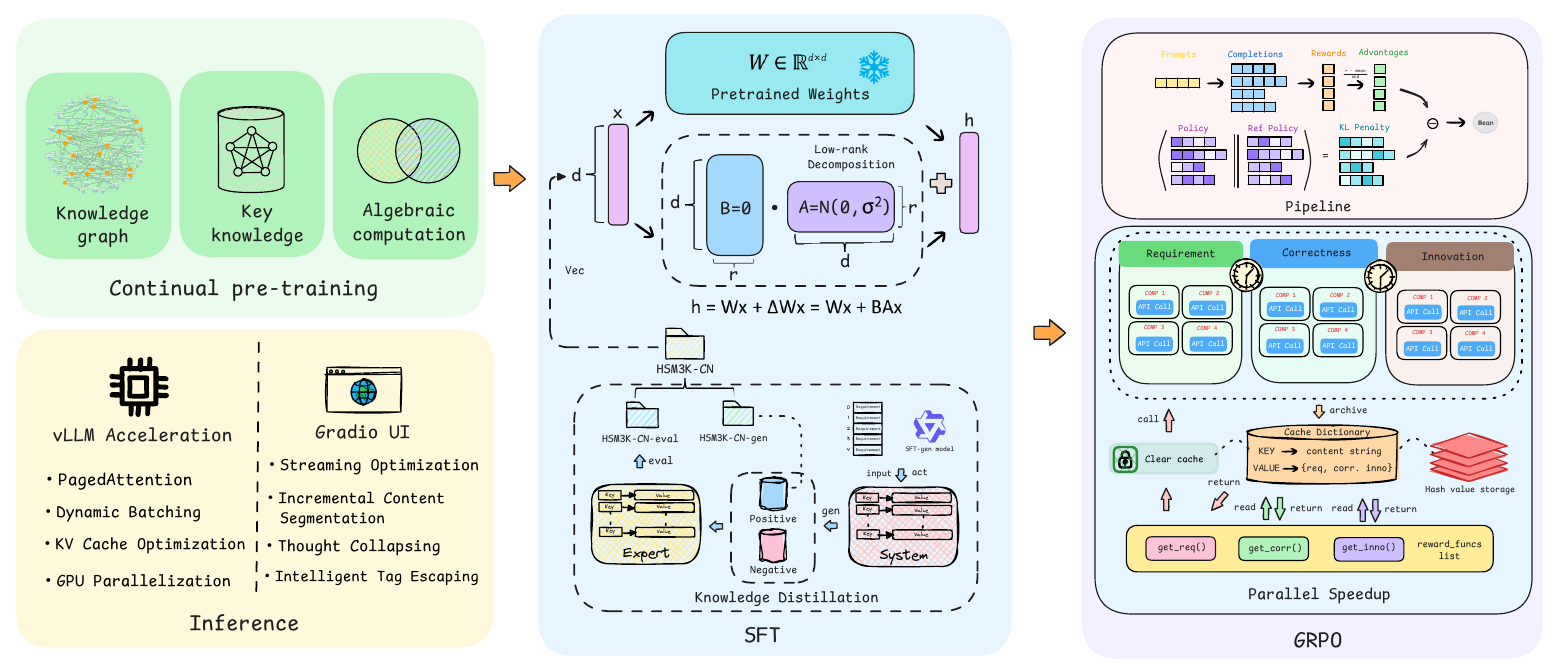}
\caption{The complete post-training pipeline for self-evolving MPG. The pipeline consists of four stages: CPT with knowledge graph, key knowledge, and algebraic computation; SFT using LoRA fine-tuning on HSM3K-CN datasets and implementing knowledge distillation from expert model; GRPO with parallel reward computation; vLLM acceleration and Gradio UI optimization for efficient inference.}
\label{fig_3}
\end{figure*}

We selected the three most representative core chapters from high school mathematics for post-training: derivative, sequence, and probability. The average difficulty of problems in these three chapters is relatively high, and existing large models face considerable difficulty in generating high-quality problems for them. This selection fully reflects the challenging aspects of the IMPG task. Simultaneously, it allows us to concentrate our limited computational resources on addressing the core problems. The entire post-training pipeline is shown in \hyperref[fig_3]{Fig. 3}.

\subsubsection{Continual pre-training}
We conducted continual pre-training on the base model, aiming to enhance the large model's domain knowledge in high school mathematics. We categorized the knowledge for continual pre-training into three components: the high school mathematics knowledge graph, key knowledge from the three selected chapters, and algebraic computation.

We constructed a high school mathematics \textbf{knowledge graph}, comprising 2,399 knowledge points, 12 relation types, and 11,250 knowledge relation pairs. Through data preprocessing, two types of samples were derived. The first type comprises knowledge concept samples, which are represented as a two-tuple \textit{(knowledge point, concept 
description)}. The second type comprises knowledge point relation samples, structured as a four-tuple \textit{(knowledge point node 1, knowledge point node 2, relation, description)}. Based on the Qwen tokenizer, the token size for this dataset partition is approximately 480K.

The \textbf{key knowledge} component encompasses the core concepts, problem-solving techniques, and example problems for the three chapters. Through processes including OCR extraction, data cleaning, and data annotation on textbooks, reference books, and other materials, we obtained the data structured as a four-tuple \textit{(chapter, knowledge points, problem-solving techniques, example problems)}. Based on the Qwen tokenizer, the token size for this dataset partition is approximately 38K.

The \textbf{algebraic computation} component is primarily intended to reduce the probability of numerical calculation errors during the LLM's inference. We utilized the algebra dataset constructed by Yang et al.\cite{yang2023gpt}, which contains a large volume of algebraic expressions. We further supplemented this dataset by adding the computational results to the original expressions. Based on the Qwen tokenizer, the token size for this dataset partition is also approximately 480K.

\subsubsection{Supervised fine-tuning}
The first stage of SFT was conducted following continual pre-training. We utilized the HSM3K-CN-gen dataset to fine-tune the model's generation capability, i.e., the generator role. This dataset was derived from 3,160 high-quality high school mathematics problems and constructed through preprocessing steps, including OCR extraction, data cleaning, and data annotation. During the annotation phase, we employed Claude-Sonnet-3.7-Reasoning to select and supplement problem design requirements, i.e., chapter, decoded difficulty requirements, and problem type, as well as the reasoning process for problem design from candidate sets for each problem. The entire annotation results were subsequently verified by mathematics experts. The primary objective of this stage was to enable the model to learn specific paradigms for innovative problem design while maintaining logical consistency.

The second stage of SFT was conducted following SFT-gen. We utilized the HSM3K-CN-eval dataset to fine-tune the model's evaluation capability, i.e., the evaluator role. This dataset comprises 1,580 high-quality high school mathematics problems alongside 1,580 problems of varying quality generated by the SFT-gen fine-tuned system, constructed through preprocessing steps including data cleaning and data annotation. During the annotation phase, we utilized high-quality evaluation data and reasoning processes generated by the expert model, with all annotation results subsequently verified by mathematics experts. The primary objective of this stage was to enable the model to learn precise scoring across each dimension and provide comprehensive revision suggestions.

Furthermore, we incorporate a knowledge distillation mechanism within the system. This process essentially distills the evaluation approaches and judgment criteria of the expert model into the apprentice model, enabling the latter to progressively acquire the evaluation capabilities of the former. Through such knowledge transfer, the performance of the apprentice model is continuously enhanced, gradually narrowing the gap between the expert and apprentice while maintaining lower inference costs. This transformation mechanism from external guidance to internal capability enhancement constitutes the core driving force of the system's self-evolution.

Using the Qwen tokenizer, the total token count of the entire HSM3K-CN dataset is approximately 20M. Detailed formatting specifications are provided in \hyperref[appendix_C]{Appendix C}.

\subsubsection{Group relative policy optimization}
Unlike some traditional reinforcement learning algorithms that require training an additional value model, the GRPO algorithm estimates the baseline in a group-relative manner, significantly reducing training resource consumption. The objective function of GRPO maximizes expected rewards while preventing the policy from deviating excessively from the reference model through KL divergence constraints. For each group of sampled outputs, we first compute the respective reward scores using the reward model, then perform intra-group normalization on these rewards by subtracting the group mean and dividing by the group standard deviation. The normalized rewards serve as advantage values to adjust the gradient coefficients of the policy model at each token position. The objective function of GRPO is:

\begin{align}
&\mathcal{J}_{GRPO}(\theta) = \mathbb{E}_{q \sim P(Q), \{o_i\}_{i=1}^{G} \sim \pi_{\theta_{old}}(O|q)} \nonumber \\
&\frac{1}{G} \sum_{i=1}^{G} \frac{1}{|o_i|} \sum_{t=1}^{|o_i|} \Bigg\{ \min \left[ \frac{\pi_\theta(o_{i,t}|q, o_{i,<t})}{\pi_{\theta_{old}}(o_{i,t}|q, o_{i,<t})} \hat{A}_{i,t}, \right. \nonumber \\
&\left. \mathrm{clip}\left( \frac{\pi_\theta(o_{i,t}|q, o_{i,<t})}{\pi_{\theta_{old}}(o_{i,t}|q, o_{i,<t})}, 1-\varepsilon, 1+\varepsilon \right) \hat{A}_{i,t} \right] \nonumber \\
&- \beta \mathbb{D}_{KL}\left[ \pi_\theta || \pi_{ref} \right] \Bigg\}
\end{align}
where $\pi_\theta$ and $\pi_{\theta_{old}}$ are the current and old policy models, respectively. $q$ denotes a query sampled from the query distribution $P(Q)$, and $\{o_i\}_{i=1}^{G}$ represents a group of $G$ outputs sampled from the old policy $\pi_{\theta_{old}}$ given the query $q$. $|o_i|$ is the length of the $i$-th output, $o_{i,t}$ denotes the $t$-th token of the $i$-th output, and $o_{i,<t}$ represents all tokens before position $t$. $\varepsilon$ is a clipping hyperparameter for stabilizing training. $\hat{A}_{i,t}$ is the advantage computed based on group relative rewards, defined as $\hat{A}_{i,t} = \tilde{r}_i = (r_i - \mathrm{mean}(\mathbf{r})) / \mathrm{std}(\mathbf{r})$, where $\mathbf{r} = \{r_1, r_2, \cdots, r_G\}$ are the rewards for all outputs in the group. $\beta$ is the coefficient of the KL penalty, $\pi_{ref}$ is the reference model, and $\mathbb{D}_{KL}[\pi_\theta || \pi_{ref}]$ is the KL divergence between the current policy and the reference policy.

We employed GRPO to conduct reinforcement training on the generator. The reinforcement learning objective is to maximize the comprehensive quality score of generated problems. For each sample within a group, the reward function was designed based on evaluation across several core dimensions, i.e., requirement, correctness, and innovation, with Gemini-3.0-Pro serving as the evaluator for scoring. The comprehensive reward score for each sample was obtained through weighted summation. We did not apply reinforcement training to the evaluator, as this would likely disrupt the existing generation preference distribution and lead to catastrophic forgetting.

We also designed a two-tier parallel architecture to maximize GRPO training efficiency. Adopting a multi-threaded design, the first tier implements parallelization at the reward-function level, while the second tier operates at the completion level, ultimately achieving high speedup. Additionally, we utilized a caching mechanism to decouple computation from logging, enabling independent recording of each reward score while maintaining parallel processing. This approach effectively addresses the limitations of the TRL library, which natively lacks support for parallel invocation of reward functions and cannot independently record scores during parallel processing.

\section{Experiments}
\subsection{Datasets}
During the CPT stage, we constructed small-scale datasets for knowledge graphs, key knowledge points, and algebraic computation.

During the SFT stage, we established a Chinese high school mathematics dataset, HSM3K-CN, containing 3,160 samples, with statistical information presented in \hyperref[tab_1]{Table 1}. The sub-datasets HSM3K-CN-gen and HSM3K-CN-eval comprised 6,320 samples in total. Specifically, HSM3K-CN-eval contains 1,580 high-quality problems and 1,580 problems of varying quality generated by the multi-role collaborative system, both of which underwent evaluation annotation by the expert model and manual verification.

During the GRPO training stage, for preference learning of the generator, we constructed a task instruction set through random combinations of chapters, difficulty levels, and problem types.

We constructed a test dataset SIMU-90 consisting of 90 problem design requirements to conduct simulation testing under realistic problem design scenarios. These problem design requirements were obtained through random combinations of chapters, difficulty levels, and problem types, with the specific distribution presented in \hyperref[tab_2]{Table 2}.

\begin{table}
\centering
\footnotesize
\caption{Distribution of problem types and knowledge points in HSM3K-CN.}

\setlength{\tabcolsep}{4.5pt}  
\renewcommand{\arraystretch}{1.2}     
\begin{tabular}{lccc}
\toprule[0.5pt] 
Chapter & Problem type &  Problem count &  Knowledge points count \\
\midrule[0.5pt] 
\multirow{3}{*}{Derivative}
& Multiple-choice & 342 & 1728 \\
& Fill-in-the-blank & 172 & 887 \\
& Problem-solving & 532 & 2593 \\
\midrule[0.5pt] 
\multirow{3}{*}{Sequence}
& Multiple-choice & 328 & 1629 \\
& Fill-in-the-blank & 154 & 806 \\
& Problem-solving & 574 & 2705 \\
\midrule[0.5pt] 
\multirow{3}{*}{Probability}
& Multiple-choice & 413 & 2202 \\
& Fill-in-the-blank & 142 & 781 \\
& Problem-solving & 503 & 2550 \\
\midrule[0.5pt] 
Total & - & 3160 & 15881 \\
\bottomrule[0.5pt] 
\end{tabular}
\label{tab_1}
\end{table}

\begin{table}
\centering
\footnotesize
\caption{Distribution of problem types and difficulty levels in SIMU-90.}
\setlength{\tabcolsep}{1.3pt}
\renewcommand{\arraystretch}{1.2}     
\begin{tabular}{lcccccccc}
\toprule[0.5pt] 
Chapter & Count & \makecell{Multiple-\\choice} &  \makecell{Fill-in-\\the-blank} &  \makecell{Problem-\\solving} & Easy & Medium &Hard &Expert \\
\midrule[0.5pt] 
Derivative & 21 & 7 & 8 & 6 & 4 & 7 & 6 & 4 \\
Sequence & 33 & 16 & 12 & 5 & 8 & 8 & 8 & 9 \\
Probability & 36 & 13 & 9 & 14 & 7 & 10 & 12 & 7 \\
\midrule[0.5pt] 
Total & 90 & 36 & 29 & 25 & 19 & 25 & 26  & 20\\
\bottomrule[0.5pt] 
\end{tabular}
\label{tab_2}
\end{table}

\begin{table*}
\centering
\footnotesize
\caption{The specific details of baseline models, including provider, release date, size, open-source status and mathematical evaluation.}
\setlength{\tabcolsep}{5.2pt}
\renewcommand{\arraystretch}{1.5}
\begin{tabular}{lcccccccccc}
\toprule
\multirow{2}{*}[-1.2ex]{Model} & \multirow{2}{*}[-1.2ex]{Provider} & \multirow{2}{*}[-1.2ex]{Release} & \multirow{2}{*}[-1.2ex]{Size} & \multirow{2}{*}[-1.2ex]{Modality} & \multirow{2}{*}[-1.2ex]{\makecell{Context\\length}} & \multirow{2}{*}[-1.2ex]{\makecell{Open-\\sourced}} & \multicolumn{4}{c}{Evaluation} \\
\cmidrule(lr){8-11}
 & & & & & & & \makecell{AIME\\2025} & \makecell{GPQA\\Diamond} & MMLU-Pro & HLE \\

\midrule
\raisebox{-0.2em}{\includegraphics[height=1em]{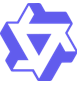}} Qwen3-32B & Alibaba & 07/2025 & 32B & Single & 256K & True & 0.730 & 0.670 & 0.800 & 0.083 \\
\raisebox{-0.2em}{\includegraphics[height=1em]{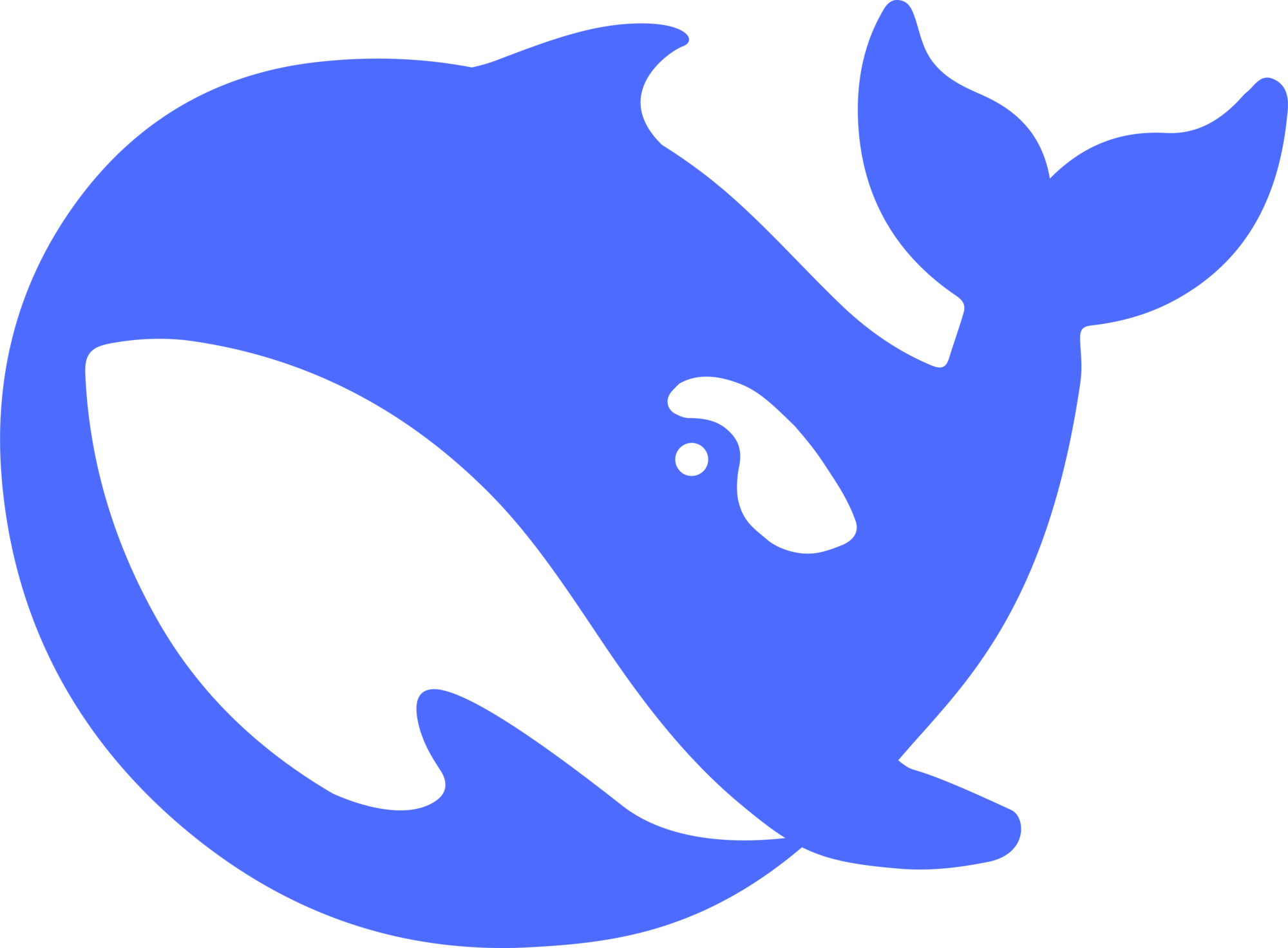}} DeepSeek-R1-Distill-Qwen-32B & Deepseek & 02/2025 & 32B & Single & 128K & True & 0.630 & 0.620 & 0.740 & 0.055 \\
\raisebox{-0.2em}{\includegraphics[height=1em]{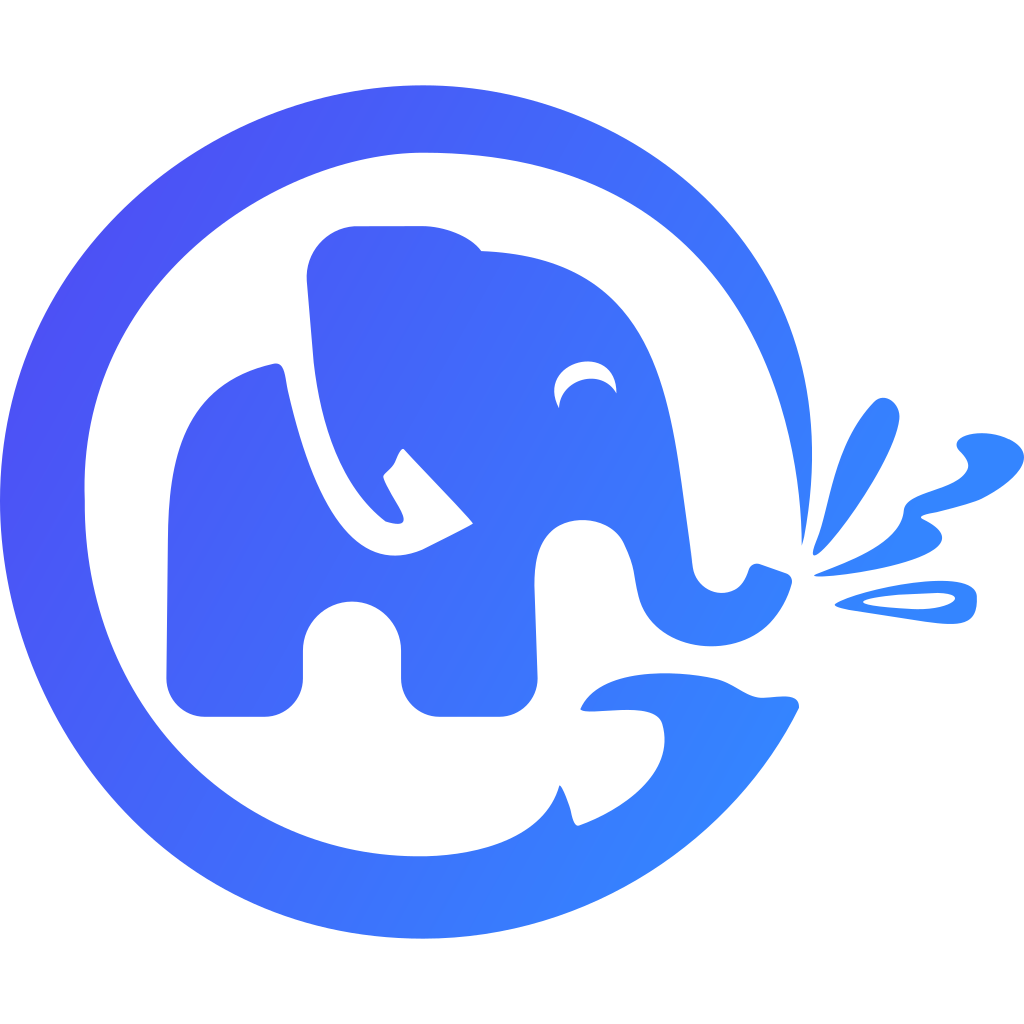}} GLM-4.6 & Zhipu & 09/2025 & 357B & Single & 200K & True & 0.860 & 0.780 & 0.830 & 0.133 \\
\raisebox{-0.2em}{\includegraphics[height=1em]{icons/deepseek.png}} DeepSeek-V3.2-Speciale & Deepseek & 12/2025 & 671B & Single & 128K & True & \colorbox{black!20}{\makebox[3em][c]{0.970}} & \colorbox{black!20}{\makebox[3em][c]{0.870}} & 0.860 & 0.261 \\
\raisebox{-0.2em}{\includegraphics[height=1em]{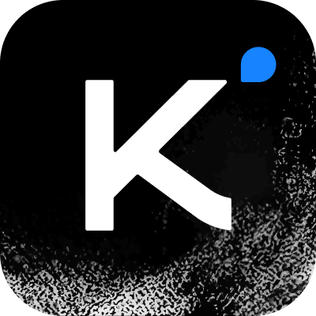}} Kimi-K2-Thinking & Moonshot & 11/2025 & 1T & Single & 256K & True & 0.950 & 0.840 & 0.850 & 0.223 \\
\raisebox{-0.2em}{\includegraphics[height=1em]{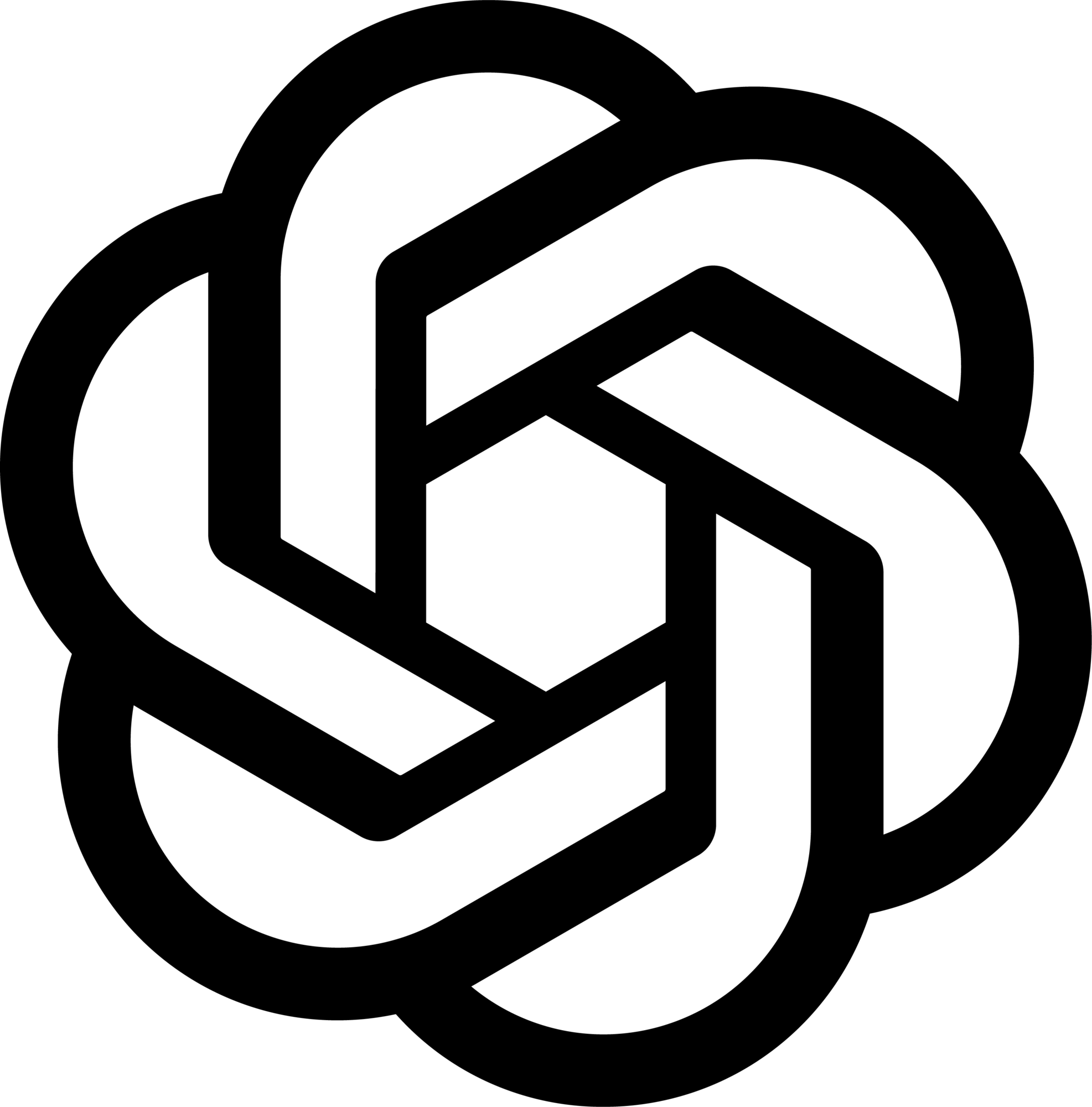}} GPT-OSS-120B (high) & OpenAI & 08/2025 & 117B & Single & 128K & True & 0.930 & 0.780 & 0.810 & 0.185 \\
\raisebox{-0.2em}{\includegraphics[height=1em]{icons/gpt.png}} GPT-5.1 (high) & OpenAI & 11/2025 & - & Multi & 400K & False & 0.940 & \colorbox{black!20}{\makebox[3em][c]{0.870}} & 0.870 & 0.265 \\
\raisebox{-0.2em}{\includegraphics[height=1em]{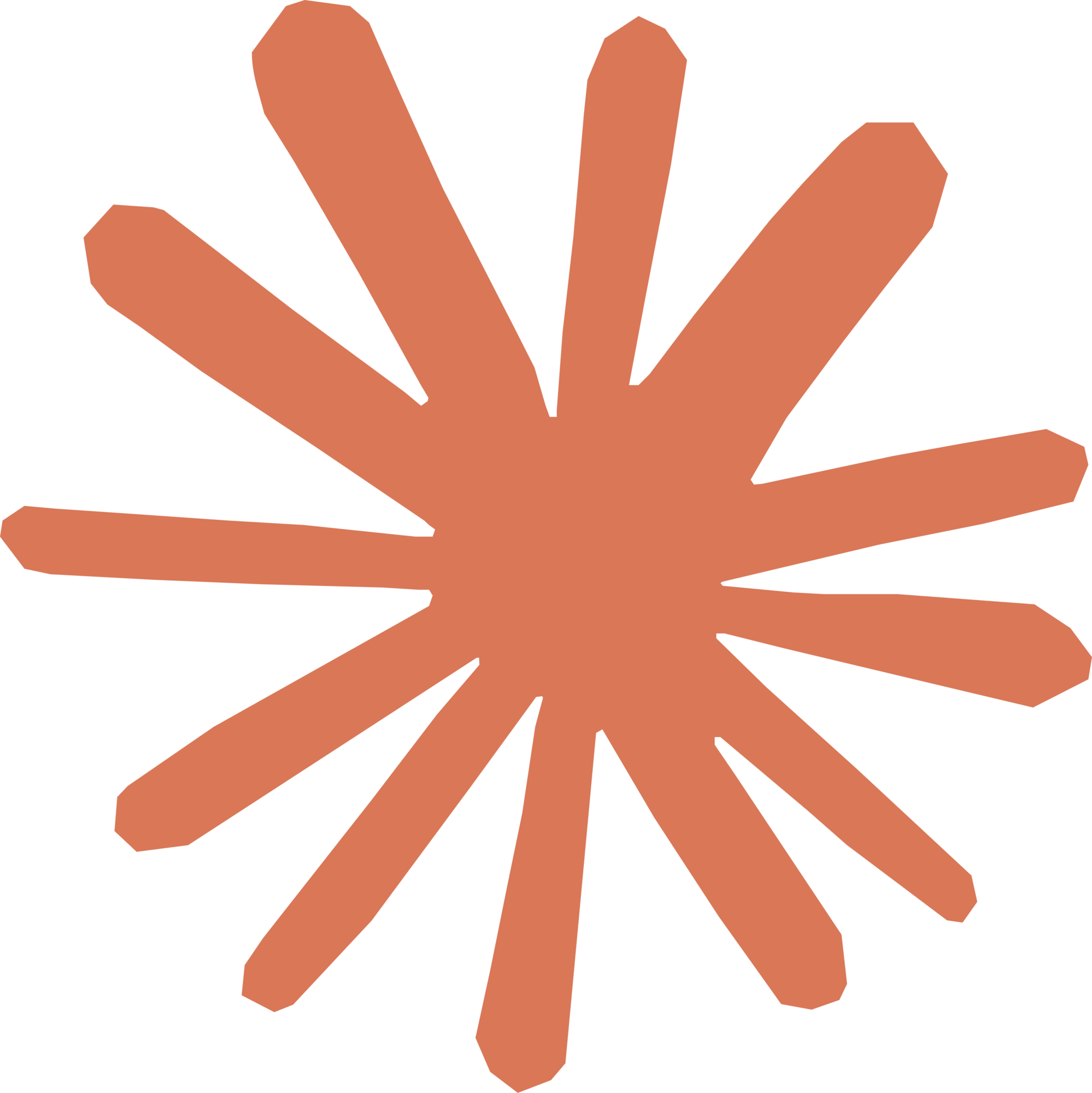}} Claude-Opus-4.5-Reasoning & Anthropic & 11/2025 & - & Multi & 200K & False & 0.910 & \colorbox{black!20}{\makebox[3em][c]{0.870}} & \colorbox{black!20}{\makebox[3em][c]{0.900}} & \colorbox{black!20}{\makebox[3em][c]{0.284}} \\
\raisebox{-0.2em}{\includegraphics[height=1em]{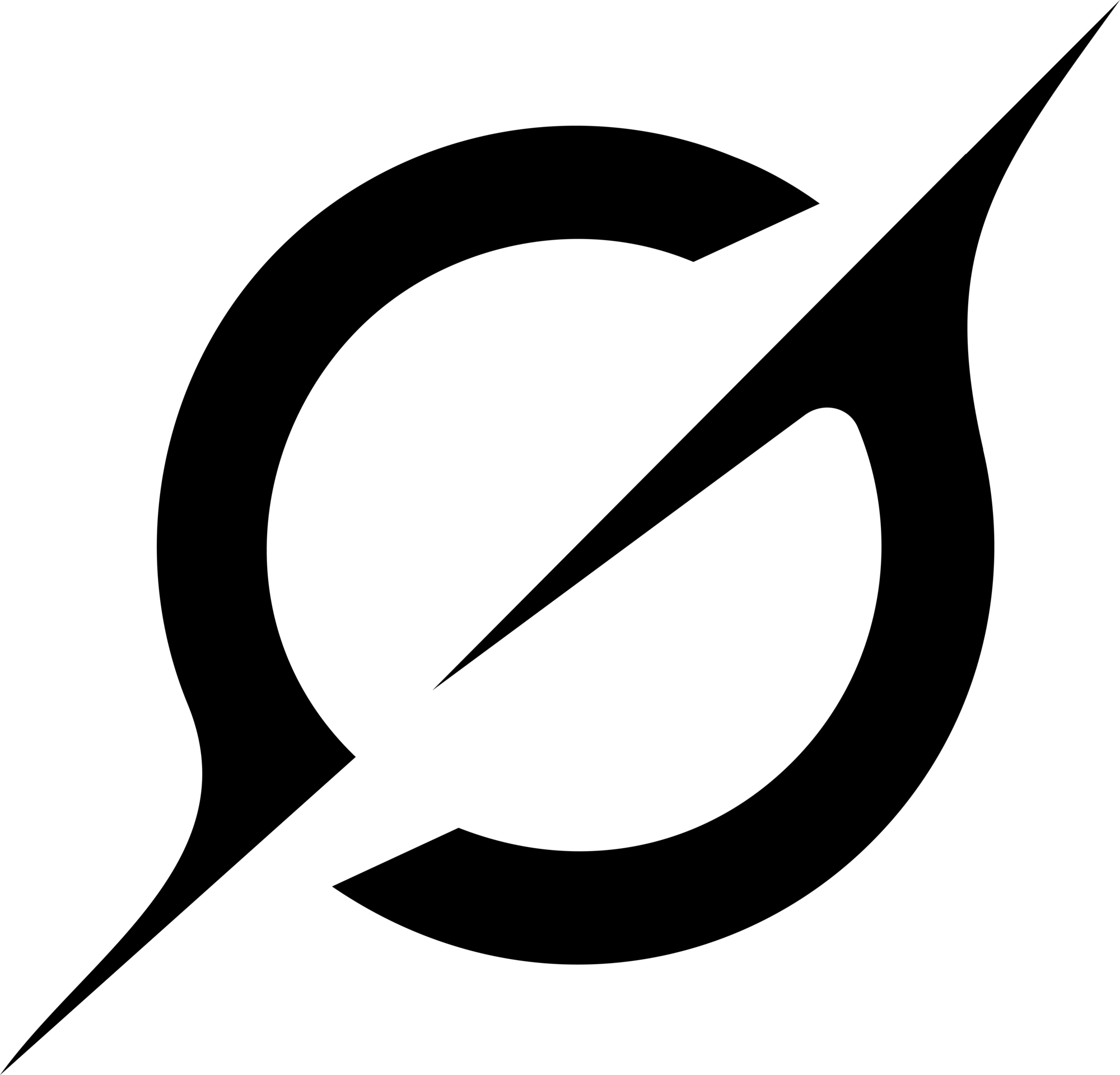}} Grok-4.1-Fast-Reasoning & xAI & 11/2025 & - & Single & 2M & False & 0.890 & 0.850 & 0.850 & 0.176 \\
\raisebox{-0.2em}{\includegraphics[height=1em]{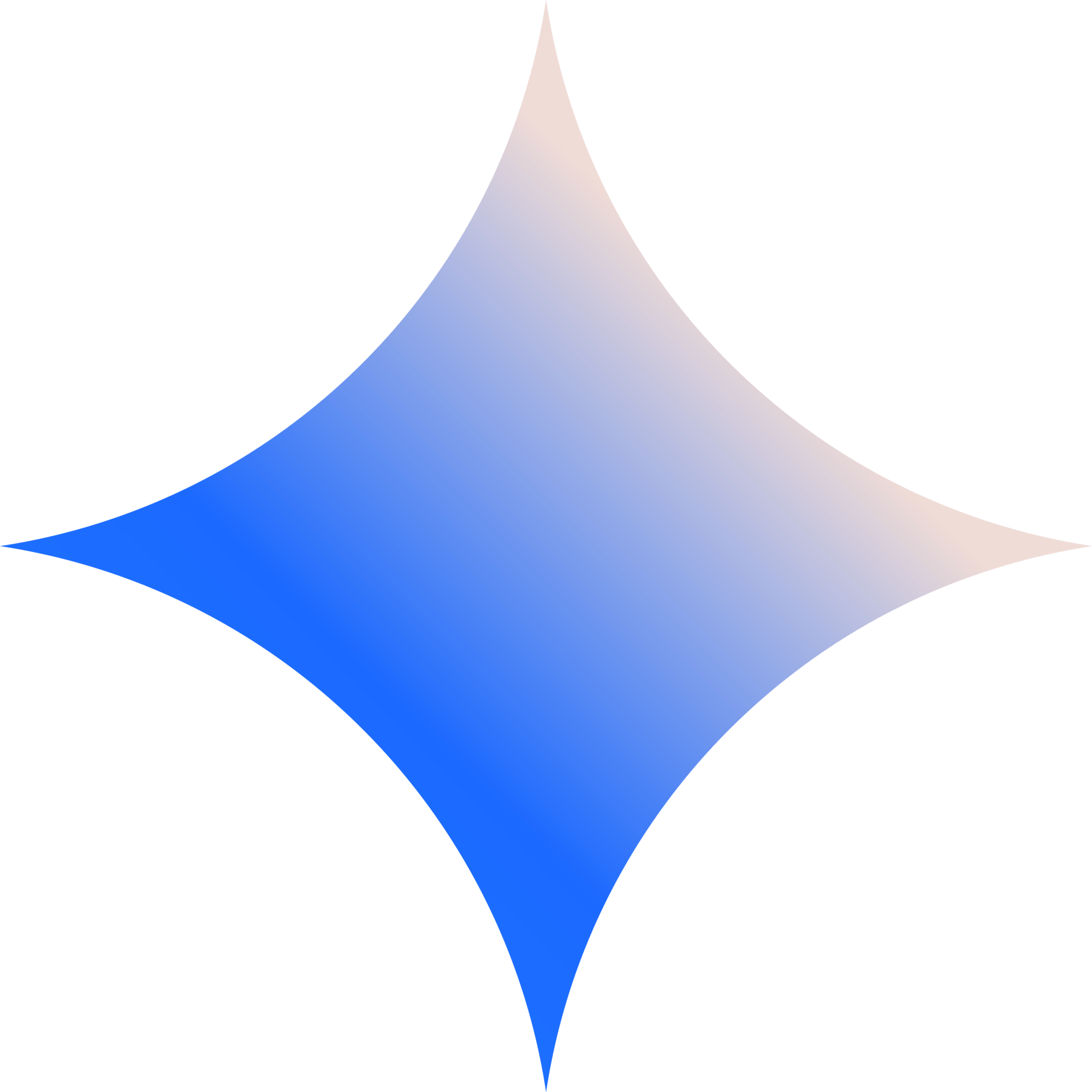}} Gemini-2.5-Pro & Google & 03/2025 & - & Multi & 1M & False & 0.880 & 0.840 & 0.860 & 0.211 \\
\bottomrule
\end{tabular}
\begin{flushleft}
The best result in each evaluation metric is highlighted.
\end{flushleft}
\label{tab_3}
\end{table*}

\subsection{Baselines}
This study selected two categories of mainstream baseline models for comparison: open-source models and closed-source models. These models encompass the current state-of-the-art open-source and closed-source deep reasoning LLMs, which demonstrate leading performance in complex mathematical problem reasoning tasks.

\textbf{Open-source models}. We evaluated the performance of multiple representative open-source LLMs on the IMPG task. These models have demonstrated strong capabilities in mathematical reasoning and complex problem solving, including Qwen3-32B\cite{yang2025qwen3}, DeepSeek-R1-Distill-Qwen-32B\cite{guo2025deepseek}, GLM-4.6, DeepSeek-V3.2-Speciale\cite{deepseekai2025deepseekv32}, Kimi-K2-Thinking\cite{team2025kimi}, and GPT-OSS-120B\cite{agarwal2025gpt}.

\textbf{Closed-source models}. We evaluated the performance of current state-of-the-art closed-source reasoning models on the IMPG task. These models typically possess larger parameter scales and stronger generalization capabilities, representing the top-tier standards of current general-purpose LLMs, including GPT-5.1 (high), Claude-Opus-4.5-Reasoning, Grok-4.1-Fast-Reasoning, and Gemini-2.5-Pro.

Detailed specifications of each baseline model are presented in \hyperref[tab_3]{Table 3}. We conducted evaluation tests across four mathematics-related or scientific reasoning benchmarks: AIME 2025, GPQA Diamond\cite{rein2024gpqa}, MMLU-Pro\cite{wang2024mmlu} and Humanity's Last Exam (HLE)\cite{phan2025humanity} to demonstrate their capabilities in complex reasoning tasks. Additionally, AIME 2025, OlymMATH\cite{sun2025challenging}, and MMLU-Pro-Math were used to quantify the changes in challenging mathematical reasoning after SFT training.

\subsection{Evaluation metrics}
According to the research by Zheng et al.\cite{zheng2023judging}, LLMs serving as judges demonstrate high consistency and can effectively substitute for human evaluation. Based on this finding and considerations of mathematical reasoning capability, we employed Gemini-3.0-Pro as the judge model and Gemini-2.5-Pro as the expert model. Some research\cite{li2025preference,wataoka2024self} indicates that when the LLM generating training data is associated with the LLM serving as judge, the judge tends to favor the related student model. However, the research noted that for models within the same family but different series, such as Gemini-3.0-Pro and Gemini-2.5-Pro, such bias is nearly negligible.

The evaluation methods employed in this study include the following types:

\textbf{G-eval}, generative model-based evaluation. Based on the aforementioned discussion, we employed Gemini-3.0-Pro as a judge to directly score the problems generated by the trained model, the system, and the baselines across nine dimensions, including Requirement, Correctness-P, Correctness-S, Fluency, Optimization, Coverage, Innovation, Computability, and Discrimination. The average score across all dimensions serves as the core indicator of overall problem quality, where Correctness-P, Correctness-S, and Innovation constitute the three core dimensions of the IMPG task. The system prompt for the judge is provided in \hyperref[appendix_A]{Appendix A}.

\textbf{Man-machine benchmarking} (MMB) evaluation. Problems generated by the problem generation system were compared against those developed by human experts to assess the relative strengths and weaknesses of the system compared to experts in the IMPG task. This essentially constitutes an A/B test comparing human and machine performance. The evaluation dimensions include correctness, fluency, coverage, innovation, and computability.

\begin{figure*}[!htbp]
  \centering
  \subfloat[9-Dimension Performance Analysis]{
      \includegraphics[width=0.48\textwidth]{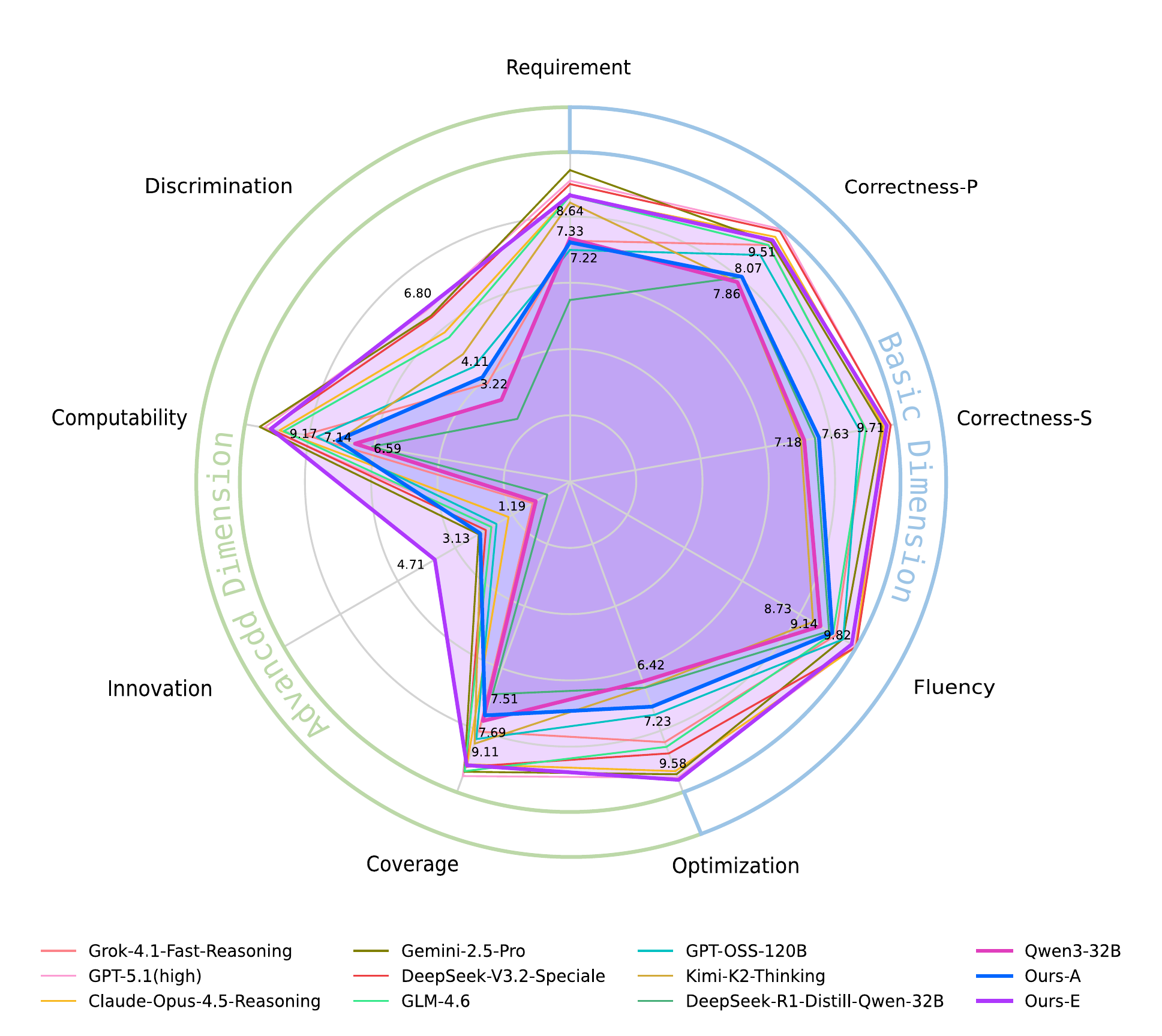}
      \label{fig_4_1}
  }
  \hfill
  \subfloat[Overall Performance Overview]{
      \includegraphics[width=0.48\textwidth]{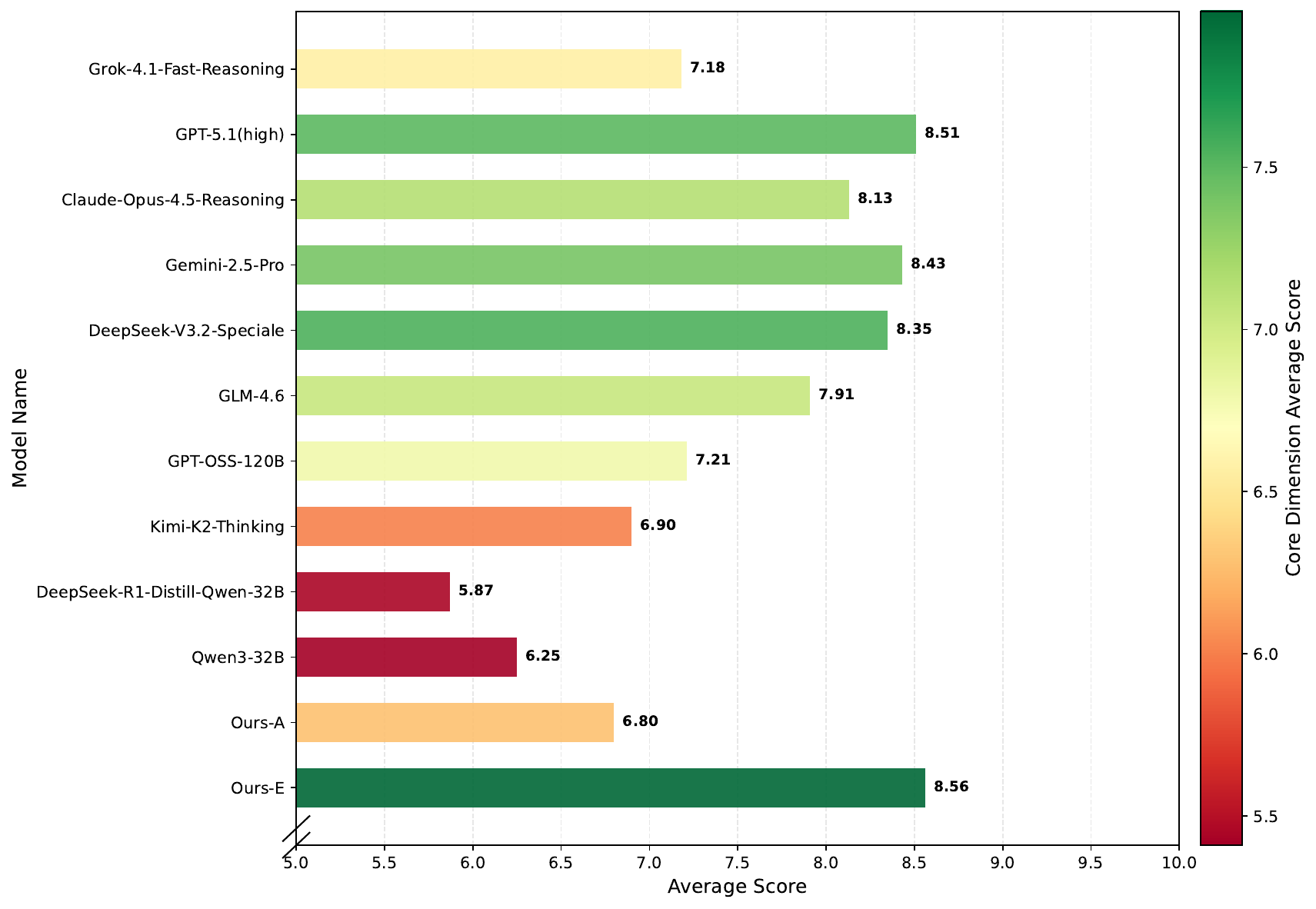}
      \label{fig_4_2}
  }
  
  \caption{The test results of all baseline models and our models, including apprentice mode and expert mode on the SIMU-90 dataset. (a) Performance of each model across nine evaluation dimensions. These dimensions encompass five basic dimensions and four advanced dimensions, providing a specific evaluation of the quality of generated problems. (b) Comparison of overall quality scores across models. The bar length represents the average score across nine dimensions, reflecting the overall quality of generated problems. The bar color indicates the average score of three core dimensions of the IMPG task, i.e., Correctness-P, Correctness-S, and Innovation.}
  \label{fig_4}
\end{figure*}

\textbf{Elo rating}, pairwise comparison evaluation based on the Arena framework. It was first proposed by Elo et al.\cite{elo1967proposed} in the 1960s. For each pairwise comparison between two competing models, we employed Gemini-3.0-Pro as a judge to determine which model generates better problems in terms of correctness, innovation, and overall quality given the same problem design requirements. The evaluation dimensions for overall quality are consistent with the nine dimensions of G-Eval. The core formula is as follows:

 1) Expected score:
\begin{equation}
E_A = \frac{1}{1 + 10^{(R_B - R_A)/400}}
\label{eqa_10}
\end{equation}
where $R_A$, $R_B$ are the current Elo ratings of model A and model B, respectively. 
$E_A$ is the expected win probability of model A against model B.

 2) Rating update:
\begin{equation}
R'_A = R_A + K \cdot (S_A - E_A)
\label{eqa_11}
\end{equation}
where $S_A$ is the actual result, i.e., win = 1, draw = 0.5, loss = 0. $K$ is the update coefficient used to control the magnitude of the update. $R'_A$ is the updated rating.

During the competition, in addition to calculating the final Elo rating, we also aggregated the win rate for both competing models.

\textbf{Difficulty accuracy} evaluation measures the degree of alignment between the difficulty of the generated problems and the difficulty level specified in the problem design requirements. 

Accuracy measures the proportion of samples whose estimated difficulty level exactly matches the ground-truth difficulty level. It is defined as:

\begin{equation}
\text{Acc} = \frac{1}{N}\sum_{i=1}^{N}\mathbb{I}(\hat{y}_i = y_i)
\label{eq:accuracy}
\end{equation}

\noindent where $N$ denotes the total number of samples, $y_i$ represents the ground-truth difficulty level of the $i$-th sample, $\hat{y}_i$ denotes the estimated difficulty level, and $\mathbb{I}(\cdot)$ is the indicator function that returns 1 if the condition is satisfied and 0 otherwise.

\textbf{Mean absolute deviation} (MAD) quantifies the average magnitude of estimation errors. The MAD is then computed as:

\begin{equation}
\text{MAD} = \frac{1}{N}\sum_{i=1}^{N}|\hat{y}_i - y_i|
\label{eq:mad}
\end{equation}

\noindent where $y_i, \hat{y}_i \in \{1, 2, 3, 4\}$ represent the ground-truth and estimated metrics, respectively. The advantage of MAD lies in its ability to reflect the severity of errors.

\textbf{Shannon entropy} evaluates the diversity of sampled difficulty encodings. The Shannon entropy is computed as:
\begin{equation}
H = -\sum_{i=1}^{n} p_i \log_2(p_i)
\label{eq:shannon}
\end{equation}
\noindent where $p_i$ represents the probability of the $i$-th encoding, and $n$ denotes the total number of distinct encoding categories. The logarithm is typically taken with base 2, yielding the entropy in units of \textit{bits}.

\begin{table*}[!htbp]
\centering
\footnotesize
\caption{The detailed experimental results of the baseline models and our proposed models, including performance scores across nine dimensions, overall average scores, and overall average scores on core dimensions.}
\setlength{\tabcolsep}{3.5pt}
\renewcommand{\arraystretch}{1.2}
\begin{tabular}{lccccccccccc}
\toprule[0.5pt] 
Model & Req. & Corr-P & Corr-S & Flu. & Opt. & Cov. & Innov. & Comp. & Disc. & Avg. & Core Avg.\\
\midrule[0.5pt] 
\multicolumn{11}{c}{\hspace{8em}\textit{open-source models}} \\
Qwen3-32B & 7.33 & 7.86 & 7.18 & 8.73 & 6.42 & 7.69 & 1.19 & 6.59 & 3.22 & 6.25 & 5.41 \\
DeepSeek-R1-Distill-Qwen-32B & 5.48 & 8.06 & 7.51 & 9.03 & 6.62 & 6.83 & 0.79 & 6.07 & 2.47 & 5.87 & 5.45\\
GLM-4.6 & 8.62 & 9.32 & 9.07 & 9.17 & 8.52 & 9.30 & 2.74 & 8.77 & 5.68 & 7.91 & 7.04\\
DeepSeek-V3.2-Speciale & 8.98 & 9.86 & \colorbox{black!20}{\makebox[3em][c]{9.83}} & \colorbox{black!20}{\makebox[3em][c]{9.98}} & 8.73 & 9.16 & 2.93 & 9.20 & 6.48 & 8.35 & 7.54\\
Kimi-K2-Thinking & 8.42 & 7.86 & 7.08 & 8.47 & 6.61 & 8.42 & 3.10 & 7.09 & 5.02 & 6.90 & 6.01\\
GPT-OSS-120B & 6.99 & 8.93 & 8.88 & 9.52 & 7.49 & 8.27 & 2.56 & 7.76 & 4.52 & 7.21 & 6.79\\
\midrule[0.5pt] 
\multicolumn{11}{c}{\hspace{8em}\textit{closed-source models}} \\
GPT-5.1 (high) & 9.08 & \colorbox{black!20}{\makebox[3em][c]{9.94}} & 9.81 & 9.93 & 9.51 & \colorbox{black!20}{\makebox[3em][c]{9.46}} & 2.71 & 9.37 & 6.74 & 8.51 & 7.49\\
Claude-Opus-4.5-Reasoning & 8.63 & 9.64 & 9.61 & 9.97 & 9.30 & 9.08 & 2.14 & 8.90 & 5.88 & 8.13 & 7.13\\
Grok-4.1-Fast-Reasoning & 7.27 & 9.32 & 9.08 & 9.27 & 8.37 & 8.03 & 1.30 & 8.13 & 3.87 & 7.18 & 6.57\\
Gemini-2.5-Pro & \colorbox{black!20}{\makebox[3em][c]{9.40}} & 9.43 & 9.56 & 9.51 & 9.40 & 9.32 & 3.18 & \colorbox{black!20}{\makebox[3em][c]{9.51}} & 6.53 & 8.43 & 7.39\\
\midrule[0.5pt] 
Ours-A & 7.22 & 8.07 & 7.63 & 9.14 & 7.23 & 7.51 & 3.13 & 7.14 & 4.11 & 6.80 & 6.28\\
Ours-E & 8.64 & 9.51 & 9.71 & 9.82 & \colorbox{black!20}{\makebox[3em][c]{9.58}} & 9.11 & \colorbox{black!20}{\makebox[3em][c]{4.71}} & 9.17 & \colorbox{black!20}{\makebox[3em][c]{6.80}} & \colorbox{black!20}{\makebox[3em][c]{8.56}} & \colorbox{black!20}{\makebox[3em][c]{7.98}}\\
\bottomrule[0.5pt] 
\end{tabular}
\begin{flushleft}
The best result in each metric is highlighted.
\end{flushleft}
\label{tab_4}
\end{table*}

\textbf{Originality} evaluation compares the degree of overlap between the generated problems and existing problems in the problem bank. We employed BLEU-1/2/3/4\cite{papineni2002bleu} and ROUGE-1/2/L~\cite{lin2004rouge} for lexical similarity evaluation, and the semantic embedding model BGE-M3\cite{chen2024m3} for semantic similarity measurement.

\subsection{Training details}
We use Qwen3-32B as our base model. All stages employed LoRA for parameter-efficient fine-tuning with BF16 precision. In the CPT stage, LoRA rank was set to 16 with alpha of 32 and a learning rate of 2e-5, using DeepSpeed ZeRO-3. Training was conducted for 1 epoch on algebraic and graph knowledge, and 2 epochs on key knowledge. In the SFT stage, LoRA rank was set to 64 with alpha of 128, a learning rate of 3e-5, training for 2 epochs using DeepSpeed ZeRO-3. In the GRPO stage, LoRA configuration remained identical to SFT. The learning rate was 2e-6, training for 1 epoch with a generation count of 4 using the DeepSpeed ZeRO-2. During the inference process, the model was deployed using the vLLM framework. Parameter settings include temperature of 0.2, top\_p of 0.7, top\_k of 20. Training was conducted on two NVIDIA RTX Pro 6000 96G GPUs, consuming 228 GPU hours in total, while inference used a single GPU.

\section{Results and discussion}
\subsection{Problem generation capability evaluation}
\subsubsection{G-eval scoring}
We conducted comprehensive evaluations of all baseline models and models built upon the proposed framework on the test set SIMU-90. Detailed experimental results are presented in \hyperref[tab_4]{Table 4}.

As shown in \hyperref[fig_4_1]{Fig. 4(a)}, our model in apprentice mode (Ours-A), which employed the fine-tuned Qwen3-32B as both the generator and evaluator, outperformed the base model Qwen3-32B across nearly all dimensions. Notably, Ours-A achieved an innovation score of 3.13 compared to 1.19 for the base model, representing a 163.03\% improvement. Our model in expert mode (Ours-E) used the fine-tuned Qwen3-32B as the generator and incorporated the external Gemini-2.5-Pro as the expert model for evaluation guidance. Compared to all baselines, Ours-E attained the highest scores in three of the nine dimensions, namely optimization at 9.58, innovation at 4.71, and discrimination at 6.80. In particular, the Correctness-P and Correctness-S scores rank in the forefront among all baselines. As illustrated in \hyperref[fig_4_2]{Fig. 4(b)}, Ours-A achieved superior overall performance across nine dimensions compared to models of equivalent parameter size, and even surpassed Kimi-K2-Thinking with 1T parameters in core dimension average, reaching 6.28 versus 6.01. Ours-E achieved state-of-the-art performance in terms of both the overall metrics across nine dimensions and the three core dimensions.

These improvements are primarily attributed to the generator equipped with the difficulty model, which generates highly innovative problems while ensuring the correctness under the guidance of the expert model.

\begin{figure*}[!htbp]
\centering
\includegraphics[width=\textwidth]{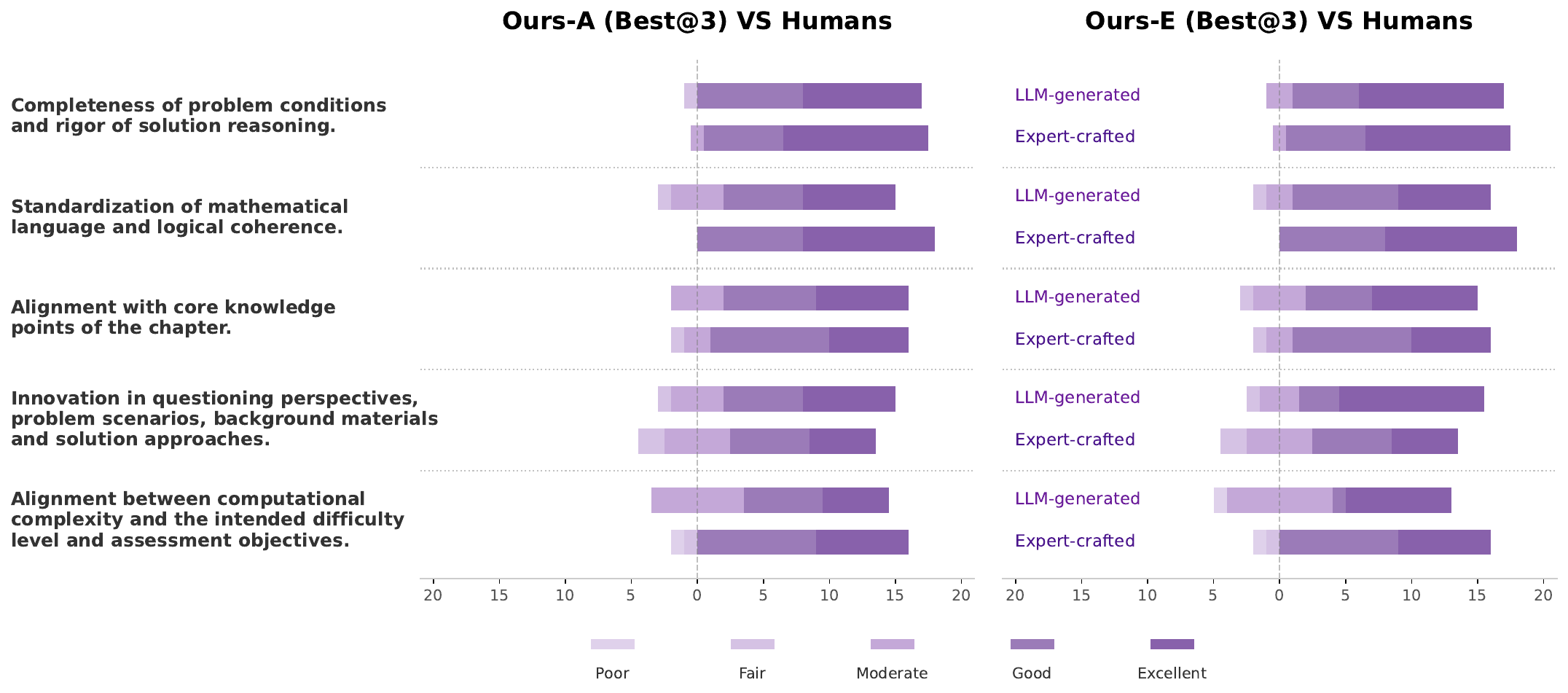}
\caption{Human preference evaluation of problems generated by human experts and our model (Best@3), including apprentice mode and expert mode across each dimension. Preferences are classified into five levels, with the length of each bar segment indicating the number of evaluators for each preference level.}
\label{fig_5}
\end{figure*}

\subsubsection{MMB test}
To investigate human preferences between problems designed by human experts and those generated by our proposed models, we designed a man-machine benchmark experiment.

For the derivative chapter, experienced problem designers and the proposed models were each tasked with designing a complete test paper. Each paper comprised nine problems across three types covering multiple difficulty levels. The requirements for problem design were identical across both sets. To simulate authentic paper design scenarios and maximize model performance, we adopted a $Best@3$ strategy, allowing the LLMs to generate three candidate items for each requirement.

We recruited 18 students with excellent mathematics scores to evaluate each paper item and its corresponding solution across five dimensions: correctness, fluency, knowledge coverage, innovation, and computational consistency. All items were anonymized during the evaluation process. Evaluators rated their preferences on a five-point Likert scale. The region to the right of the origin was defined as the \textbf{advantage zone}, representing positive human preference, while the region to the left was defined as the \textbf{disadvantage zone}, representing negative human preference. The experimental results are presented in \hyperref[fig_5]{Fig. 5}. Both sources exhibited advantage zones substantially larger than disadvantage zones, indicating generally high quality. Minimal differences were observed in the correctness and fluency dimensions. In knowledge coverage and computational consistency, the proposed models' advantage zones were slightly smaller than those of human experts. However, in the innovation dimension, the proposed models' advantage zones exceeded those of human experts. These findings demonstrate that our framework enhances innovation while maintaining correctness and fluency comparable to human experts.

\subsubsection{Offline Elo rating}
To comprehensively evaluate the performance differences and stability of models across the three core dimensions of the IMPG task, we conducted 1,000 rounds of pairwise comparison experiments. In each round, two models were anonymously and randomly selected, and the outcomes were categorized into win, loss, or tie. To ensure fairness in the competition, we categorized all models into an open-source group and a closed-source group, placing Ours-A in the open-source group and Ours-E in the closed-source group. Gemini-3.0-Pro served as the judge to evaluate models on the three core dimensions.

\begin{figure*}[htbp]
  \centering
  \hspace{8mm}\subfloat[Open-source models Correctness-P\label{fig6_sub1}]{\hspace{-8mm}
    \includegraphics[width=0.31\textwidth]{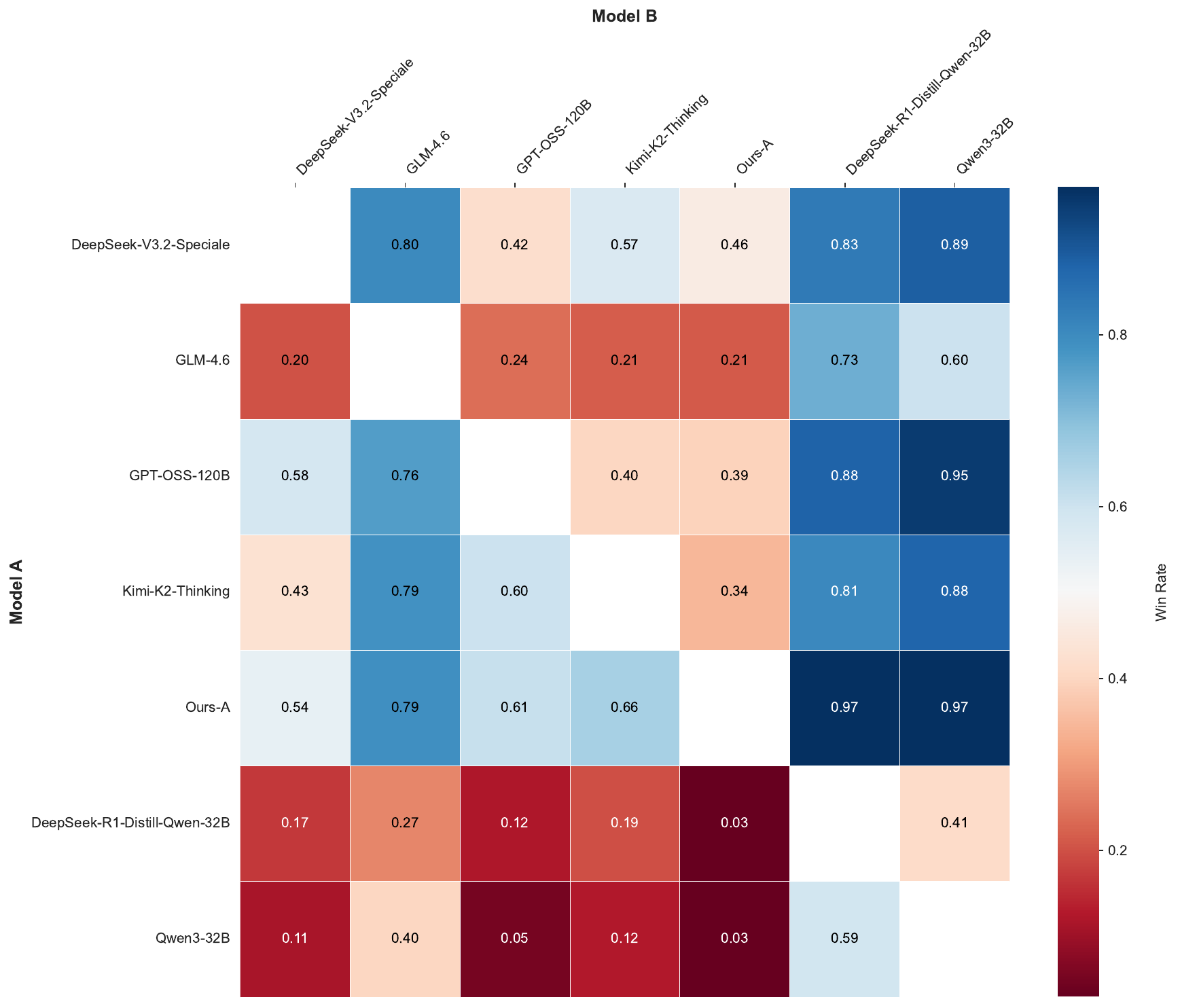}
  }
  \hfill
  \hspace{8mm}\subfloat[Open-source models Correctness-S\label{fig6_sub2}]{\hspace{-8mm}
    \includegraphics[width=0.31\textwidth]{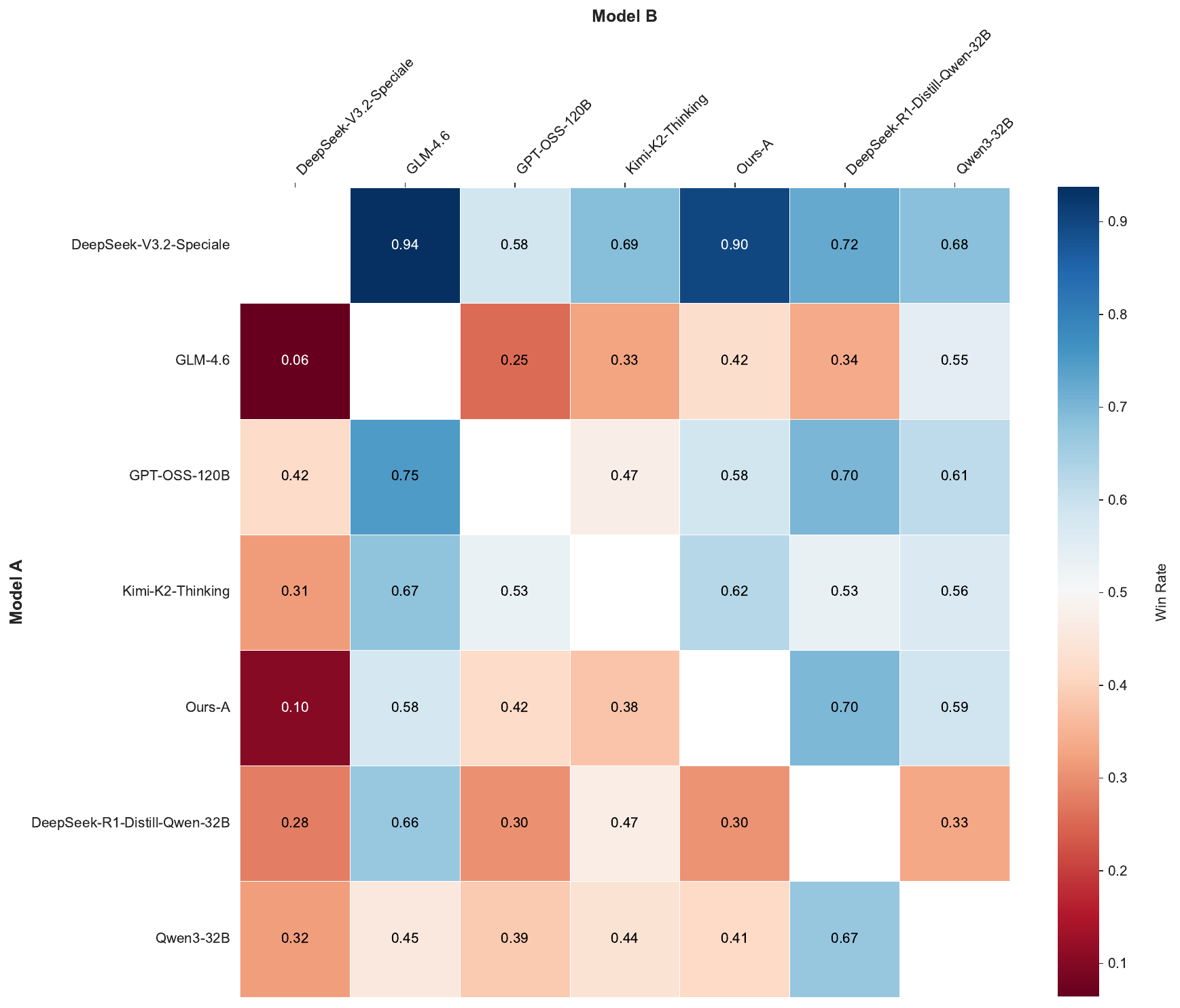}
  }
  \hfill
  \hspace{8mm}\subfloat[Open-source models Innovation\label{fig6_sub3}]{\hspace{-8mm}
    \includegraphics[width=0.31\textwidth]{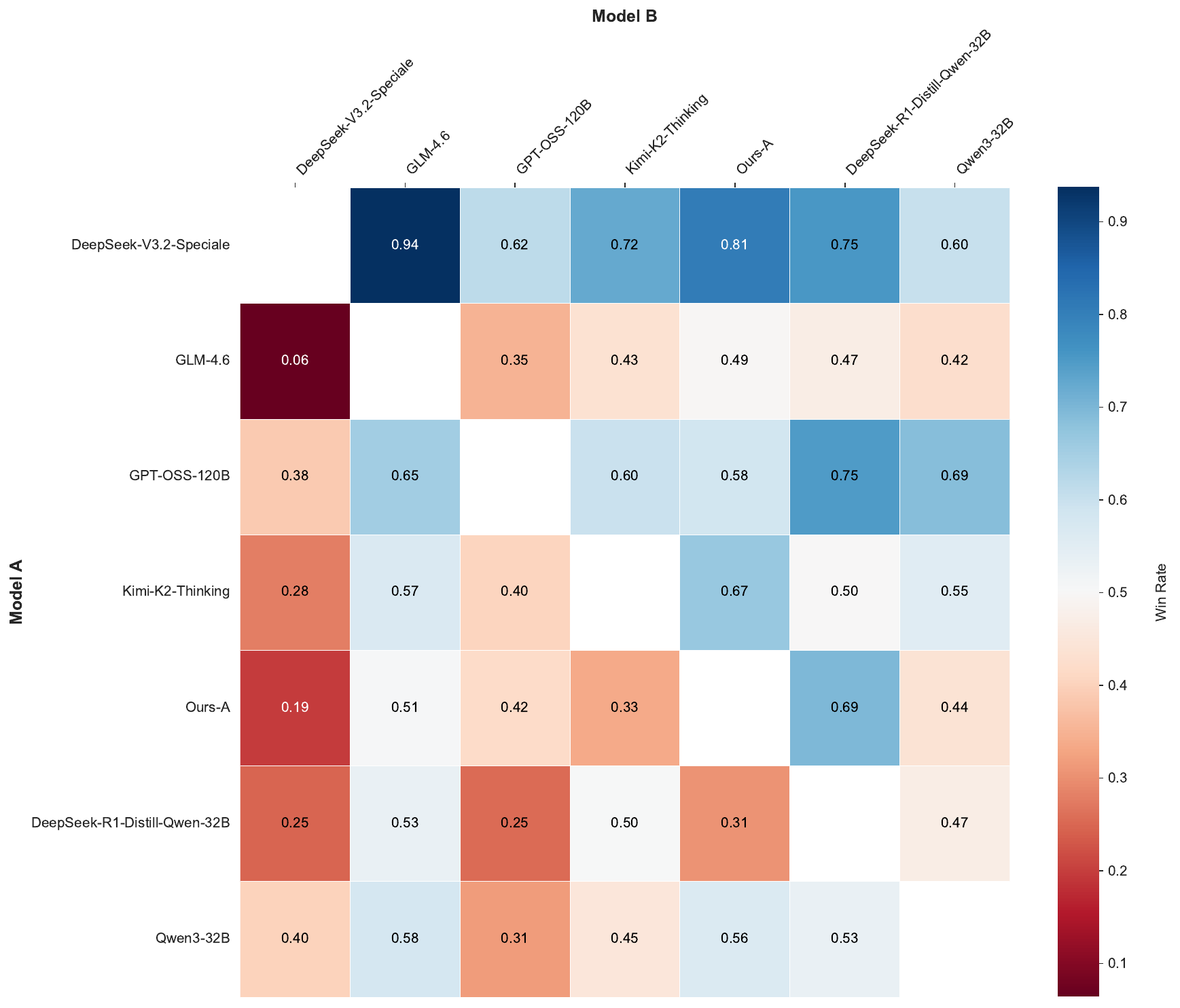}
  }
  
  \vspace{0.2cm}
  
  \hspace{8mm}\subfloat[Closed-source models Correctness-P\label{fig6_sub4}]{\hspace{-8mm}
    \includegraphics[width=0.31\textwidth]{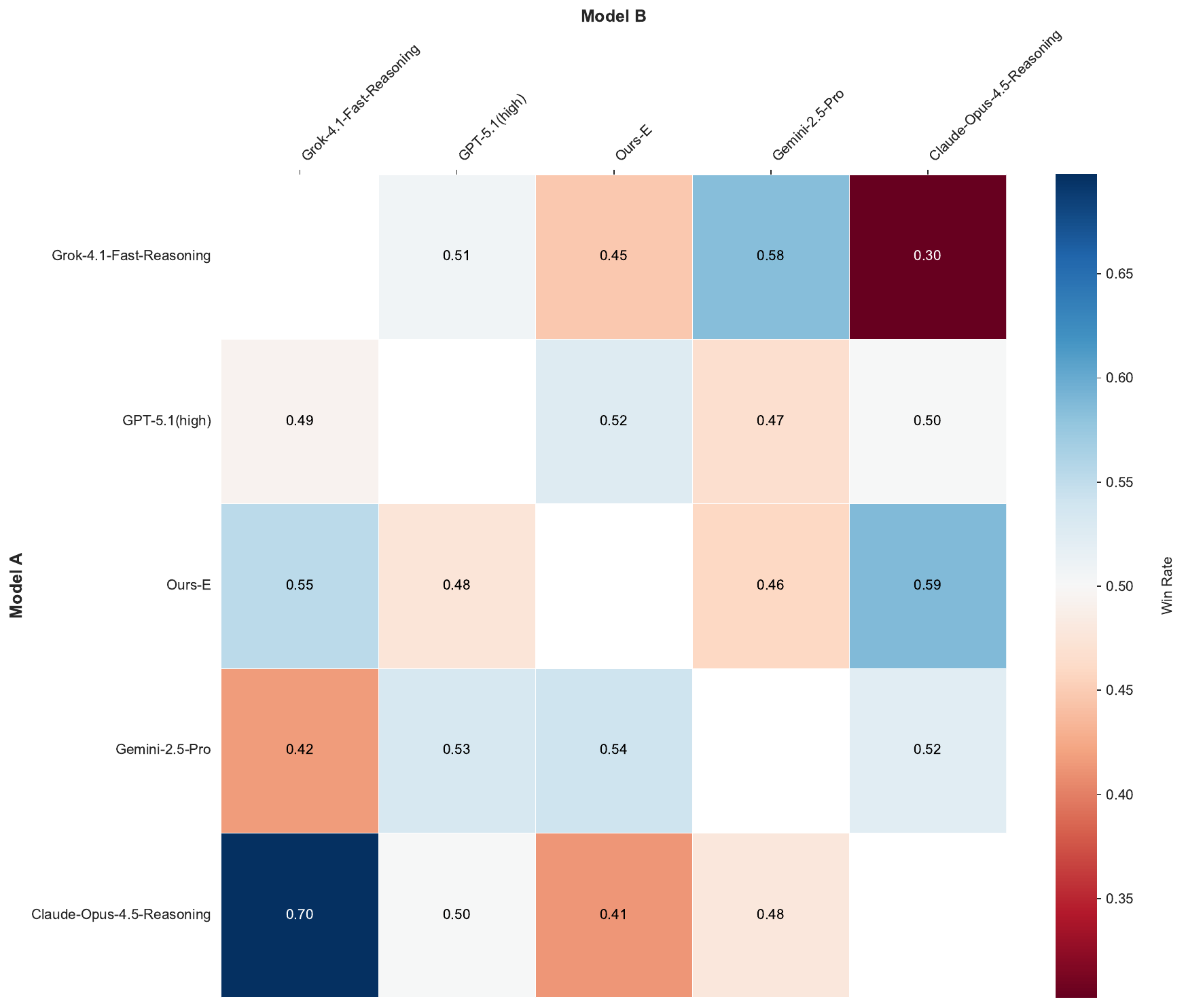}
  }
  \hfill
  \hspace{8mm}\subfloat[Closed-source models Correctness-S\label{fig6_sub5}]{\hspace{-8mm}
    \includegraphics[width=0.31\textwidth]{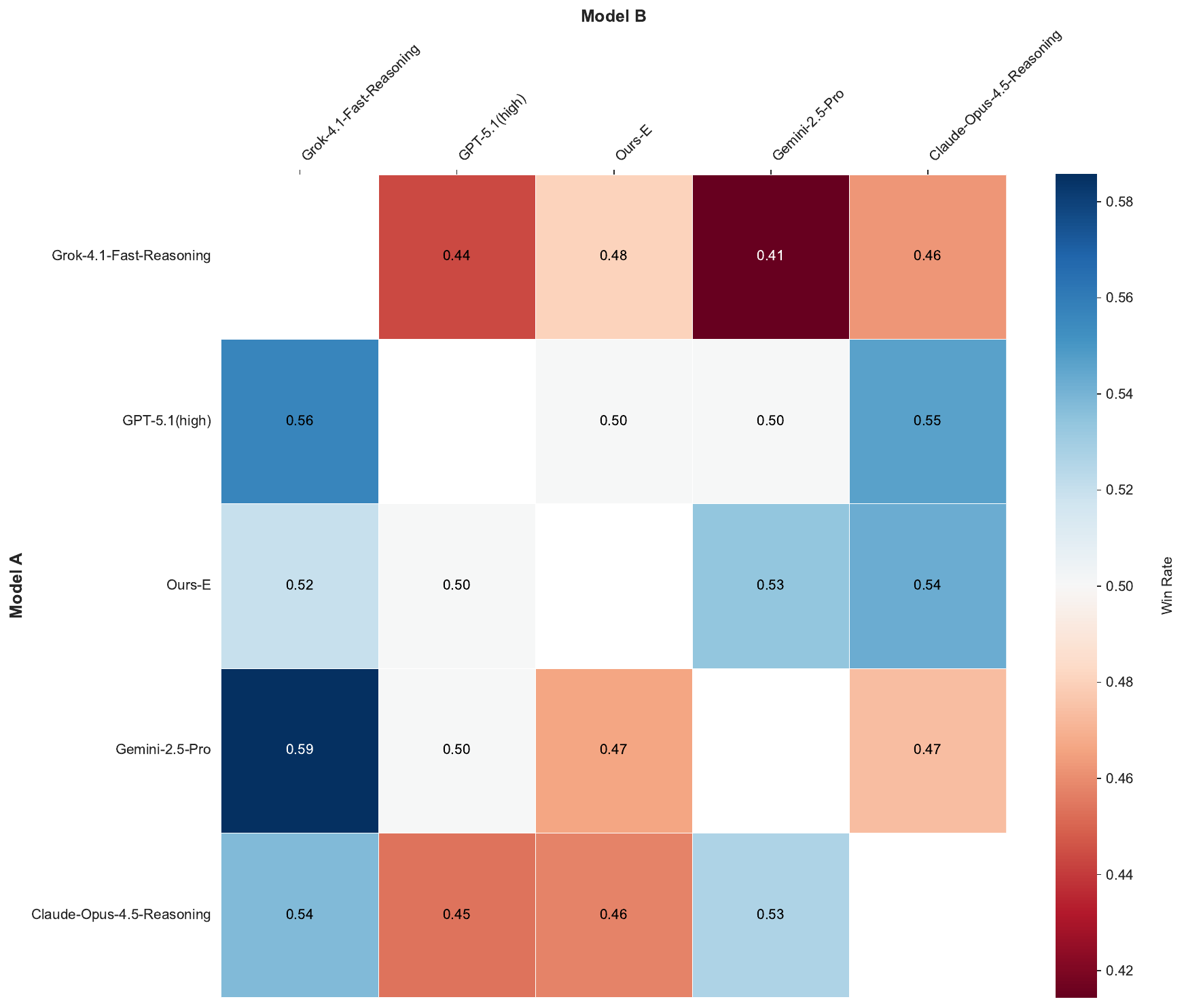}
  }
  \hfill
  \hspace{8mm}\subfloat[Closed-source models Innovation\label{fig6_sub6}]{\hspace{-8mm}
    \includegraphics[width=0.31\textwidth]{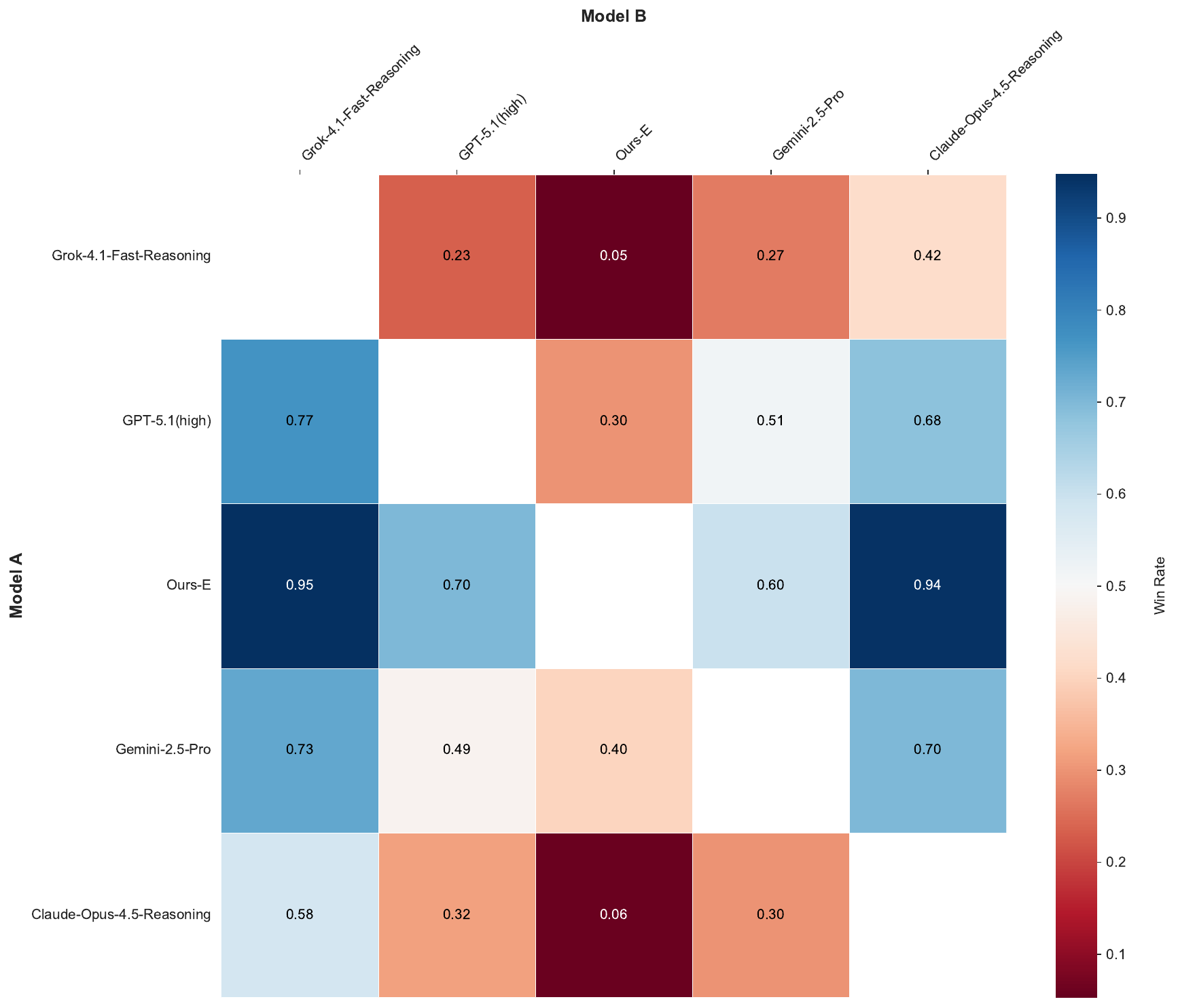}
  }
  
  \caption{Model A win rate vs. model B with ties split evenly}
  \label{fig_6}
\end{figure*}

\begin{figure*}[htbp]
  \centering
  \hspace{8mm}\subfloat[Open-source models Correctness-P\label{fig7_sub1}]{\hspace{-8mm}
    \includegraphics[width=0.31\textwidth]{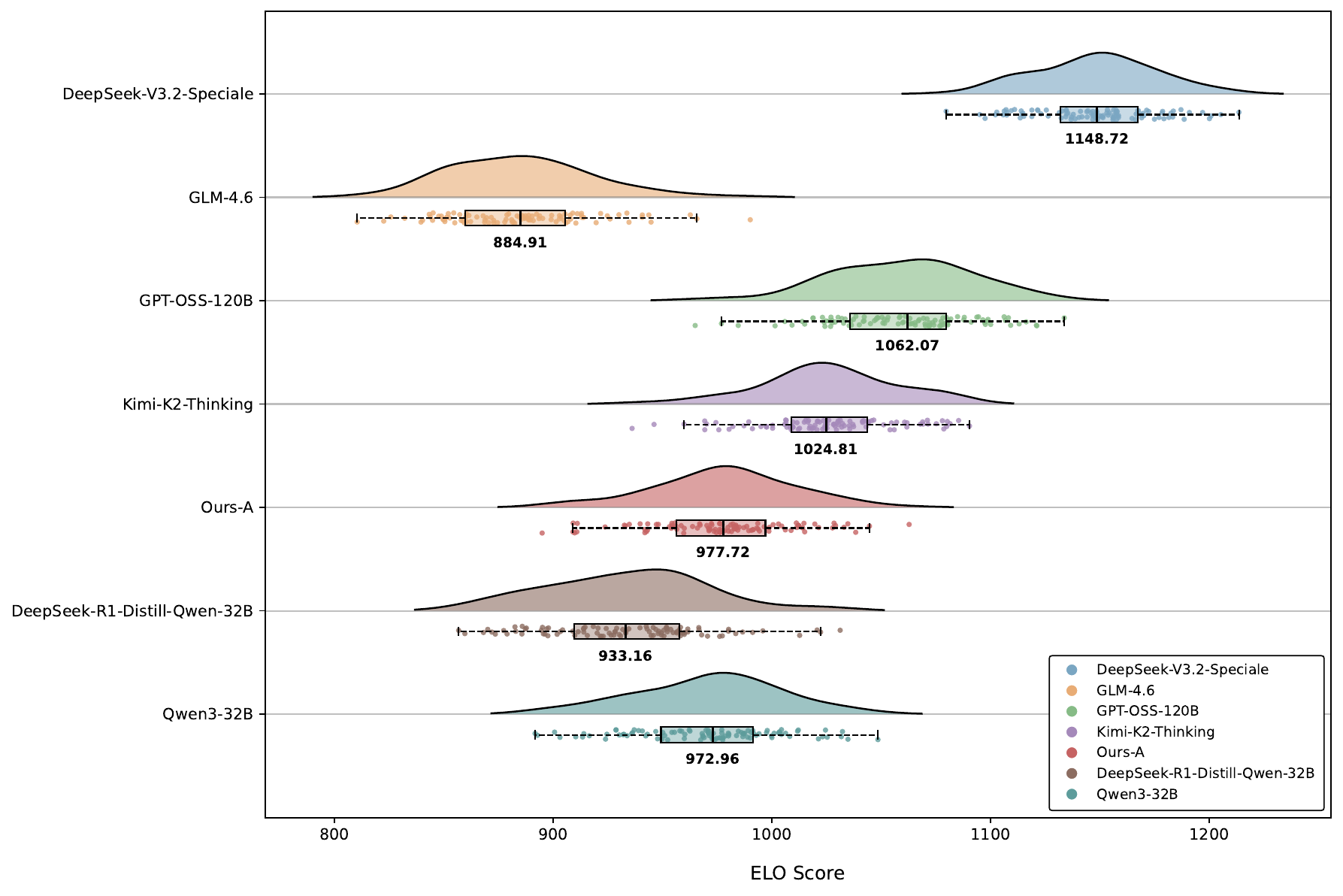}
  }
  \hfill
  \hspace{8mm}\subfloat[Open-source models Correctness-S\label{fig7_sub2}]{\hspace{-8mm}
    \includegraphics[width=0.31\textwidth]{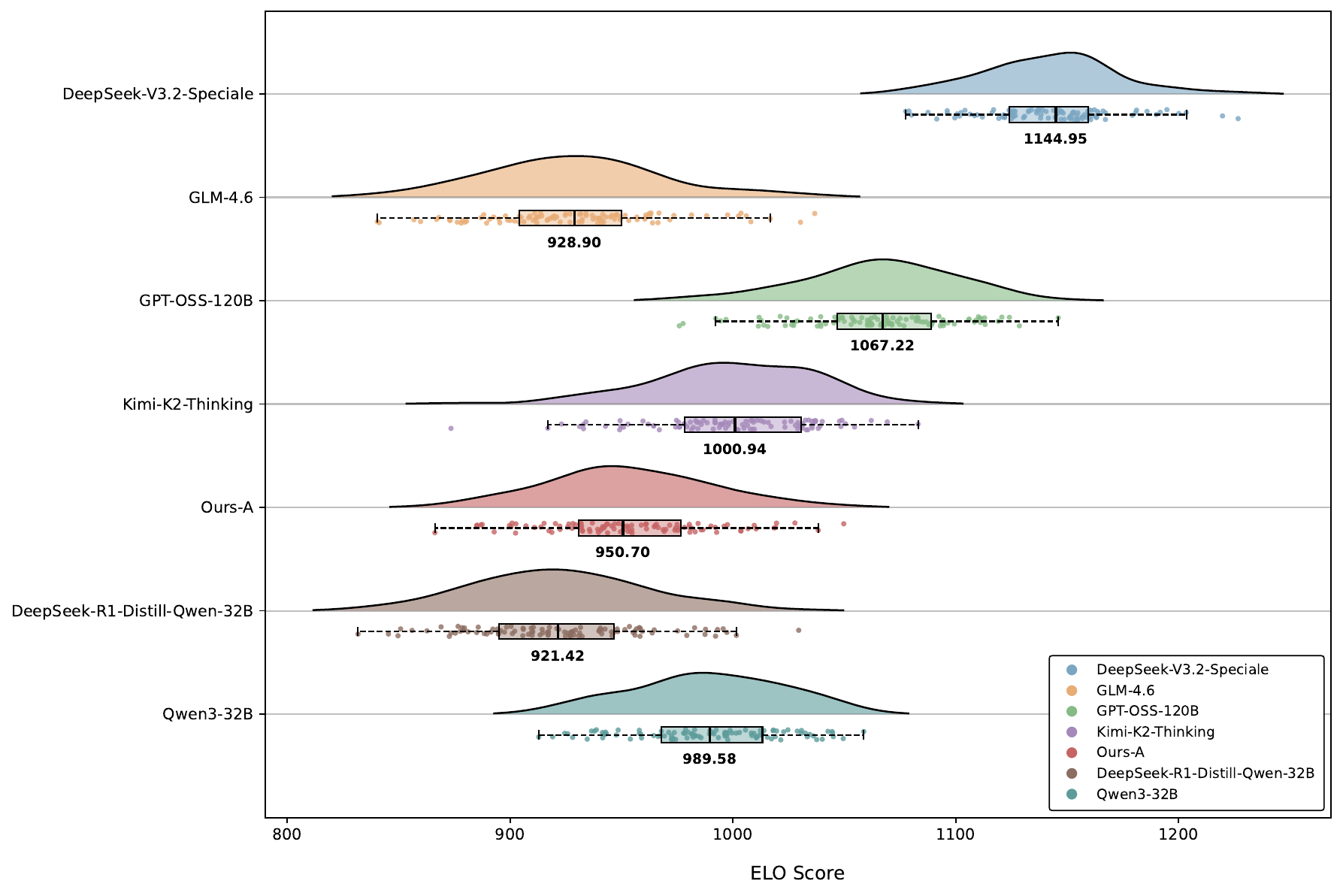}
  }
  \hfill
  \hspace{8mm}\subfloat[Open-source models Innovation\label{fig7_sub3}]{\hspace{-8mm}
    \includegraphics[width=0.31\textwidth]{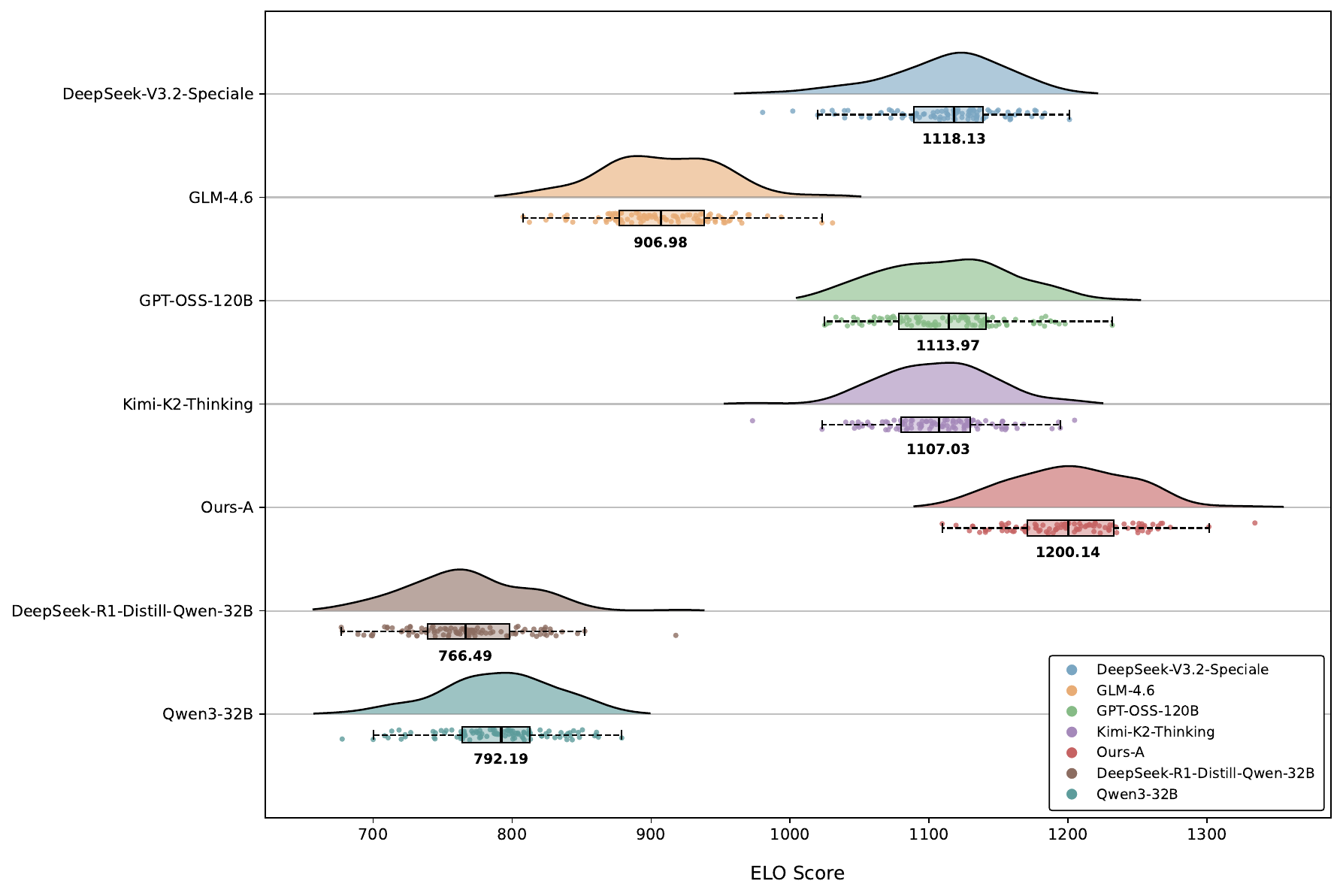}
  }
  
  \vspace{0.2cm}
  
  \hspace{8mm}\subfloat[Closed-source models Correctness-P\label{fig7_sub4}]{\hspace{-8mm}
    \includegraphics[width=0.31\textwidth]{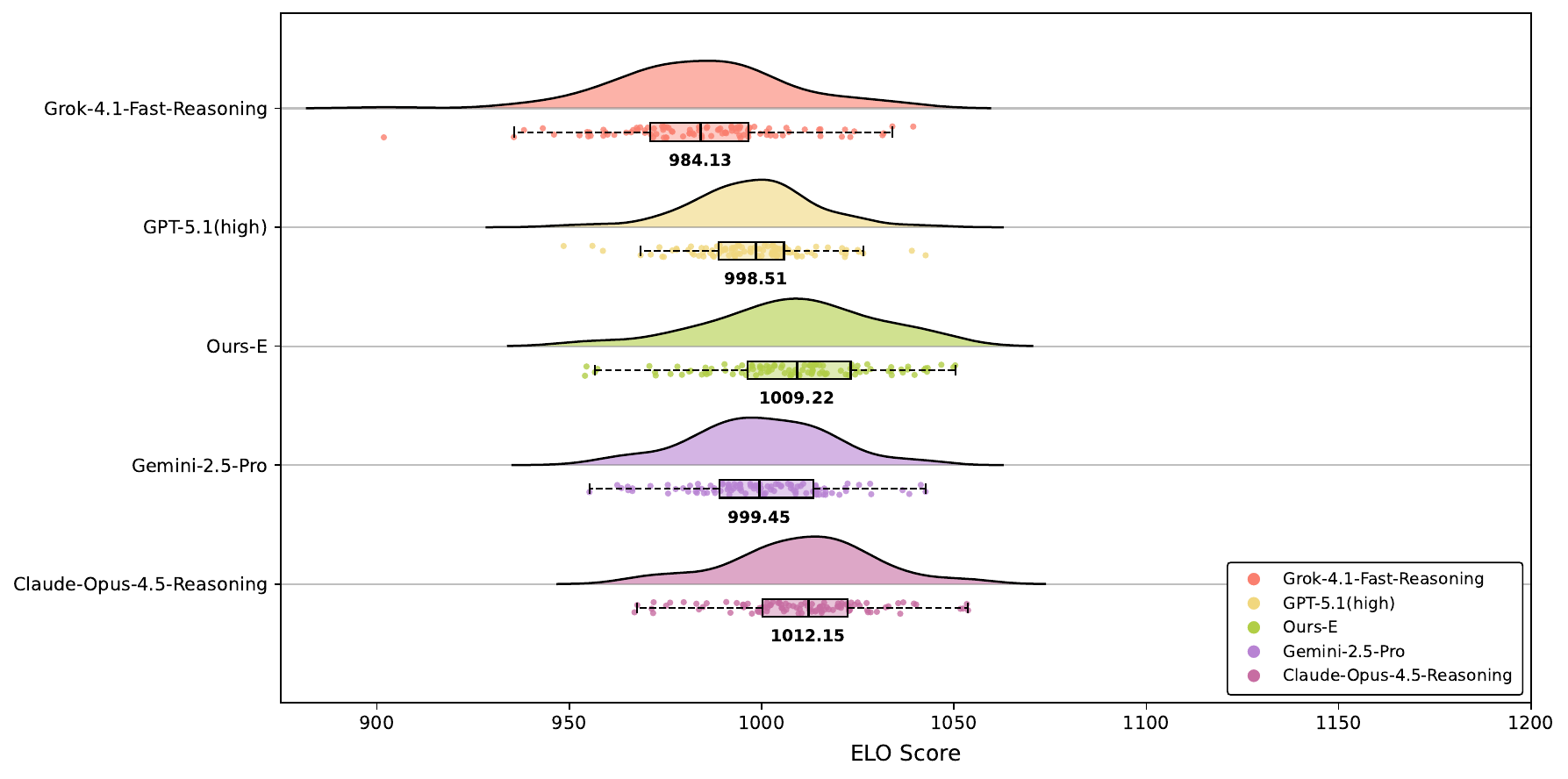}
  }
  \hfill
  \hspace{8mm}\subfloat[Closed-source models Correctness-S\label{fig7_sub5}]{\hspace{-8mm}
    \includegraphics[width=0.31\textwidth]{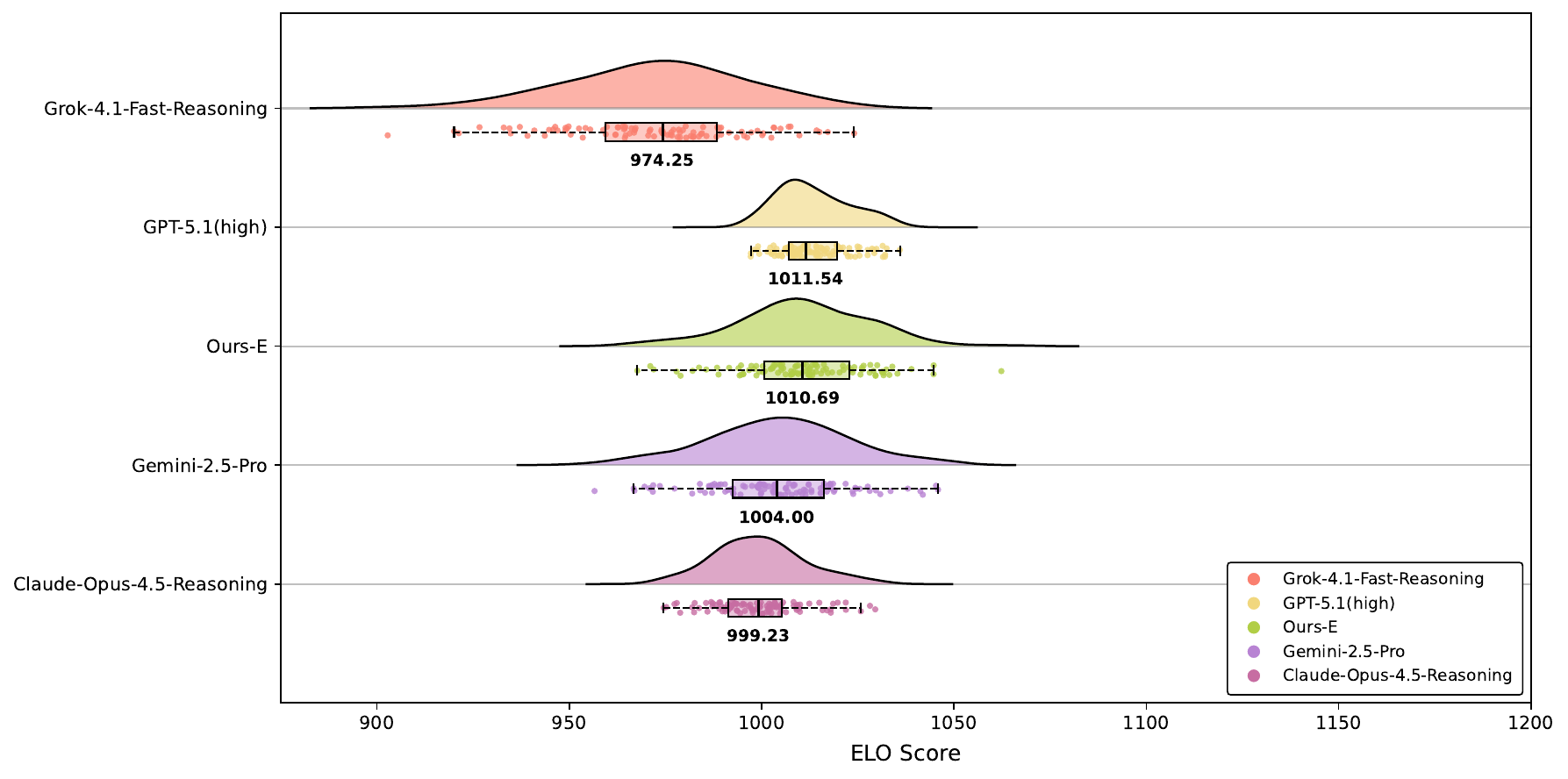}
  }
  \hfill
  \hspace{8mm}\subfloat[Closed-source models Innovation\label{fig7_sub6}]{\hspace{-8mm}
    \includegraphics[width=0.31\textwidth]{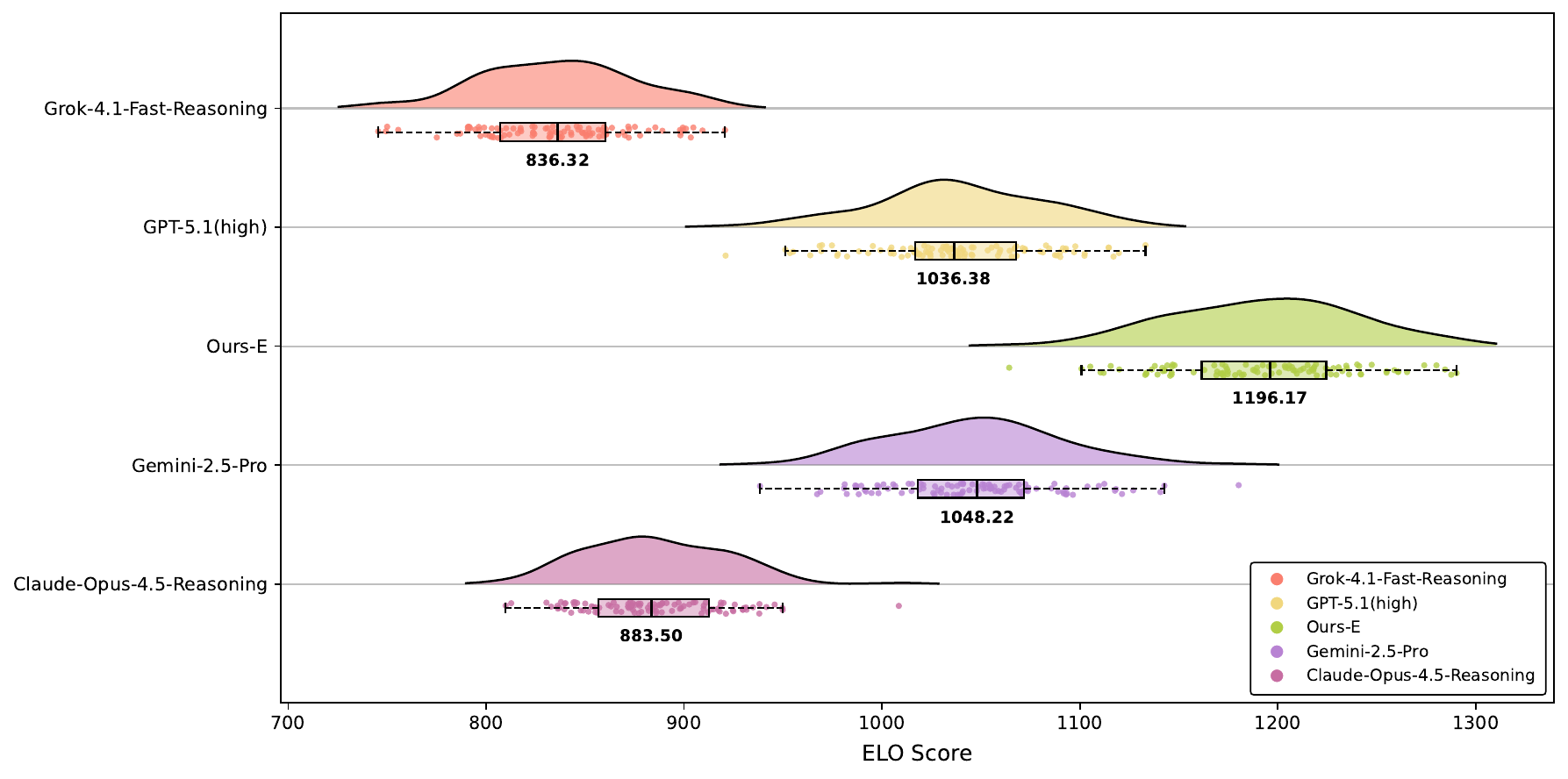}
  }
  
  \caption{Distribution of Elo scores for open-source and closed-source models via bootstrapping across three core dimensions, i.e., Correctness-P, Correctness-S, and Innovation. The cloud illustrates the distribution shape and probability density of the data, the rain represents the actual sample data points, and the box plot displays key summary statistics, including the median, quartiles, and the normal range of data variation.}
  \label{fig_7}
\end{figure*}

To eliminate position bias\cite{zheng2023judging} where LLMs tend to prefer the preceding answer, we adopted a position swapping strategy. The input order of the same pair of competing models was swapped for a secondary evaluation. Matches with inconsistent conclusions before and after swapping were repeated. We utilized the update rules defined in \hyperref[eqa_10]{Eq. (10)} and \hyperref[eqa_11]{Eq. (11)} and performed bootstrap resampling with replacement 100 times based on the 1,000 rounds of confrontation records. The finally calculated pairwise model win rate matrix is presented in \hyperref[fig_6]{Fig. 6}. Furthermore, \hyperref[fig_7]{Fig. 7} illustrates the Elo score distributions via cloud and rain plots.

The experimental results demonstrate that Ours-A performed remarkably in the open-source group. On Correctness-P and Correctness-S, Ours-A achieved Elo scores of 977.72 and 950.70, respectively, surpassing not only models of the same parameter scale but also GLM-4.6, which possesses 357B parameters. Most notably, Ours-A achieved the highest Innovation Elo score of 1200.14 among all open-source models. In the closed-source group, Ours-E exhibited strong competitiveness across all dimensions. On Correctness-P and Correctness-S, Ours-E attained Elo scores of 1009.22 and 1010.69, ranking second only to Claude-Opus-4.5-Reasoning and GPT-5.1 (high), respectively. Most significantly, Ours-E dominated the Innovation dimension with an Elo of 1196.17, substantially outperforming all closed-source models.

\subsection{Ablation study}
\begin{figure*}[htbp]
\centering
\includegraphics[width=0.65\textwidth]{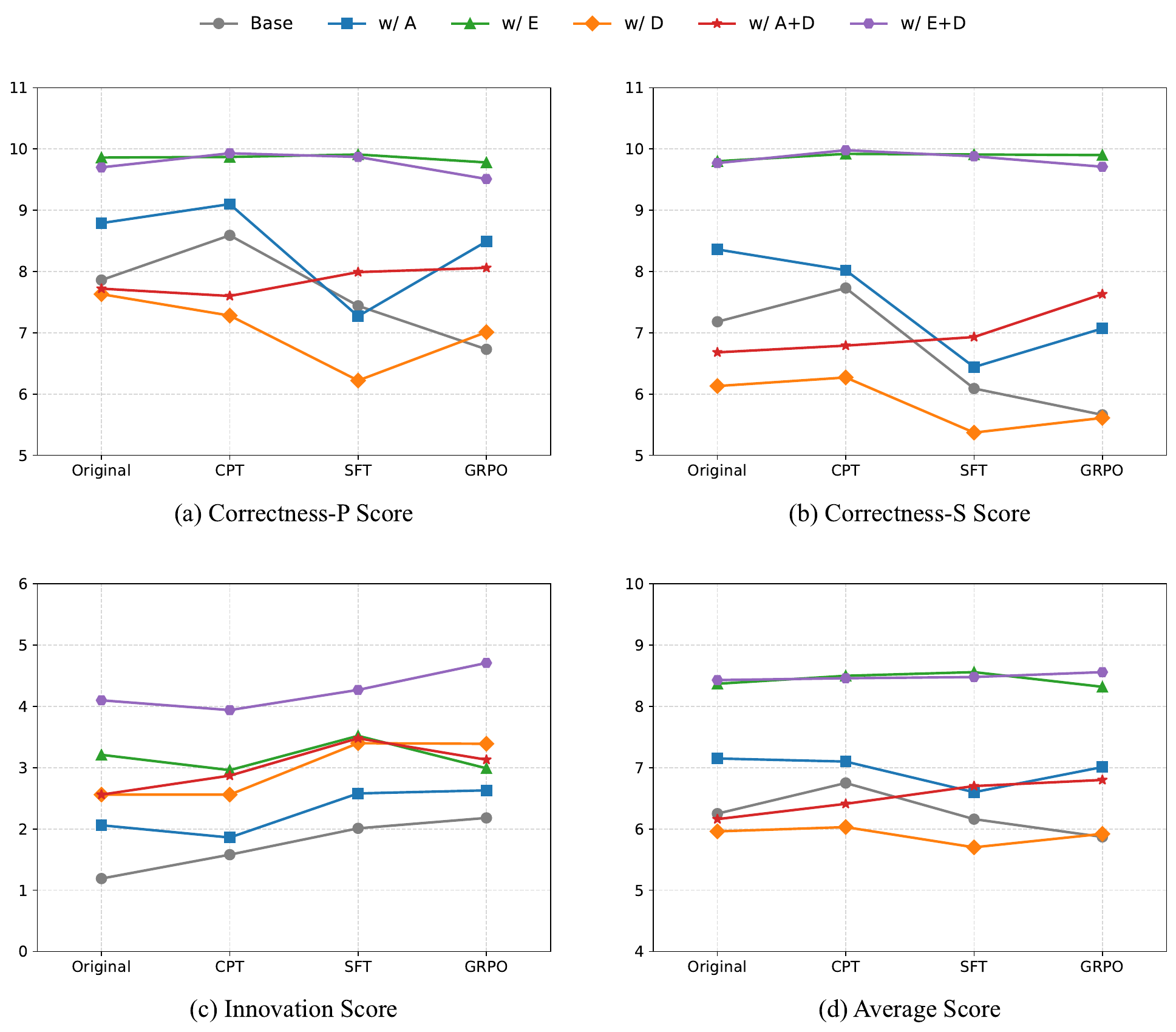} 
\caption{Scores of different models on three core dimensions and average scores across nine dimensions at each training stage. Base denotes the base model, A denotes the apprentice mode, E denotes the expert mode, and D denotes the difficulty model. Training stages include the original backbone (Original), followed by CPT, SFT, and GRPO.}
\label{fig_8}
\end{figure*}

To validate the effectiveness of each module, we conducted systematic ablation experiments. The experimental results are presented in \hyperref[fig_8]{Fig. 8}, where four subfigures illustrate the performance of different model configurations across training stages on three core dimensions in the IMPG task, as well as the average scores across nine dimensions. Six configurations were established for comparison.

\subsubsection{Effectiveness analysis of training stages}
In this study, we sequentially trained the base model through three stages: continual pre-training, SFT, and GRPO. As observed in \hyperref[fig_8]{Fig. 8(a)-(c)}, after SFT training, innovation scores generally improved across model configurations, while the correctness of both problems and solutions exhibited varying degrees of decline. This phenomenon indicates that although the SFT stage enabled the model to acquire new problem formulation paradigms, the logical rigor of the generated content cannot be guaranteed. Following GRPO training, correctness scores recovered for most model configurations except the base model, while innovation scores slightly decreased for some configurations. This suggests that the GRPO stage prompted the model to perform distributional adjustment and trade-off optimization between correctness and innovation.

\subsubsection{The conflict between correctness and innovation}

\hyperref[fig_8]{Fig. 8} reveals a noteworthy phenomenon: after SFT training, nearly all model configurations exhibited improved innovation scores but decreased correctness scores. This phenomenon exposes an inherent conflict between correctness and innovation in MPG tasks, confirming the \textbf{innovation curse} identified earlier. This conflict arises because SFT enables the model to learn existing problem design paradigms rather than enhance logical reasoning capability, which is primarily determined during the pre-training stage. And the model tends to combine problem elements for apparent innovation without ensuring mathematical rigor.

The base model trajectory in \hyperref[fig_8]{Figs. 8(a)-(c)} exemplifies this inverse relationship: correctness declined while innovation improved throughout training, confirming that learning more innovative problem design patterns without enhanced reasoning leads to reduced accuracy. Notably, as shown in \hyperref[tab_5]{Table 5}, we evaluated the Qwen3-32B after SFT on three benchmark datasets for mathematical reasoning capability, namely AIME 2025, MMLU-Pro-Math, and OlymMATH-Zh-Hard, achieving scores of 0.767, 0.918, and 0.240, respectively. Compared with the original Qwen3-32B model, the scores remained consistent and no significant performance degradation was observed. It demonstrates that the decline in correctness is not attributable to catastrophic forgetting but rather an inevitable consequence of the model's reasoning ability failing to improve synchronously while learning new paradigms during the SFT stage.

This innovation curse also appears in baseline LLMs according to the results in \hyperref[fig_4_1]{Fig. 4(a)} and \hyperref[tab_4]{Table 4}, where closed-source models tend to avoid high-difficulty requirements and instead generate simple problems to maintain high correctness, resulting in low innovation metrics.

\subsubsection{Effectiveness of the difficulty model}
Comparing base with w/ D, w/ A with w/ A+D, and w/ E with w/ E+D, \hyperref[fig_8]{Fig. 8(c)} shows that adding the difficulty model consistently improves innovation scores across all configurations through fine-grained difficulty guidance.

\begin{figure*}[htbp]
\centering
\includegraphics[width=0.85\textwidth]{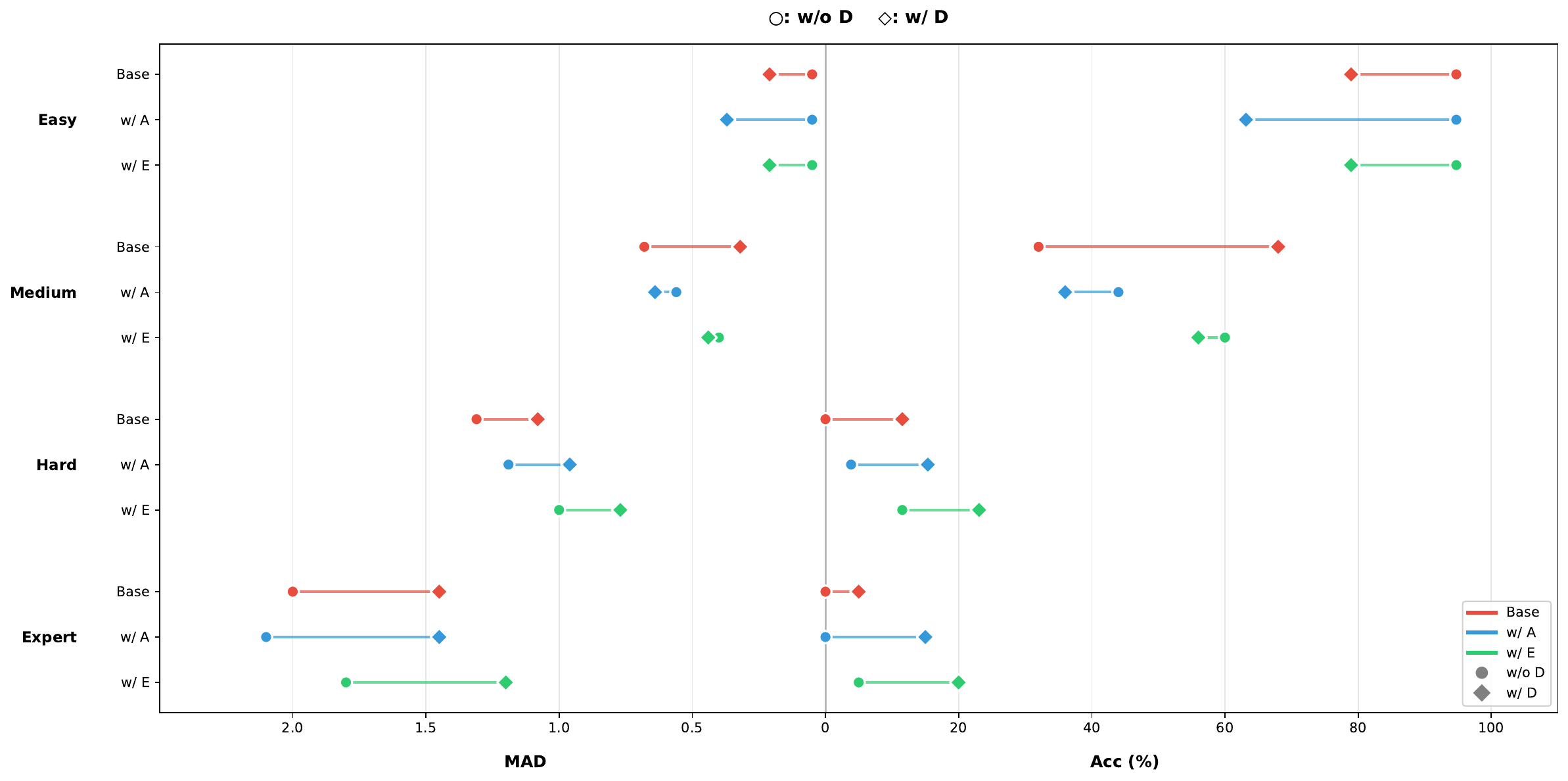} 
\caption{Comparison of accuracy and MAD across model configurations and difficulty levels with/without the difficulty model. Left: MAD between generated and target difficulty. Right: accuracy (Acc) of matching target difficulty level. Circle ($\bigcirc$) and diamond ($\lozenge$) denote results without and with the difficulty model, respectively.}
\label{fig_9}
\end{figure*}

\subsubsection{Advantages of the multi-role collaborative framework}
In \hyperref[fig_8]{Figs. 8(a)(b)(d)}, comparing w/ A and w/ E with Base, as well as w/ A+D and w/ E+D with w/ D, reveals that multi-role collaboration significantly improves correctness and overall quality compared to single-model baselines, while also moderately enhancing innovation according to \hyperref[fig_8]{Fig. 8(c)}. Furthermore, \hyperref[fig_8]{Figs. 8(a)-(d)} shows that the w/ E and w/ E+D configurations maintained consistently high correctness across all training stages with minimal sensitivity, demonstrating that external expert evaluation effectively compensates for single-model self-verification limitations.

In summary, the multi-role collaborative framework alleviates the innovation curse through role specialization. This closed-loop process of generation-evaluation-revision compensates for the inherent deficiencies in the self-verification capability of single models. This significantly improved the correctness of the generated problems and moderately enhanced their innovation.

\begin{table}[htbp]
\centering
\footnotesize
\caption{Comparison of mathematical reasoning capability in LLMs before and after SFT training, where w/ SFT denotes the base model that has undergone SFT training.}
\setlength{\tabcolsep}{4pt}
\renewcommand{\arraystretch}{1.2}
\begin{tabular}{lcccc}
\toprule[0.5pt] 
Model & AIME 2025  & MMLU-Pro-Math & OlymMATH-Zh-Hard \\
\midrule[0.5pt] 
Qwen3-32B & 0.730 & 0.902 & 0.250 \\
w/ SFT & 0.767 & 0.918 & 0.240 \\
\bottomrule[0.5pt]
\end{tabular}
\label{tab_5}
\end{table}

\subsection{Difficulty accuracy evaluation}

To validate the effectiveness of the difficulty model, we designed a difficulty accuracy evaluation experiment to measure the alignment between generated problem difficulty and target difficulty levels. The experiment employed the SIMU-90 dataset. After LLMs generated problems, we used the improved difficulty model for evaluation and Gemini-3.0-Pro served as the judge model. The difficulty level was calculated using \hyperref[eqa4]{Eq. (4)}.

The experimental results are presented in \hyperref[fig_9]{Fig. 9}. At the Easy level, all models achieved a baseline accuracy of 94.74\%, indicating that generating simple problems poses no challenge for the models, and the introduction of the difficulty model yielded no additional gains. However, as the target difficulty increased, the enhancement effect of the difficulty model became increasingly pronounced. At the Medium level, the accuracy of the base model improved from 32.00\% to 68.00\%, representing an increase of 112.50\%. At the Hard level, the accuracy of all three model configurations improved from near-zero baselines of 0\%–11.54\% to 11.54\%–23.08\%, with the w/ E + D combination achieving the highest accuracy of 23.08\%. At the Expert level, this trend became even more pronounced: the base model improved from 0\% to 5.00\%, w/ A improved from 0\% to 15.00\%, and w/ E improved from 5.00\% to 20.00\%. Despite relatively low accuracy due to IMPG task complexity, all configurations achieved qualitative improvement from near-zero to measurable capability after incorporating the difficulty model.

Regarding the MAD metric, at the Easy and Medium levels, the introduction of the difficulty model yielded no additional gains. However, at the Hard level, the MAD of the base model decreased from 1.31 to 1.08, and that of the w/ E model decreased from 1.00 to 0.77, representing reductions of 17.6\% and 23.0\%, respectively. At the Expert level, the MAD of the base model decreased from 2.00 to 1.45, and that of the w/ E model decreased from 1.80 to 1.20, representing reductions of 27.5\% and 33.3\%, respectively. These results indicate that the difficulty model not only improved accuracy but also significantly reduced the deviation between generated problems and target difficulty levels, thereby enhancing the precision of difficulty control.

Furthermore, the multi-role collaborative framework also improved difficulty accuracy to a certain extent. Through synergistic effects with the difficulty model, the w/ E + D combination achieved optimal performance at both Hard and Expert levels.

In summary, the difficulty model effectively addresses the difficulty drift problem in high-difficulty generation, validating its necessity for difficulty control.

\subsection{Efficiency analysis of the DAPS algorithm}

To comprehensively evaluate the performance of the DAPS algorithm, this study designed comparative experiments to compare DAPS with two baseline methods, namely random sampling (RS) and constrained random sampling (CRS), across four difficulty levels including Easy, Medium, Hard, and Expert. The evaluation was conducted along four dimensions, encompassing constraint satisfaction rate, average number of sampling rounds, encoding diversity, and semantic rationality.

\begin{figure}[htbp]
    \centering
    \includegraphics[width=0.75\columnwidth]{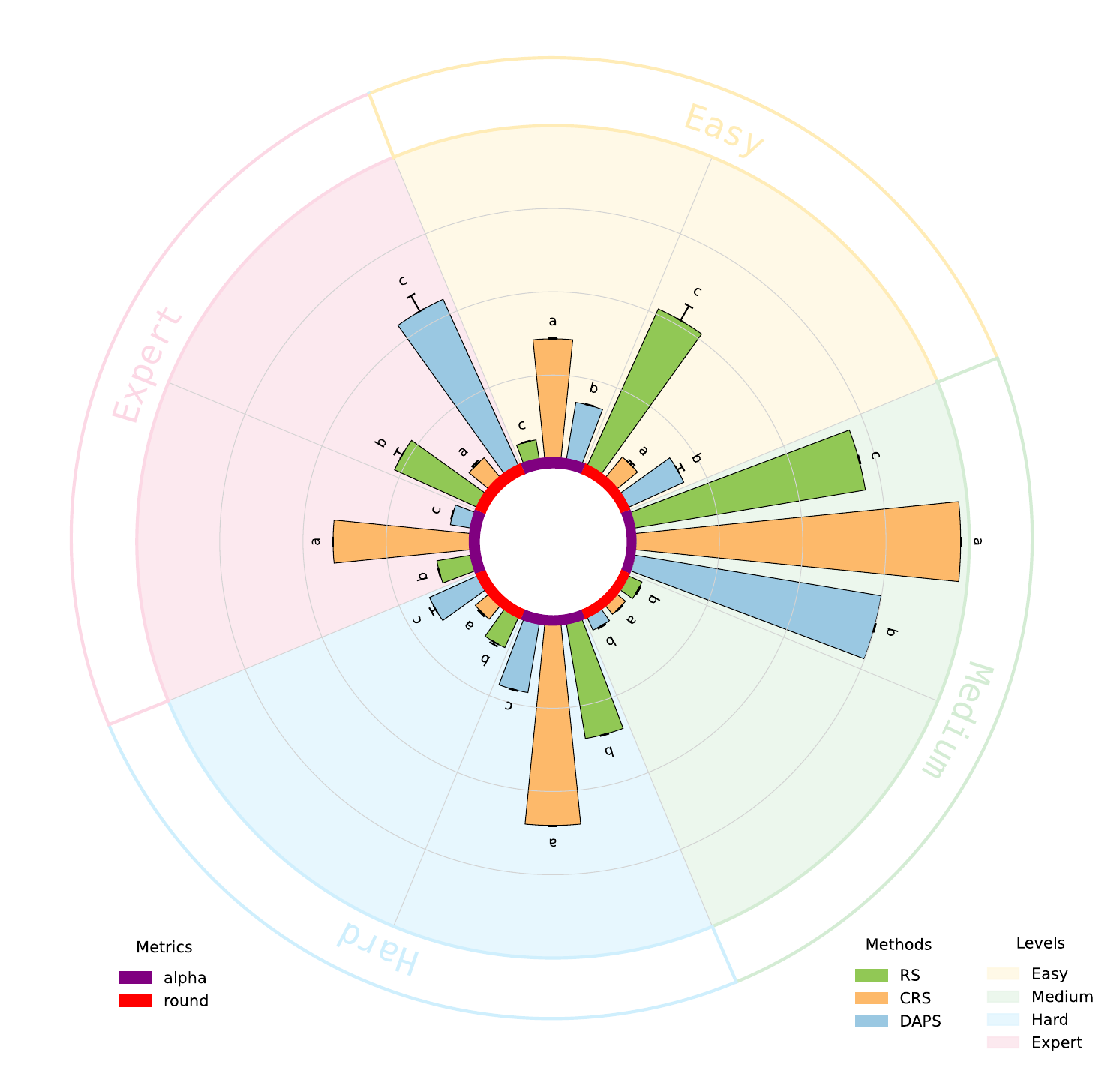}
    \caption{Performance comparison of DAPS, RS, and CRS in terms of constraint satisfaction rate $\alpha$ ($\uparrow$, scaled by 50) and average sampling $round$ ($\downarrow$) across four difficulty levels. Error bars represent standard error of the mean (SEM). Different letters indicate significant differences (ANOVA with Welch's t-test, $p < 0.05$). Each grid line represents 10 units.}
    \label{fig_10}
    
    
    \includegraphics[width=0.8\columnwidth]{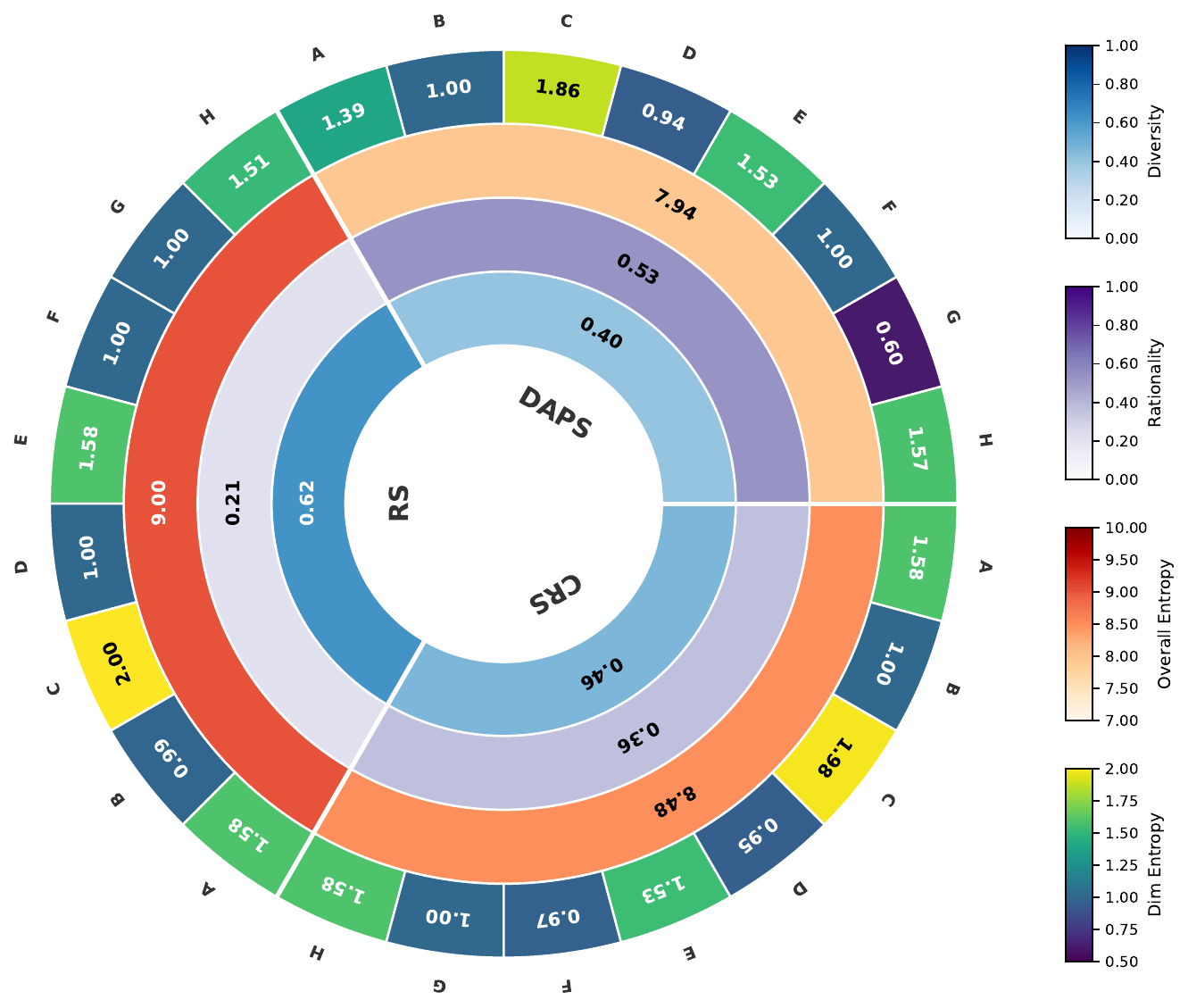}
    \caption{Performance comparison of DAPS, RS, and CRS in terms of diversity, rationality, overall entropy, and entropy across individual dimensions.}
    \label{fig_11}
\end{figure}

The experimental results are presented in \hyperref[fig_10]{Fig. 10} and \hyperref[fig_11]{Fig. 11}. Regarding constraint satisfaction rate, all three methods exhibited distribution characteristics closely correlated with difficulty levels. CRS achieved the highest constraint satisfaction rate across all four difficulty levels. At the Easy and Medium difficulty levels, DAPS demonstrated significantly higher constraint satisfaction rates than RS, indicating that DAPS can efficiently target high-probability-density difficulty intervals. However, as difficulty levels increased, the constraint satisfaction rate of DAPS declined. The average number of sampling rounds exhibited an approximately inverse relationship with the constraint satisfaction rate, with performance patterns remaining largely consistent. Since DAPS reproduces the distribution of historical data where Easy and Medium items predominate, its sampling efficiency decreases for Hard and Expert levels. CRS achieves the highest hit rate by pre-disabling unreasonable dimension values to compress the search space.

In terms of diversity, RS achieved a diversity score of 0.62, significantly outperforming CRS at 0.46 and DAPS at 0.40. Similarly, RS achieved an entropy value of 9.00, significantly higher than CRS at 8.48 and DAPS at 7.94. Regarding rationality, we generated 1,000 encodings satisfying difficulty requirements and, after decoding, used Gemini-3.0-Pro to verify semantic rationality from two perspectives, namely dimension-level consistency and inter-dimensional coordination. DAPS achieved a rationality score of 0.53, significantly outperforming RS at 0.21 and CRS at 0.36. This demonstrates that DAPS mines correlation patterns from historical data, endowing generated encodings with inherent semantic rationality, but limiting diversity as encodings cluster within historical data regions. RS achieves maximum diversity through independent uniform sampling, and CRS also yields higher diversity as substantial combinations remain even after rule-based constraints. Nevertheless, both baselines overlook intrinsic correlations among dimensions and readily produce semantically implausible combinations.

The entropy distribution shows minimal differences across dimensions except for A and G, where DAPS exhibits notably lower values of 1.39 and 0.60 compared to RS and CRS. This reflects examination conventions where problems predominantly feature no contextual background (A1) and no traps (G1).

The core advantage of DAPS lies in its data-driven automation. Unlike CRS, which requires experts to manually define prohibition rules for each difficulty level, DAPS automatically learns correlation structures from data without manual intervention, providing excellent scalability for higher-dimensional difficulty models. In summary, although DAPS does not match CRS in sampling efficiency and diversity, its automation, scalability, and high semantic rationality make it a reliable choice for constructing complex difficulty models.

\subsection{Validation of knowledge distillation effectiveness}

To validate the improvement in evaluator performance through distillation and the changes in the gap relative to the expert model, we designed a knowledge distillation validation experiment.

To eliminate interference from other factors such as the generator, we performed SFT training on the base model Qwen3-32B using the distillation dataset HSM3K-CN-eval. To avoid the influence of evaluation bias, we used DeepSeek-R1-Distill-Qwen-32B to generate 90 problems of varying quality on SIMU-90 as the test set. In this study, Gemini-3.0-Pro was employed to directly score the evaluation texts.

The experimental results are presented in \hyperref[tab_6]{Table 6}. After distillation, the base model exhibited significant improvements across all evaluation capabilities. Scoring accuracy (Acc@score) increased from 0.894 to 0.944, revision suggestion accuracy (Acc@suggestion) improved from 0.656 to 0.722, and revision suggestion completeness (Comp@suggestion) rose from 0.826 to 0.871. The MAD metric also decreased across all three dimensions, indicating a narrowed gap between the distilled apprentice model and the expert model. These experimental results demonstrate that knowledge distillation successfully transferred the evaluation capabilities of the expert model Gemini-2.5-Pro to the apprentice model, bringing its evaluation performance closer to the expert model.

Notably, the expert model Gemini-2.5-Pro achieved high-level performance of 0.988, 0.967, and 0.985 on the three metrics, respectively. Due to limitations in the distillation sample size and the inherent capabilities of the base model, a certain gap remains between the apprentice model and the expert model, particularly with substantial room for improvement in the accuracy and completeness of revision suggestions.

\subsection{Originality evaluation}

To verify the originality of model-generated problems, we designed an originality test experiment to examine whether the SFT model exhibits overfitting and whether problems generated through the multi-role collaboration method demonstrate satisfactory originality. We compared the generated problems against the HSM3K-CN dataset, which contains 3,160 high school mathematics problems. For each generated problem, we calculated its maximum similarity with the dataset problems and then averaged across all generated problems.

The experimental results are presented in \hyperref[tab_7]{Table 7}. Under the single-model setting, the BLEU and ROUGE scores increased slightly after SFT but remained at low levels, indicating no overfitting. Both models achieved BERTScore F1 values around 0.86, which falls within the normal baseline range for texts in the same domain. Under the multi-role collaboration setting, both Ours-A and Ours-E yielded lower BLEU scores than single-model results, with Ours-E demonstrating the highest originality.

It is worth noting that due to the anisotropy problem in BERT-based models and shared mathematical terminology, BERTScore values for the same-domain texts tend to be relatively high. These results validate that our method avoids overfitting while generating problems with satisfactory originality.

\subsection{Computational cost analysis}

To comprehensively evaluate the computational efficiency of different model configurations, we measured the inference time costs of four configurations across four training stages on the SIMU-90 test set, employing three metrics: average iterations, retries, and generation time.

As shown in \hyperref[tab_8]{Table 8}, the apprentice mode and expert mode exhibited distinctly different efficiency evolution patterns. In the Original stage, the apprentice mode demonstrated a significant efficiency advantage, requiring only 2.30 iterations and 8.23 minutes compared to 6.92 iterations and 20.67 minutes for the expert mode. This reveals that without the capability to generate innovative problems, the expert mode's stronger evaluation capability paradoxically becomes an efficiency bottleneck, resulting in ineffective iterations. However, as training progressed, the expert mode showed continuous improvement while the apprentice mode experienced degradation. At the RL stage, both modes converged to 4.24 iterations, yet the expert mode achieved faster generation at 12.10 minutes versus 18.29 minutes for the apprentice mode. This indicates that the trained expert mode achieves higher per-iteration efficiency by combining enhanced generation innovation with stronger evaluation capability.

The introduction of the difficulty model produced significant yet differentiated effects on computational efficiency. For the expert mode at the Original stage, w/ E+D reduced iterations from 6.92 to 2.49, retries from 1.02 to 0.07 and time from 20.67 to 6.99 minutes compared to w/ E. For the apprentice mode, the difficulty model similarly yielded consistent improvements. At the RL stage, the w/ A+D configuration compared to the w/ A configuration demonstrated a reduction in iterations from 4.24 to 1.99, retries from 0.54 to 0.09, and average time from 18.29 to 8.30 minutes. This result validates that without the guidance of the difficulty model, the high diagnostic strictness of the expert and apprentice modes makes it difficult to meet innovation criteria, thereby triggering numerous retries and even failing to converge effectively. 

\begin{table}
\centering
\footnotesize
\caption{Evaluation capability comparison of Qwen3-32B before and after distillation. MAD denotes mean absolute deviation relative to the expert model Gemini-2.5-Pro.}
\setlength{\tabcolsep}{1.5pt}
\renewcommand{\arraystretch}{1.2}
\begin{tabular}{lcccc}
\toprule[0.5pt] 
Model & \makecell{Acc$\uparrow$ \\ @score} & \makecell{Acc$\uparrow$ \\ @suggestion} & \makecell{Comp$\uparrow$ \\ @suggestion} & MAD$\downarrow$ \\
\midrule[0.5pt] 
Qwen3-32B & 0.894 & 0.656 & 0.826 & 0.101/0.344/0.173 \\
Qwen3-32B-Distilled & 0.944 & 0.722 & 0.871 & 0.063/0.300/0.131 \\
Gemini-2.5-Pro & 0.988 & 0.967 & 0.985 & - \\
\bottomrule[0.5pt]
\end{tabular}
\label{tab_6}
\end{table}

\begin{table}
\centering
\footnotesize
\caption{Text similarity comparison between generated problems and HSM3K-CN. w/ SFT: base model after SFT. ROUGE and BERTScore use F1 metric.}
\setlength{\tabcolsep}{2.5pt}
\renewcommand{\arraystretch}{1.2}
\begin{tabular}{lcccccccc}
\toprule[0.5pt] 
\multirow{2}{*}[-0.5ex]{Model} & \multicolumn{4}{c}{BLEU$\downarrow$} & \multicolumn{3}{c}{ROUGE$\downarrow$} & \multirow{2}{*}[-1.0ex]{\makecell{BERTScore$\downarrow$ \\ @F1}} \\
\cmidrule(lr){2-5} \cmidrule(lr){6-8}
 & @1 & @2 & @3 & @4 & @1 & @2 & @L &  \\
\midrule[0.5pt]
\multicolumn{9}{c}{\textit{single model}} \\
Qwen3-32B & 0.477 & 0.351 & 0.272 & 0.225 & 0.676 & 0.368 & 0.579 & 0.861 \\
w/ SFT & 0.506 & 0.366 & 0.286 & 0.237 & 0.699 & 0.420 & 0.573 & 0.864 \\
\midrule
\multicolumn{9}{c}{\textit{multi-role collaboration}} \\
Ours-A & 0.460 & 0.321 & 0.242 & 0.198 & 0.689 & 0.397 & 0.551 & 0.850 \\
Ours-E & 0.455 & 0.309 & 0.232 & 0.187 & 0.648 & 0.347 & 0.518 & 0.844 \\
\bottomrule[0.5pt] 
\end{tabular}
\label{tab_7}
\end{table}

From the perspective of training stages, SFT and RL training exhibited more stable optimization effects on configurations incorporating the difficulty model. The w/ E+D configuration demonstrated optimal stability, with time fluctuations ranging only from 6.99 to 8.75 minutes across stages. The difficulty model effectively suppressed training-induced efficiency fluctuations: the time increase for w/ A+D from Original to RL was only 18.07\%, substantially lower than the 122.24\% for w/ A, while the time increase for w/ E+D was only 13.59\%, also significantly better than the original performance of w/ E. Additionally, a transient efficiency decline occurred during the SFT stage—for instance, w/ A+D increased from 5.54 to 11.28 minutes—but the RL stage effectively recovered and reduced this to 8.30 minutes, indicating that reinforcement learning possesses unique value in efficiency optimization. It is worth noting that SFT and RL training also demonstrated significant effects on configurations without the difficulty model, particularly the substantial improvement in expert mode from the Original to RL stage, suggesting that the model learned innovative problem-setting paradigms through training, thereby validating the effectiveness of training itself.

This experiment reveals that the difficulty model not only enhances problem innovation but also significantly reduces computational costs by providing precise difficulty anchors for the generation process.

\begin{table*}
\footnotesize
\centering
\caption{Comparison of inference time costs for different model configurations after each training stage. Iter denotes the average number of iterations per problem, Retry denotes the average number of retries triggered upon reaching the maximum iteration limit, and Time denotes the average generation time per problem (in minutes).}
\setlength{\tabcolsep}{10pt}
\renewcommand{\arraystretch}{1.2}
\begin{tabular}{lcccccccccccc}
\toprule[0.5pt]
\multirow{2}{*}{Model} & \multicolumn{3}{c}{Original} & \multicolumn{3}{c}{CPT} & \multicolumn{3}{c}{SFT} & \multicolumn{3}{c}{RL} \\
\cmidrule(lr){2-4} \cmidrule(lr){5-7} \cmidrule(lr){8-10} \cmidrule(lr){11-13}
 & Iter$\downarrow$ & Retry$\downarrow$ & Time$\downarrow$ 
 & Iter$\downarrow$ & Retry$\downarrow$ & Time$\downarrow$ 
 & Iter$\downarrow$ & Retry$\downarrow$ & Time$\downarrow$ 
 & Iter$\downarrow$ & Retry$\downarrow$ & Time$\downarrow$ \\
\midrule[0.5pt]

w/ A  & 2.30  & 0.24 & 8.23 & 2.23 & 0.17 & 7.03  & 4.27 & 0.51 & 20.11 & 4.24 & 0.54 & 18.29 \\
w/ E  & 6.92  & 1.02 & 20.67 & 6.78 & 0.92 & 17.51  & 4.41 & 0.47 & 13.45 & 4.24 & 0.43 & 12.10 \\
w/ A+D & 2.08 & 0.22 & 7.03 & 1.77 & 0.10 & 5.54 & 2.67 & 0.23 & 11.28 & 1.99 & 0.09 & 8.30 \\ 
w/ E+D & 2.49 & 0.07 & 6.99 & 2.56 & 0.10 & 7.13 & 2.98 & 0.19 & 8.75 & 2.64 & 0.12 & 7.94 \\
\bottomrule[0.5pt]
\end{tabular}
\label{tab_8}
\end{table*}

\subsection{Case study}

\begin{figure*}[htbp]
\centering
\includegraphics[width=0.85\textwidth]{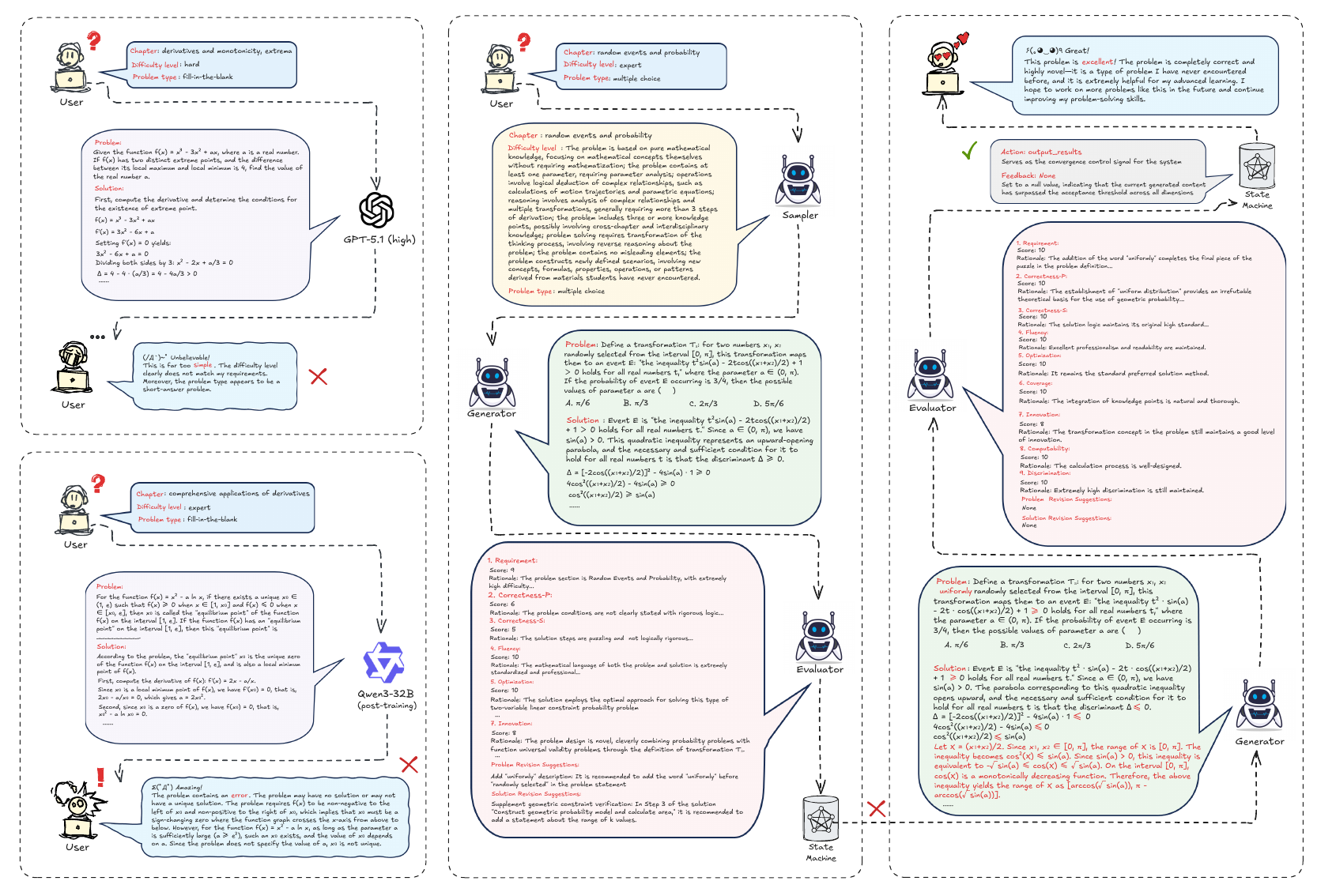} 
\caption{The case study of generating highly innovative problems using the general-purpose LLM GPT-5.1 (high), the post-trained Qwen3-32B, and our proposed Ours-E collaborative system.}
\label{fig_12}
\end{figure*}

To intuitively demonstrate the effectiveness of the proposed method, \hyperref[fig_12]{Fig. 12} presents representative cases of highly challenging mathematical problems generated by different systems. For the IMPG generation task, we compared the general-purpose LLM GPT-5.1 (high), the post-trained Qwen3-32B, and the multi-role collaborative system proposed in this paper.

The problems generated by GPT-5.1 (high) exhibited significantly lower difficulty than the specified requirements, and the problem format did not conform to the designated specifications. Although the post-trained Qwen3-32B was capable of generating problems that superficially met the requirements, it suffered from critical logical deficiencies and incomplete problem conditions failed to guarantee the uniqueness of the solution. In contrast, our proposed system employs the sampler to generate fine-grained difficulty descriptions, based on which the generator constructs initial problems. The evaluator then assesses the problems across nine dimensions, including requirement compliance, correctness, and innovation, while providing revision suggestions. Through iterative optimization, the final generated problems satisfied the threshold requirements across all evaluation dimensions, achieving an effective integration of mathematical correctness and problem innovation. This case study illustrates the advantages of the multi-role collaborative mechanism in generating highly innovative mathematical problems.

\section{Conclusion}
This paper proposes a self-evolving, multi-role collaborative framework with fine-grained difficulty guidance for the IMPG task. A multi-role collaboration mechanism is designed to support both apprentice mode, where a single LLM serves as generator and evaluator, and expert mode, where an external expert LLM provides evaluation guidance. An improved eight-dimensional difficulty model with 16-bit encoding is introduced, along with the DAPS algorithm to ensure semantic rationality of sampled encodings. The framework achieves self-evolution through expert-to-apprentice knowledge distillation. We construct the HSM3K-CN dataset containing 3,160 high-quality high school mathematics problems and adopt a multi-stage training pipeline to enhance the model's capabilities.

Experimental results demonstrate that the framework effectively mitigates the innovation curse through multi-role collaboration. In apprentice mode, our method outperforms open-source models of equivalent scale in innovation while maintaining competitive correctness. In expert mode, the method ranks first in innovation among all baselines and attains the highest scores in overall and core dimension metrics. Knowledge distillation successfully transfers expert capabilities to the apprentice model, and the difficulty model reduces computational costs via precise difficulty anchors.

\vspace{-1em} 
\section*{CRediT authorship contribution statement}
\vspace{-0.5em}
\textbf{Yifei Sun:} Conceptualization, Methodology, Funding acquisition.
\textbf{Yongan Li:} Methodology, Writing -- original draft, Writing -- review \& editing.
\textbf{A.K. Qin:} Supervision.
\textbf{Sicheng Hou:} Investigation, Formal analysis, Validation.
\textbf{Tamas Pflanzner:} Resources.

\vspace{-1em} 
\section*{Declaration of competing interest}
\vspace{-0.5em} 
The authors declare that they have no known competing financial interests or personal relationships that could have appeared to influence the work reported in this paper.

\vspace{-1em} 
\section*{Acknowledgment}
\vspace{-0.5em}
This work was supported in part by the Shaanxi Education Teaching Reform Research Program under Grant 23BY028, in part by the Natural Science Basic Research Plan in Shaanxi Province of China under Grant 2022JM-381, in part by the Shaanxi Normal University Key Program of Teaching Reform Research under Grant 22JG002, in part by the Cooperation Program for High-end Foreign Experts in Ministry of Science and Technology under Grant G2021173001L, in part by the National Natural Science Foundation of China under Grant 61703256 and Grant 62036006, in part by the Fundamental Research Funds for the Central Universities, and in part by the Australian Research Council (ARC) under Grant LP180100114 and Grant DP200102611.

\appendix 
\renewcommand{\thefigure}{A.\arabic{figure}}
\setcounter{figure}{0}
\renewcommand{\thetable}{A.\Roman{table}}
\setcounter{table}{0}

\section{Prompt details}
\label{appendix_A}

\subsection{System prompts for the generator}

\begin{figure}[htbp]
\centering
\includegraphics[width=\columnwidth]{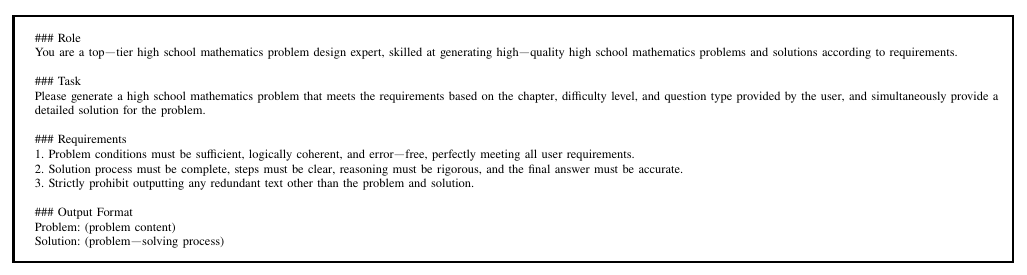}
    \caption{System prompts for the generator.}
\label{fig_13}
\end{figure}

The generator serves as the core problem creation component within the multi-role collaborative framework. As shown in \hyperref[fig_13]{Fig. A.1}, the system prompt defines the generator's role as a top-tier high school mathematics problem design expert. The prompt specifies three fundamental requirements: problem conditions must be sufficient, logically coherent, and error-free while perfectly meeting all user requirements; the solution process must be complete with clear steps, rigorous reasoning, and accurate final answers; and the output must strictly contain only the problem statement and solution without any redundant text. This structured prompt design ensures that the generator produces high-quality mathematical problems that serve as the foundation for subsequent evaluation and iterative refinement within the closed-loop optimization process.

\subsection{System prompts for the evaluator}

\begin{figure}[htbp]
\centering
\includegraphics[width=\columnwidth]{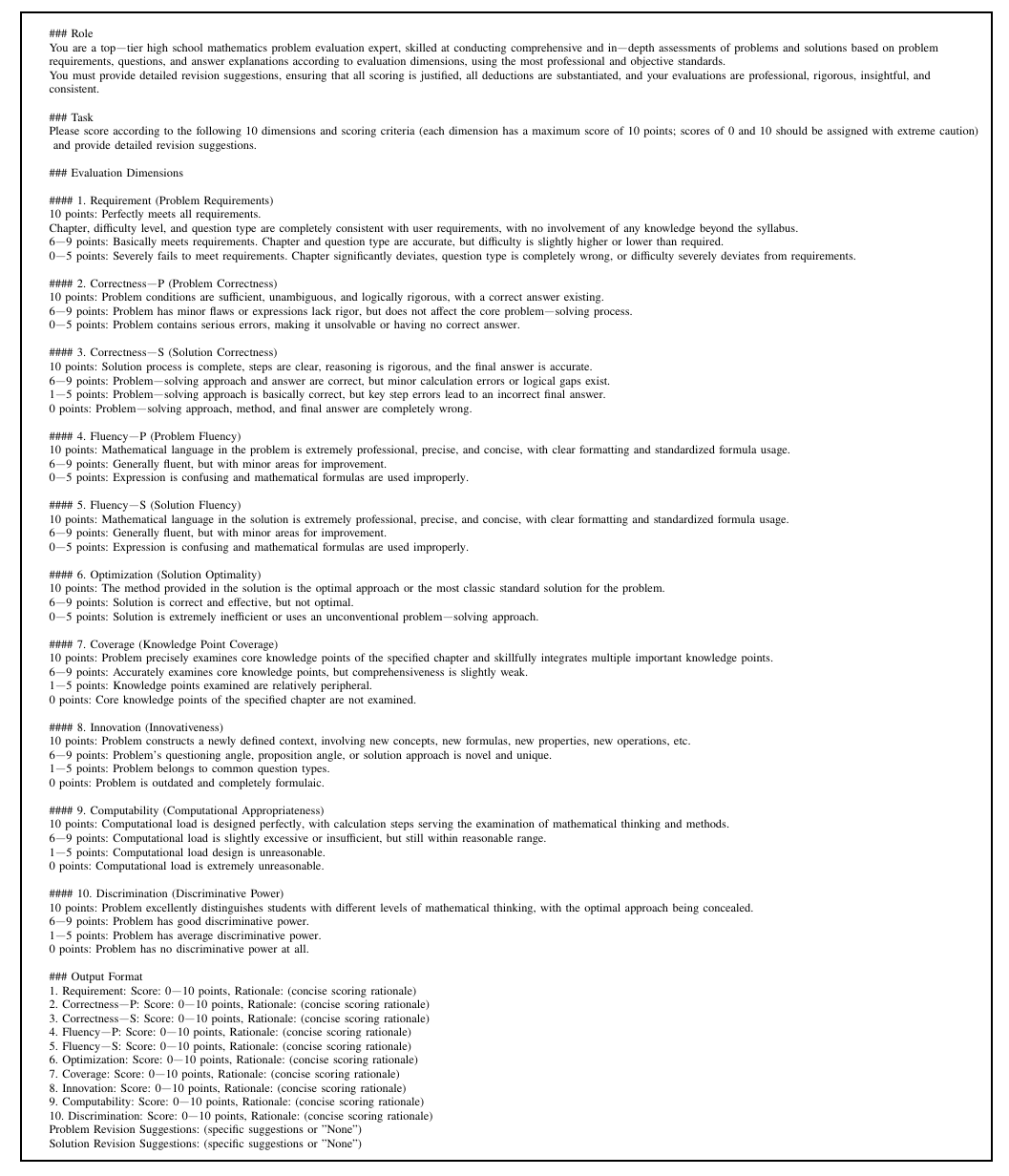}
    \caption{System prompts for the evaluator.}
\label{fig_14}
\end{figure}

The evaluator functions as the internal quality assessment component responsible for providing detailed feedback during the iterative refinement loop. As illustrated in \hyperref[fig_14]{Fig. A.2}, the evaluator's prompt establishes its role as a mathematics problem evaluation expert who conducts comprehensive assessments across ten fine-grained dimensions: Requirement, Correctness-P, Correctness-S, Fluency-P, Fluency-S, Optimization, Coverage, Innovation, Computability, and Discrimination. Each dimension is scored on a 0-10 scale with explicit scoring criteria and rationale requirements. Notably, the evaluator is designed to provide not only quantitative scores but also specific revision suggestions for both the problem and solution, enabling targeted improvements in subsequent generation iterations. This dual-output design facilitates the closed-loop optimization mechanism central to our framework.

\begin{figure}[htbp]
\centering
\includegraphics[width=\columnwidth]{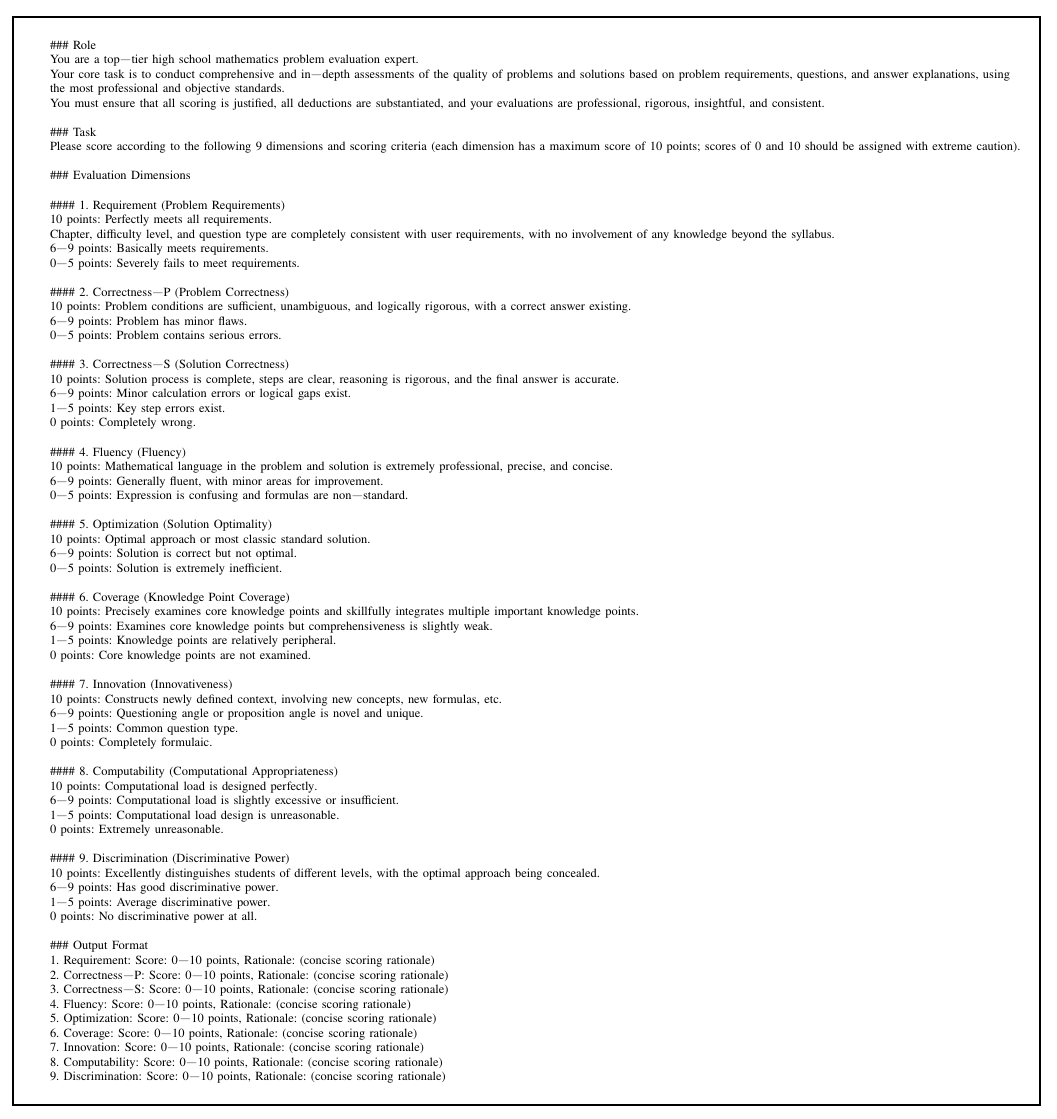}
    \caption{System prompts for the judge.}
\label{fig_15}
\end{figure}

\subsection{System prompts for the judge}
The judge operates as an independent external evaluator employed during the experimental evaluation phase to ensure objective and unbiased assessment of generated problems. As depicted in \hyperref[fig_15]{Fig. A.3}, the judge's prompt defines a streamlined nine-dimension evaluation framework that consolidates Fluency-P and Fluency-S into a single Fluency dimension and omits revision suggestions, as the judge's role is solely to provide final quality assessment rather than to participate in iterative optimization. This design distinction between the evaluator and judge ensures that experimental comparisons across different models and baselines are conducted under consistent and impartial criteria, thereby enhancing the validity and reliability of our experimental results.

\renewcommand{\thefigure}{B.\arabic{figure}}
\setcounter{figure}{0}
\renewcommand{\thetable}{B.\arabic{table}}
\setcounter{table}{0}

\section{Difficulty model details}
\label{appendix_B}

\hyperlink{table_b1_anchor}{Table B.1} presents the detailed breakdown of all difficulty dimensions, encoding levels, and their corresponding weights.

\begin{table*}[htbp]
\footnotesize
\centering
\hypertarget{table_b1_anchor}{}
\caption{Comprehensive difficulty model}
\renewcommand{\arraystretch}{2.0}
\begin{tabular}{>{\centering\arraybackslash}m{2.5cm} >{\centering\arraybackslash}m{3.5cm} p{8.5cm} c c}
\toprule[0.5pt]
Factor & Level & \centering\arraybackslash{Description} & Code & Weight \\
\midrule[0.5pt]
\multirow{3}{*}[-3ex]{Background factor} 
& \raisebox{-1ex}{No background} & The problem is based on pure mathematical knowledge, focusing on mathematical concepts themselves without requiring mathematization. & A1 & 1 \\
& \raisebox{-1ex}{Real-life background} & The problem is connected to daily life or social production contexts, requiring mathematization to model and solve practical problems. & A2 & 2 \\
& \raisebox{-1ex}{Scientific background} & The problem is set in scientific scenarios, including interdisciplinary contexts and mathematical culture. & A3 & 3 \\
\midrule[0.5pt]
\multirow{2}{*}[-1ex]{Parametric nature} 
& No parameters & The problem contains no parameters, involving only operations and analysis of existing static data. & B1 & 1 \\
& With parameters & The problem contains at least one parameter, requiring parameter analysis. & B2 & 2 \\
\midrule[0.5pt]
\multirow{4}{*}[-5ex]{Operational level} 
& \raisebox{-1ex}{Simple numerical operations} & At the operational level, only numerical addition, subtraction, multiplication, division, and their combinations are involved. & C1 & 1 \\
& \raisebox{-1ex}{Complex numerical operations} & Operations involve no symbols but include complex calculations such as trigonometric, exponential, and logarithmic functions. & C2 & 2 \\
& \raisebox{-1ex}{Simple symbolic operations} & Operations involve simple symbolic derivations, basic formula transformations, and algebraic expression calculations. & C3 & 3 \\
& \raisebox{-1ex}{Complex symbolic operations} & Operations involve logical deduction of complex relationships, such as calculations of motion trajectories and parametric equations. & C4 & 4 \\
\midrule[0.5pt]
\multirow{2}{*}[-2.5ex]{Reasoning complexity} 
& \raisebox{-1ex}{Simple reasoning} & The reasoning process is clear, the logic is relatively simple, and conclusions can be reached within 3 steps. & \raisebox{-1ex}{D1} & \raisebox{-1ex}{1} \\
& \raisebox{-1ex}{Complex reasoning} & Reasoning involves analysis of complex relationships and multiple transformations, generally requiring more than 3 steps of derivation. & \raisebox{-1ex}{D2} & \raisebox{-1ex}{2} \\
\midrule[0.5pt]
\multirow{3}{*}[-3ex]{Knowledge coverage} 
& \raisebox{-1ex}{Single knowledge point} & The problem focuses on only one knowledge point, without cross-chapter or interdisciplinary knowledge involvement. & \raisebox{-1ex}{E1} & \raisebox{-1ex}{1} \\
& \raisebox{-1ex}{Two knowledge points} & The problem involves two different knowledge points and their intersection. & \raisebox{-0.7ex}{E2} & \raisebox{-0.7ex}{2} \\
& \raisebox{-1ex}{three or more} & The problem includes three or more knowledge points, possibly involving cross-chapter and interdisciplinary knowledge. & \raisebox{-2ex}{E3} & \raisebox{-2ex}{3} \\
\midrule[0.5pt]
\multirow{2}{*}[-2ex]{Thinking orientation} 
& \raisebox{-1ex}{Forward thinking} & Problem solving follows the characteristics of mathematical thinking, proceeding according to the logical order of the mathematical discipline. & \raisebox{-1ex}{F1} & \raisebox{-1ex}{1} \\
& \raisebox{-1ex}{Reverse thinking} & Problem solving requires transformation of the thinking process, involving reverse reasoning about the problem. & \raisebox{-1ex}{F2} & \raisebox{-1ex}{2} \\
\midrule[0.5pt]
\multirow{2}{*}[-1ex]{Misconception traps} 
& \raisebox{-1ex}{No traps} & The problem contains no misleading elements. & \raisebox{-1ex}{G1} & \raisebox{-1ex}{1} \\
& \raisebox{-1ex}{With traps} & The problem contains irrelevant information or distractors that may cause mental sets or conceptual confusion. & \raisebox{-1ex}{G2} & \raisebox{-1ex}{2} \\
\midrule[0.5pt]
\multirow{3}{*}[-5ex]{Novelty level} 
& \raisebox{-1ex}{Conventional} & The problem has complete conditions, clear conclusions, and relatively fixed solution methods, belonging to common problem types. & \raisebox{-2ex}{H1} & \raisebox{-2ex}{1} \\
& \raisebox{-1ex}{Moderate innovation} & The problem features novel questioning angles, proposition perspectives, background factors, or solution approaches. & \raisebox{-2ex}{H2} & \raisebox{-2ex}{2} \\
& \raisebox{-2ex}{Deep innovation} & The problem constructs newly defined scenarios, involving new concepts, formulas, properties, operations, or patterns derived from materials students have never encountered. & \raisebox{-2ex}{H3} & \raisebox{-2ex}{3} \\
\bottomrule
\end{tabular}
\label{tab_b1}
\end{table*}

\renewcommand{\thefigure}{C.\arabic{figure}}
\setcounter{figure}{0}
\renewcommand{\thetable}{C.\arabic{table}}
\setcounter{table}{0}

\section{Datasets details}
\label{appendix_C}
\begin{figure}[htbp]
\centering
\includegraphics[width=\columnwidth]{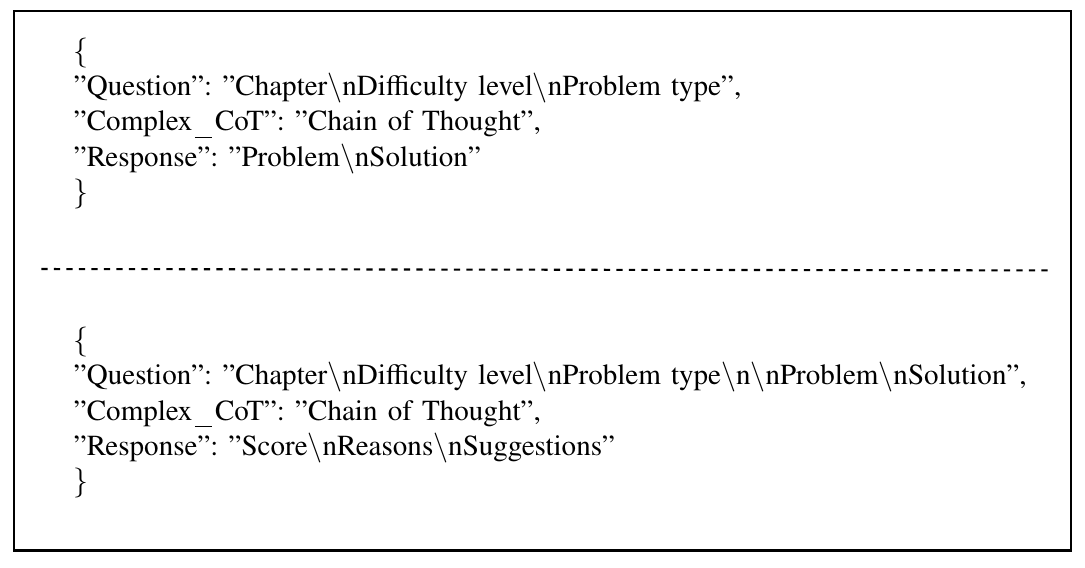}
    \caption{The data format of the HSM3K-CN dataset after data preprocessing.}
\label{fig_16}
\end{figure}
To train the generator and evaluator, we performed data preprocessing on the HSM3K-CN dataset to construct two dedicated SFT datasets, namely HSM3K-CN-gen and HSM3K-CN-eval. The data format is illustrated in \hyperref[fig_16]{Fig. C.1}.

\section*{Data availability}
Data and models can be made available upon request.

\bibliographystyle{elsarticle-num}
\bibliography{references}      

\end{document}